\documentclass{article}
\usepackage{iclr2026_conference,times}

\usepackage{amsmath,amsfonts,bm}

\def\eqref#1{equation~\ref{#1}}
\def\1{\bm{1}}

\DeclareMathAlphabet{\mathsfit}{\encodingdefault}{\sfdefault}{m}{sl}
\SetMathAlphabet{\mathsfit}{bold}{\encodingdefault}{\sfdefault}{bx}{n}

\usepackage{hyperref}
\hypersetup{
    colorlinks=true,
    linkcolor=red,
    citecolor=black,
    urlcolor=blue
}
\usepackage{url}
\usepackage{graphicx}
\usepackage{cleveref}
\usepackage{amsmath,amssymb}
\usepackage[compatibility=false]{caption}
\usepackage{subcaption}
\usepackage{wrapfig}
\usepackage{multicol}
\usepackage{multirow}
\usepackage{booktabs}
\usepackage[dvipsnames]{xcolor}
\usepackage[most]{tcolorbox}
\tcbuselibrary{breakable}
\usepackage{booktabs,multirow,makecell,tabularx,wrapfig}
\usepackage{enumitem}
\setlist[itemize]{leftmargin=*}
\setlist[enumerate]{leftmargin=*}

\newtcolorbox{infobox}[2][]{%
  breakable,
  enhanced,
  title={#1},
  colback=#2!8,
  colframe=#2!70!black,
  coltitle=white,
  fonttitle=\bfseries,
  boxrule=0.8pt,
  arc=2mm,
  left=6pt,right=6pt,top=6pt,bottom=6pt
}

\title{Scaling Laws and Symmetry,\\ Evidence from Neural Force Fields}

\iclrfinalcopy

\author{Khang Ngo $^{1~2}$, Siamak Ravanbakhsh $^{1~2}$ \\
$^1$ Mila - Quebec AI Institute, $^2$ School of Computer Science, McGill University \\
\texttt{\{khang.ngo,
siamak.ravanbakhsh\}@mila.quebec} \\
}

\begin{document}

\maketitle

\begin{abstract}
 We present an empirical study in the geometric task of learning interatomic potentials, which shows equivariance matters even more at larger scales; we show a clear power-law scaling behaviour with respect to data, parameters and compute with ``architecture-dependent exponents''.
 In particular, we observe that equivariant architectures, which leverage task symmetry, scale better than non-equivariant models. Moreover, among equivariant architectures, higher-order representations translate to better scaling exponents.
 Our analysis also suggests that for compute-optimal training, the data and model sizes should scale in tandem regardless of the architecture.
At a high level, these results suggest that, contrary to common belief, we should not leave it to the model to discover fundamental inductive biases such as symmetry, especially as we scale, because they change the inherent difficulty of the task and its scaling laws. \footnote{Code is available at \ttfamily{https://github.com/nnkhang19/scaling-laws-and-symmetry}}
\end{abstract}

\begin{figure}[h]
    \centering
      \begin{subfigure}[b]{0.49\textwidth} 
        \centering
        \includegraphics[width=\linewidth]{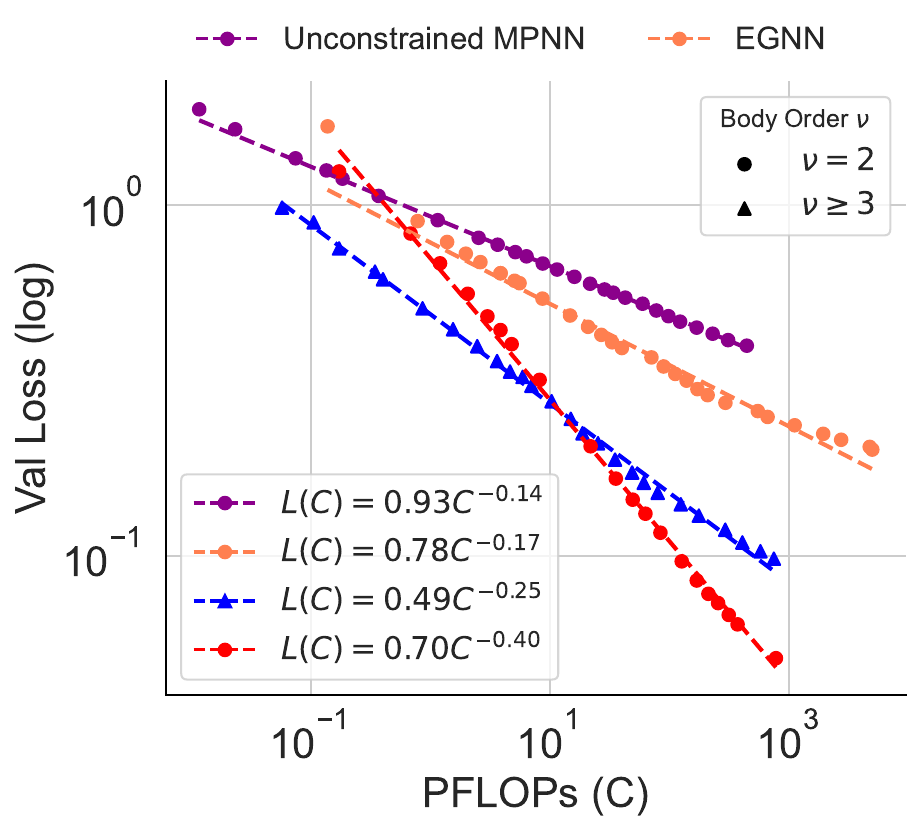}
        \label{fig:subfigA}
    \end{subfigure}
    \begin{subfigure}[b]{0.49\textwidth}
        \centering
        \includegraphics[width=\linewidth]{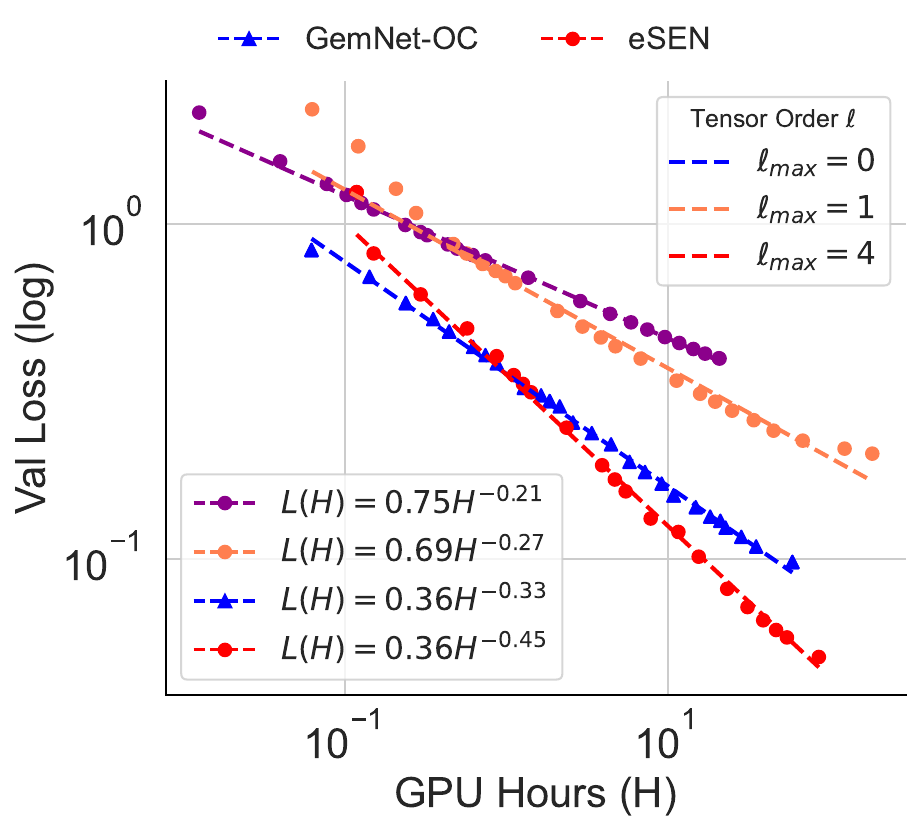}
        \label{fig:subfigB}
    \end{subfigure} 
    \vspace{-10pt}
    \caption{Performance of neural network interatomic potentials follows a power law (linear in log-log space) in training compute (PFLOPs, GPU-hours). The scaling behaviour varies with architectural complexity: the slope of the performance curve improves as the architecture changes from unconstrained to low-order to high-order, implying that performance gaps widen with increasing compute. \textit{Body
order} $\nu$: number of nodes whose states define a message within a layer. \textit{Tensor order} $\ell$: order of geometric features processed by the models. \textbf{Left}: Empirical scaling laws along the FLOPs-optimal frontier. \textbf{Right}: Empirical scaling laws along the train-time-optimal frontier. }
    \label{fig:champion}
    \vspace{-1em}
\end{figure}

\section{Introduction}
Recent years have witnessed extensive study of neural scaling laws across various machine learning domains, including natural language and vision.
%\citep{kaplan2020scaling, hoffmann2022chinchilla} and vision  \citep{zhai2022scaling}. 
% Empirical findings underpinning these theories demonstrate that scaling datasets and models with optimal compute substantially enhances the performance and generalization of deep learning models. 
The general observation supported by the theory is that test errors exhibit a power-law relationship with the scale of training data points, model parameters, and the amount of compute in floating-point operations (FLOPs). These laws identify the optimal scaling of model size with dataset size for a given compute budget, enabling an optimal use of resources at scale. 
% They have emerged as highly effective practical guidance for the development of large models in the current era of large-scale pre-training. As a direct result, billion parameter models pre-trained on massive amount of data and compute, have demonstrated impressive zero-shot capabilities in tackling a diverse array of tasks, ranging from question answering to conditional image generation. Motivated by these observations, we ask:  
% \begin{equation}
%     \textit{Do message-passing NNIPs follow power scaling laws?}
%     \label{question_1}
% \end{equation}

A common view is that the scaling behaviour is consistent across various expressive architectures for a task---i.e., the choice of architecture can only increase or decrease the loss by a multiplicative factor that remains constant across sufficiently large scales. This belief is supported by both theoretical results \citep{utkarsh2022scalingmanifold,bahri2024explaining_nn_scaling} and empirical studies in different domains \citep{ahmad1998scaling, hestness2017deep}, including language \citep{kaplan2020scaling, hoffmann2022chinchilla} and vision \citep{zhai2022scaling}, %; for example, comparing the scaling of Transformers and Long Short Term Memory (LSTM) networks.
% These observations have reinforced Sutton's bitter lesson; here we quote a  particularly relevant passage:
% \begin{quote} ``we should stop trying to find simple ways to think about the contents of minds, such as simple ways to think about space, objects, multiple agents, or symmetries. All these are part of the arbitrary, intrinsically-complex, outside world. They are not what should be built in, as their complexity is endless; instead we should build in only the meta-methods that can find and capture this arbitrary complexity.``
% \end{quote}
%Moreover, for finite groups, it is easy to foresee that leveraging the symmetry can at best improve the efficiency by a constant factor corresponding to the group size. It would appear surprising if, for continuous groups, this boosting factor were growing with scale. We report such an unexpected behaviour.
and it is further reinforced by Sutton’s bitter lesson \citep{sutton2019bitter}, which highlights that attempts to encode inductive biases, such as symmetry, explicitly are often outperformed in the long run, since models can learn these structures on their own when scaled.

The specific inductive bias of symmetry, and in particular 
 Euclidean and rotational symmetry have been successfully leveraged in many domains, including for molecular force fields. %\citep{thomas2018tensor, anderson2019cormorant, batatia2022mace, liao2024equiformerv}.%, and protein/materials generation \citep{jiao2023diffscp, xie2022cdvae}. 
 The success of these networks is often attributed to their improved generalization and robustness to out-of-distribution data \citep{batatia2022mace, petrache2023approximation}.  However, one may argue that equivariant networks are harder to scale as their specialized operations, such as tensor products, spherical harmonics \citep{thomas2018tensor, anderson2019cormorant, liao2023equiformer}, or high-order message passing \citep{Gasteiger2020Directional, gemnet}, are complex and computationally expensive. At the same time, several works in protein folding \citep{abramson2024alphafold3}, molecule conformer generation \citep{wang2024swallowing}, and {Neural Network Interatomic Potentials (NNIPs)} \citep{deng2023chgnet, qu2024scalable, rhodes2025orb} demonstrate that non-equivariant networks can perform well in geometric tasks. \cite{brehmer2025does} also show that non-equivariant architectures trained with data augmentation can perform on par with their equivariant counterparts when given sufficient compute.  
All of this paints a picture in favour of forgoing equivariance and scaling simpler non-equivariant models.

% combined with data augmentations or symmetry-constraint loss functions
% An overgeneralization of Sutton's bitter lesson, combined with the theoretical and empirical evidence mentioned above, has led to a growing trend of ignoring inductive biases, such as symmetry in the architecture, and instead adopting simple, expressive, and GPU-friendly architectures. Indeed, widely successful architectures that are trained at scale, such as Alphafold 3 \citep{abramson2024alphafold3} in molecular biology and ORB \citep{rhodes2025orb} in molecules and materials, are non-equivariant, despite having exact domain/task symmetries. 

This paper presents a careful empirical study that questions this growing mindset and shows that equivariance matters even more as we scale. We report a clear architecture-dependent scaling exponent in model size, data size, and compute, {for several widely used scalable NNIPs architectures that encode rotational and permutation symmetry to varying degrees}. This translates to a performance gap that grows with scale, favouring scalable models with a higher-order symmetry bias at larger scales; see \cref{fig:champion}. 

Our target domain for this study has witnessed a growing number of deep learning techniques for predicting quantum properties of atomistic systems in recent years, where neural models approximate computationally demanding {ab initio} calculations, such as density functional theory. The most promising progress is being made on NNIPs, which map molecular systems to their energies and forces. NNIPs' foundation models are unlocking new possibilities through efficient and accurate molecular dynamics, and our findings in this domain identify the most promising direction for the design of models that are trained at scale. 
\vspace{-10pt}
\subsection{Key Findings and Contributions}
In this work, we conduct comprehensive scaling-law experiments, drawing on best practices and insights from prior work on the expressive power of (geometric) message passing neural networks (MPNNs) \citep{Loukas2020What, joshi2023gwl}, maximal update parametrization ($\mu$P) \citep{yang2021muP}, and compute-optimal scaling \citep{hoffmann2022chinchilla}. Our key findings are:

\begin{itemize}
\item \textbf{\textit{Clear power law scaling.}} Message-passing NNIPs obey power-law scaling with respect to compute, data, and model size. For compute, unlike prior studies that report only FLOPs within architectures, we characterize scaling with both FLOPs and wall-clock training time (GPU-hours). Given that equivariant networks can be less GPU-friendly, this approach provides a more complete view for practical purposes.
While prior work in geometric domains has shown architecture-dependent scaling with respect to only the dataset size \citep{nequip}, to our knowledge, none of them provide a complete and comparable picture to ours.
\item \textbf{\textit{Architecture-dependent exponents.}} Power-law exponents increase as the ``degree'' of equivariance grows, from non-equivariant (unconstrained) models to lower- to higher-order equivariant designs. 
% \Cref{fig:champion} summarizes our main results. 
% This means an optimal investment of additional compute favours model-size for equivariant networks compared to non-equivariant ones.

\item \textbf{\textit{Compute-optimal scaling.}}
We find that the power-law exponents for dataset and model size in a compute-optimal scaling are similar across non-equivariant and equivariant architectures of different representation degrees. This means that a compute-optimal scaling should increase the model and dataset size in tandem; this mirrors the findings of \cite{hoffmann2022chinchilla} in natural language.

\item {\textbf{\textit{Multi-epoch training and data-augmentation.}}
While our main results consider a single epoch regime, we show that the same scaling laws hold across tens of epochs in a multi-epoch setting. This is because, at scale -- even with 1\% of our training set -- the effect of overfitting is negligible. For non-equivariant models,  data augmentation is required to avoid overfitting and maintain the scaling coefficients.  We also consider inference-time augmentation for the unconstrained model, and show that it only changes the multiplicative coefficient (rather than the exponent) in the scaling law, and its benefit saturates quickly with the number of augmentations for this task.}

\item \textbf{\textit{Scaling effect of symmetry loss.}} Enforcing symmetry through loss does NOT seem to provide the same benefits as having an equivariant architecture.
% \textbf{Inference-time scaling.} Inference-time scaling using augmentation does not show a power-law trend....  \\

\item \textbf{\textit{Trend in optimal depth.}} For a fixed parameter and compute budget, the optimal depth of the network is correlated with the ``degree'' of equivariance among the architectures we studied; with equivariant networks, the benefit of depth saturates at higher values, and for higher rotation order networks this value grows larger; this corroborates the claims of \cite{joshi2023gwl, push_limit_of_md_100m}.\\
% \textbf{Trend for Batch size} In contrast with language and vision domains \citep{kaplan2020scaling, bachmann2023scalingmlps}, smaller batch-sizes perform better, for fixed FLOPs.\\
% \textbf{anything else?} \textcolor{blue}{I'm working on some stuffs to show that feature learning may explain the change of slope} 
\end{itemize}
\vspace{-20pt}
\paragraph{Organization.} The rest of the paper is organized as follows.  \Cref{sec:design_principles} outlines problem setup and symmetry constraints through architectures and loss. \Cref{sec:experiments} discusses our experiments, including key hyperparameters and scaling strategies. \Cref{sec:scaling_laws} presents our results and their analysis. 
\Cref{sec:conclusion} concludes with an emphasis on limitations of our work, and important directions for near-future work.  Details on related works, background, experimental setup, and additional results are included in the Appendix.
% {Appendix includes detailed related works in neural scaling laws and existing results on molecular graphs in \cref{sec:related_works}, geometric message passing in \cref{sec:background_mpnn}, experimental setup in \cref{sec:experimental_setup}, uncertainty in scaling laws in \cref{sec:uncertainty}, the effect of scaling vector channels \cref{sec:increase_vector_channels}, results on more diverse dataset \cref{sec:diverse_dataset}, effect of test-time augmentation in \cref{sec:test-time},  ablation study on translation invariance in \cref{sec:translation_invariance}, and instabilities of training vanilla transformers for this tasks in \cref{sec:instabilites_transformer}}.

\vspace{-10pt}
\section{Setup}
\label{sec:design_principles}
An atomistic system can be represented as a point cloud $X=\{(z_1,{x}_1), \dots, (z_n,{x}_n) \}$, where $z_i\in\mathbb{N}$ and ${x}_i\in\mathbb{R}^3$ are the the atomic number and the position respectively. The potential energy  $e(X)$ is a scalar that is invariant to global translations and rotations, while forces ${f}_i(X) \;=\; -\,{\partial {e}(X)} / {\partial {x}_i}$ are vectors that are translation-invariant and rotation-equivariant. The task of our NNIPs is to train a neural network $\phi_\theta: \mathbb{N} \times \mathbb{R}^{3 \times n} \to \mathbb{R}^{1 + 3\times n}$ that takes $X$ as input and predicts the potential energy (scalar) and atom-level forces, one for each atom -- i.e.,  $\phi_\theta: X \mapsto ({e}_\theta(X), \{{f}_{\theta,1}(X), \ldots {f}_{\theta,n}(X)\})$. While it is sufficient to learn the energy for predicting conservative forces, direct force prediction is significantly more scalable. {Using this approach enables one to benefit from the dense signal during pre-training. In post-training, the force prediction can be removed and the model can be fine-tuned to predict conservative forces through backpropagation via the predicted energies, ensuring a good balance between computational cost and accuracy \citep{bigi2025the, fu2025eSEN}.} We minimize the per-atom mean absolute error (MAE) and mean squared error of forces \citep{fu2025eSEN,wood2025uma}:
\begin{equation}
\mathcal{L}(\phi_\theta, X)
= \frac{\lambda_e}{n}\, \big\|  e_\theta(X) -  e(X) \big\|_1
+ \frac{\lambda_f}{n}\, \sum_{i=1}^{n} \big\| {f}_{\theta, i}(X) - f_i(X) \big\|_2,
\label{eq:task_loss}
\end{equation}
with $\lambda_e,\lambda_f >0$ are the coefficients that control the relative importance of energy and force predictions; we use $\lambda_e = \lambda_f$.  %= 10
% This makes the two tasks equally important, although signals from force tasks are stronger since they come from node-level targets.

% \begin{table}[h]
%     \caption{Architectures and their expressivity.}
%     \centering
%     \begin{tabular}{lccc}
%         \toprule 
%          Architectures  & Characteristic & Tensor Order ($\ell$) & Body Order ($\nu$) \\
%          \midrule
%          \midrule
%          Baseline & unconstrained & -  & 2 \\   
%         GemNet-OC & invariant & 0 & 4 \\
%          MC-EGNN & equivariant & 1 & 2 \\
%          eSEN & equivariant & $\ge 2$ & 2 \\
%          \bottomrule
%     \end{tabular}
%     \label{tab:model_expressivity}
% \end{table}

\subsection{Architectures} 
Since we observed instability issues when scaling vanilla transformers for this task, we focused on message-passing architectures. Here, in addition to 
a basic unconstrained MPNN, following the classification of MPNNs in \cite{joshi2023gwl}, we considered three widely adopted equivariant architectures that cover various body  and tensor orders.
The body order corresponds to $S_n$ representations, and refers to the number of nodes participating in a message function. The tensor order $\ell$ corresponding to $SO(3)$ representations, and  denotes the order of the geometric tensor embeddings processed by each model. Below we briefly enumerate these; for more background on these architectures, see \cref{sec:background_mpnn}:
\begin{enumerate}
\item \textbf{\emph{unconstrained}}: a vanilla MPNN that directly processes geometric features, i.e., relative position vectors, without any symmetry constraints.
\item \textbf{\emph{invariant scalars}}: geometric message passing neural network (GemNet-OC) \citep{gemnet_oc} is a variation of GemNet \citep{gemnet} adapted for large and diverse molecular dataset. Although it uses invariants such as interatomic distances and angles, and therefore has a tensor order $\ell=0$, it can approximate equivariant functions from edge-based invariant features, because it performs geometric message passing with two-hop and edge-directional embeddings; see Theorem 3 in \citet{gemnet}. Because GemNet-OC incorporates dihedral-angle information, its two-hop messages depend simultaneously on the states of four nodes, and thus it is classified as four-body. 
\item \textbf{\emph{Cartesian vectors}}:  E(n)-equivariant graph neural network (EGNN) \citep{egnn}; in particular, the extension of \cite{levy2023using}, which allows for more than one equivariant vector channel. We use a specific $\mu$P informed scaling, in which scalar channels scale quadratically wrt number of vector channels, see \cref{sec:background_mpnn} for details. 
\item \textbf{\emph{high-order spherical tensors}}: equivariant Smooth Energy Network (eSEN) \citep{fu2025eSEN}, which uses higher-order irreducible representaions of rotation group ($\ell \ge 2$). Unlike other architectures in the same category \citep[e.g.,][]{thomas2018tensor, nequip, liao2023equiformer}, we found eSEN  more scalable because it uses frame alignment to sparsify the tensor product, allowing it to eliminate Clebsch–Gordan coefficients and to directly parameterize kernels  with linear layers \citep{passaro2023reduce}.
\end{enumerate}

\subsection{Symmetry loss}
Symmetry-based losses have been used in different settings from self-supervised learning \citep{dangovski2021equivariant, bai2025regularization}, to physics-informed settings \citep{tara2023liepoint, yang2024symminformed}, generative modelling \citep{tong2025rao} and symmetry discovery~\citep{escriche2025learning}.
A canonical choice is a loss that penalizes deviations from equivariance constraints for randomly sampled global transformations  \citep[e.g.,][] {kim2023softloss,elhag2025relaxed, bai2025regularization}: 
\begin{equation}
{\mathcal{L}}_{\text{sym}}(\phi_\theta; x,y)
\;=\;
\frac{1}{M}\sum_{i=1}^{M}
\mathcal{L}\!\big(\phi_\theta(\rho_{\text{in}}(g_i)\,x),\;\rho_{\text{out}}(g_i)\,y\big),
\label{eq:symmetry_loss}
\end{equation}
where $g_i \sim \mu_G$ is a sampled from the Haar measure, $\rho_{\text{in}}$ and $\rho_{\text{out}}$ define linear actions on inputs $x$ and targets $y$, respectively, and $\mathcal{L}$ is the task loss. The symmetry-augmented term is added to the base objective in \cref{eq:task_loss} when training $f_\theta$.

For our task, the translation part of the special Euclidean group $SE(3) = SO(3) \ltimes T(3)$ is accounted for by centring the molecule at its center of mass, and the loss is only for the rotation group. \footnote{We also tried using regularization that measures invariance by differentiating along infinitesimal generators, similar to \cite{rhodes2025orb}, but we could not achieve stable training.}

\vspace{-10pt}
\section{Experiments}
\label{sec:experiments}
% \subsection{Setup}
\subsection{Dataset} We conduct our experiments on the OpenMol neutral-molecule subset \citep{levine2025openmol}, with 34M training samples and 27K held-out validation samples.\footnote{We use the neutral subset rather than the full 100M-molecule corpus due to main-memory constraints.} Treating atom nodes as tokens, the training set corresponds to approximately $D \approx 9.2 \times 10^8$ tokens. Following scaling studies in LLMs, we consider a single-epoch training regime, where each sample is observed exactly once. While a multi-epoch setting can be more practical for AI4Science due to smaller datasets compared to language, our goal was to stay faithful to existing methodologies and avoid possible confounding effects \citep{muennighoff2023dataconstraintscaling}.

\subsection{Optimization} Following \citet{choshen2025hitchhiker}, which shows that estimating scaling laws from intermediate checkpoints yields more robust results, we track validation losses throughout training and fit these points—excluding the first $1\%-10\%$ of steps—to a standard scaling-law functional form \citep{kaplan2020scaling, hoffmann2022chinchilla}. A well-known caveat in scaling-law analyses is the sensitivity to learning-rate schedules, particularly when predefined decay steps are used \citep{hoffmann2022chinchilla, hu2024minicpm}. To address this, we adopt scheduler-free AdamW-style optimizers \citep{defazio2024schedulerfree}, which not only remove the need for tuning decay schedules but also allow us to capture model training dynamics within a single run—without retraining from scratch at each data ratio or relying on checkpoint restarts \citep{hu2024minicpm}. Crucially, this approach enables more accurate measurement of training time by mitigating hardware-related artifacts and helps us derive scaling laws directly with respect to training time, measured in GPU-hours.

\subsection{Hyper-parameter Tuning}
Investigating scaling behaviours of neural networks necessitates evaluations under optimal conditions for both efficiency and performance. We, therefore, perform systematic experiments to determine critical hyperparameters affecting the scaling behaviours of those architectures. Our analysis include non-equivariant MPNN, EGNN, and GemNet-OC. Due to the higher computational cost of eSEN, we adopt the optimal hyperparameters from \cite{passaro2023reduce, fu2025eSEN}. 

\paragraph{Learning Rates and Batch Sizes.} When fine-tuning the model, we swept over 12 configurations for each architecture by testing three learning rates, $\{1e-4, 5e-4, 1e-3\}$, and four batch sizes, $\{64,128,256, 512\}$. We performed the tuning with approximately one million parameters. As shown in \cref{fig:batch_size_learning_rate}, smaller batch sizes and larger learning rates resulted in lower validation losses. This finding about small batch sizes aligns with observations in \citep{dimenet_pp, neural_scale_of_chemical_models}.  

\paragraph{Saturation Depth.}
Depth $d$ and width $w$ (embedding dimension) govern the parameter count $N$. While \citet{kaplan2020scaling} found Transformer performance depends mainly on $N$ and is independent of architectural factors such as $d$ and $w$, recent MPNNs work shows that depth choice can undermine power-law scaling behaviour \citep{liu2024graphscaling,sypetkowski2024gnnscaling}. We therefore probe the depth-saturation point on 3D geometric graphs—the depth beyond which validation error no longer improves for fixed capacity \citep{Loukas2020What}. To isolate depth/width from model size, we fix $N\!\approx\!10^6$ and sweep $(d,w)$. As shown in \cref{fig:batch_size_learning_rate}, non-equivariant MPNNs degrade with increasing depth (over-smoothing/over-squashing \citep{topping2022oversquashing}), whereas equivariant models continue to improve, with validation losses plateauing at depths $L\in \{12, \ldots, 16\}$, consistent with prior reports \citep{joshi2023gwl,passaro2023reduce,pengmei2025pushing}.
% Depth $d$ and width $w$ (embedding dimension) are the two key hyperparameters that dominate a network’s parameter count $N$. While \cite{kaplan2020scaling} reported that Transformer performance in language modelling depends primarily on total trainable parameters—largely independent of architectural factors such as $d$ and $w$—recent studies on MPNNs suggest that an arbitrary choice of depth can undermine their power-law scaling behaviour \citep{liu2024graphscaling, sypetkowski2024gnnscaling}. This contrast motivates us to examine the depth-saturation point for MPNNs on 3D geometric graphs, i.e., the critical depth beyond which validation error no longer improves, given a fixed capacity $w \times d$ \citep{Loukas2020What}. To isolate depth and width effects from model capacity, we fixed $N \approx$ $1$M, and varied $(d, w)$, thereby eliminating model-size confounds. As shown in \cref{fig:batch_size_learning_rate}, non-equivariant MPNNs degrade with increasing depth, consistent with over-smoothing and over-squashing \citep{topping2022oversquashing}. In contrast, deeper equivariant architectures continue to improve, with validation losses plateauing around $L \in \{12, \ldots, 16\}$. Notably, this trend for deeper equivariant models aligns with prior observations \citep{joshi2023gwl, passaro2023reduce, pengmei2025pushing}. 

\paragraph{Infinite-Width Scaling.} Our depth-saturation experiments are motivated by the universality condition for message-passing-based architectures, which requires “sufficient depth” and unbounded width; see Corollary 3.1 in \cite{Loukas2020What}. For each architecture type, we train a series of models with an increasing number of channels (width) along a scaling ladder. We fix the optimal hyperparameters—including depth, learning rate $\eta^* = 1\text{e}{-}3$, and batch size—for $\approx$1M-parameter models with base width $w_{\text{base}}$, and use $\mu$P \citep{yang2021muP} to transfer $\eta^*$ to other widths $w$ via $\eta(w) = \eta^* \cdot \frac{w_{\text{base}}}{w}$. We increase model size until the memory of a single NVIDIA A100 (40 GB) GPU is exhausted. In other words, we keep depth and batch size constant across model sizes and scale the width as high as our hardware allows.
\vspace{-10pt}
\section{Scaling Laws}
\label{sec:scaling_laws}
\subsection{Scaling Compute} 
Nominal FLOPs are hardware-agnostic, yet equivariant models often have lower GPU utilization, so FLOPs may understate practical cost. We therefore fit scaling laws in both theoretical FLOPs $C$ and wall-clock training hours $H$, training all models on identical hardware; see \Cref{sec:experimental_setup}. 

\begin{figure}[ht]
    \centering
      \begin{subfigure}[b]{0.24\textwidth} 
        \centering
        \includegraphics[width=\linewidth]{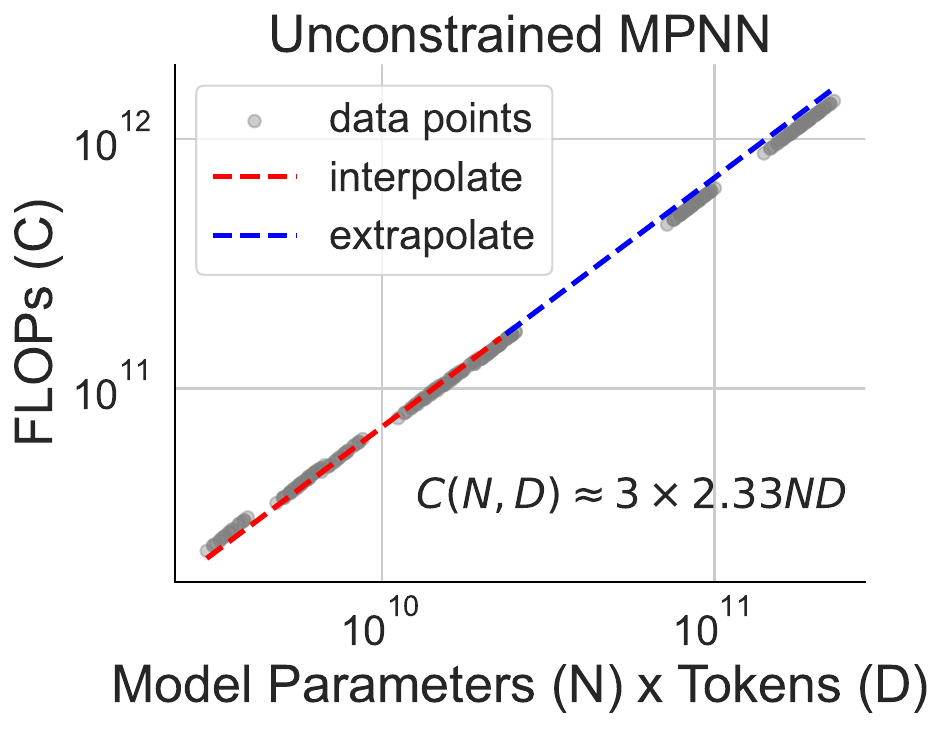} 
        \label{fig:subfigA}
    \end{subfigure}
    \begin{subfigure}[b]{0.24\textwidth}
        \centering
        \includegraphics[width=\linewidth]{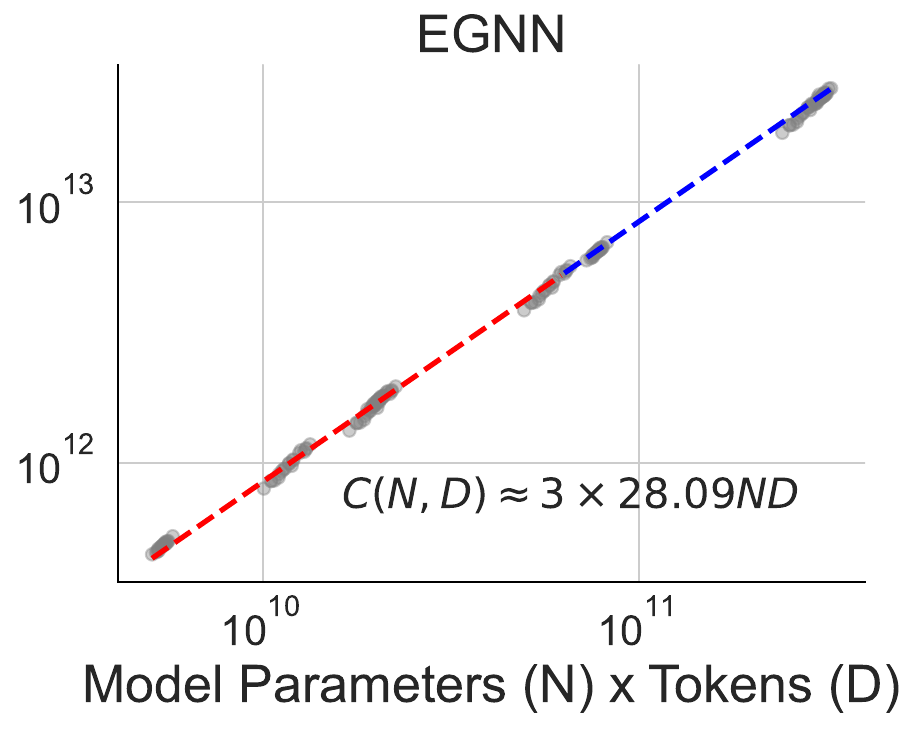}
        \label{fig:subfigB}
    \end{subfigure} 
    \begin{subfigure}[b]{0.24\textwidth}
        \centering
        \includegraphics[width=\linewidth]{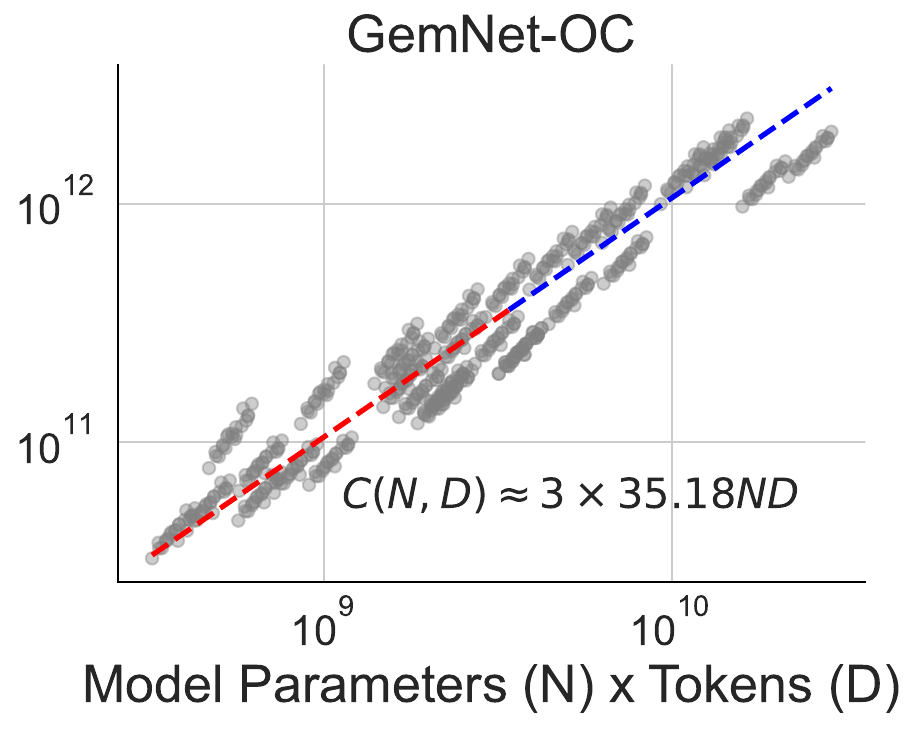}
        \label{fig:subfigB}
    \end{subfigure} 
    \begin{subfigure}[b]{0.24\textwidth}
        \centering
        \includegraphics[width=\linewidth]{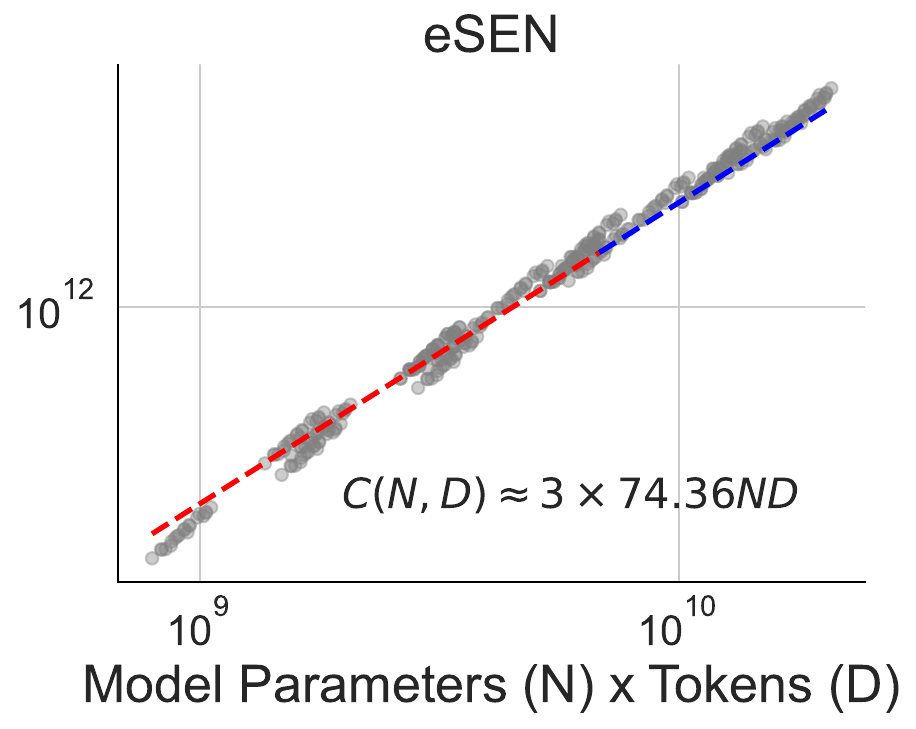}
        \label{fig:subfigB}
    \end{subfigure} 
    \vspace{-10pt}
    \caption{Estimation of $\kappa$ for architectures used in our study. }
    \label{fig:flops_used}
\end{figure}
\begin{figure}[h]
    \centering
      \begin{subfigure}[b]{0.24\textwidth} 
        \centering
        \includegraphics[width=\linewidth]{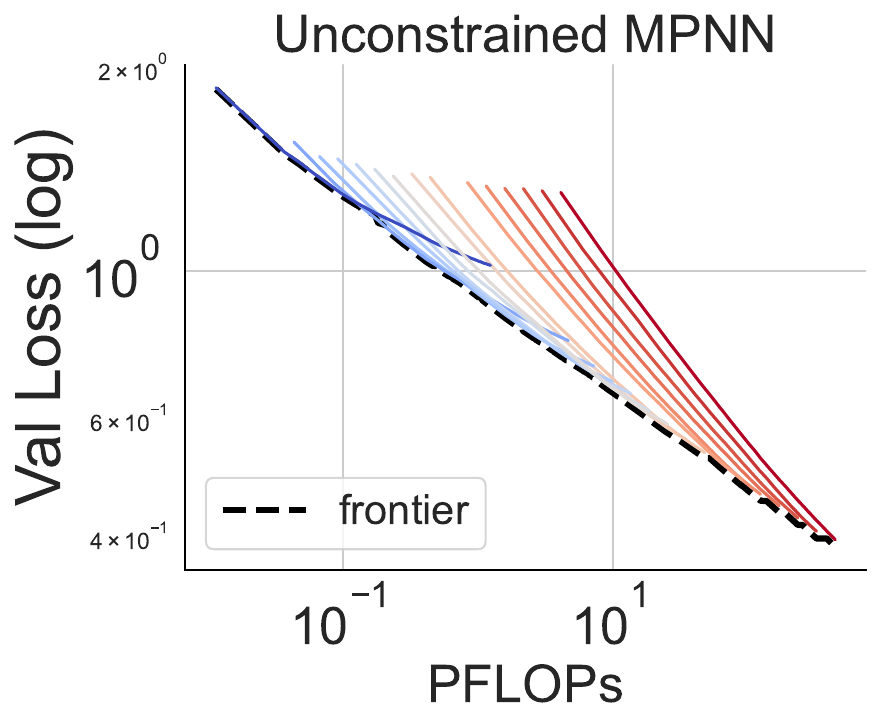}
        \label{fig:subfigA}
    \end{subfigure}
    \begin{subfigure}[b]{0.24\textwidth}
        \centering
        \includegraphics[width=\linewidth]{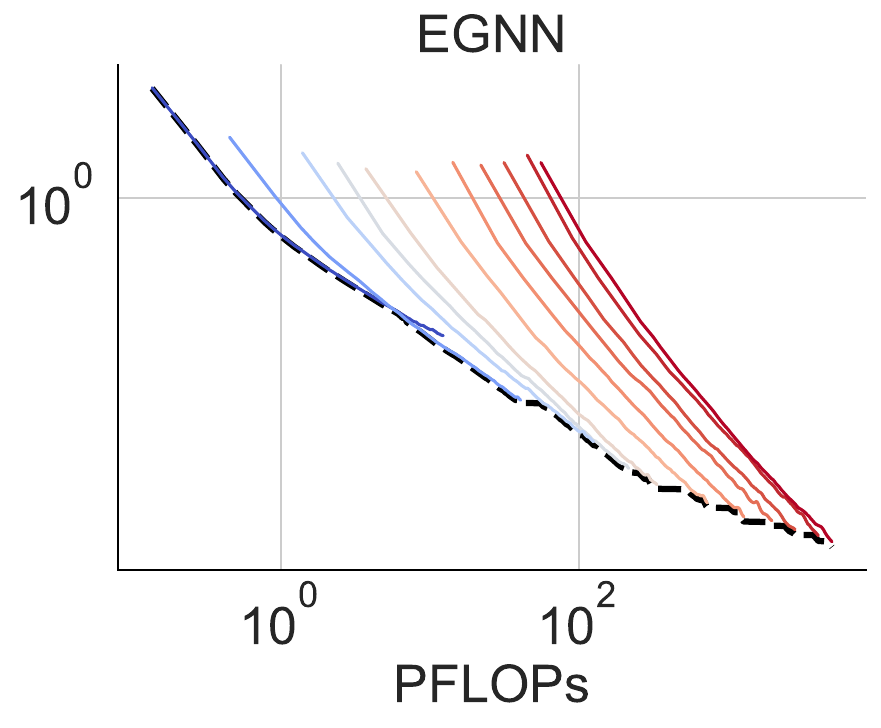}
        \label{fig:subfigB}
    \end{subfigure} 
       \begin{subfigure}[b]{0.24\textwidth}
        \centering
        \includegraphics[width=\linewidth]{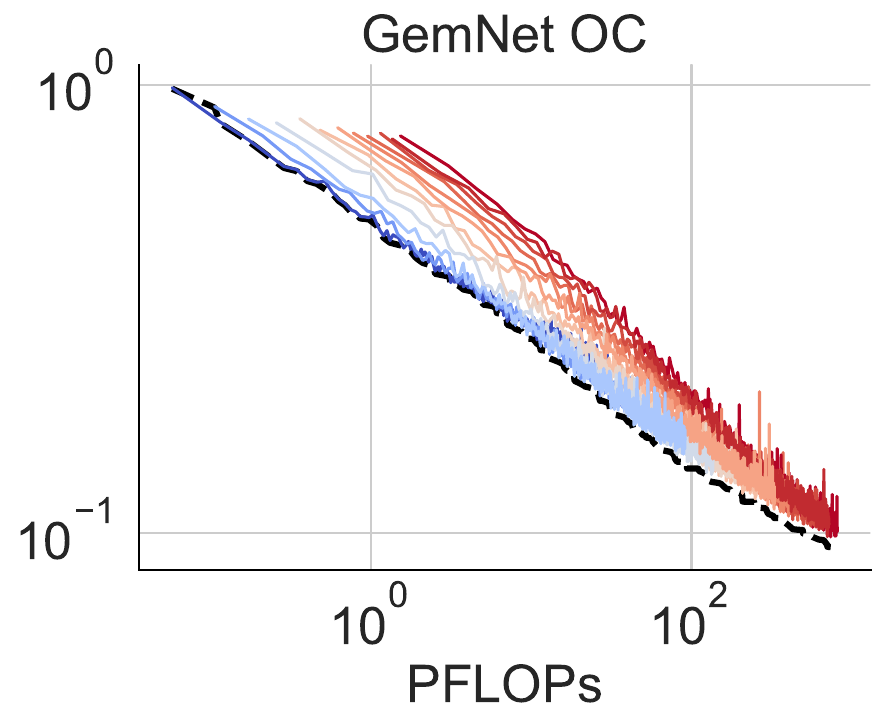}
        \label{fig:subfigB}
    \end{subfigure} 
    \begin{subfigure}[b]{0.24\textwidth}
        \centering
        \includegraphics[width=\linewidth]{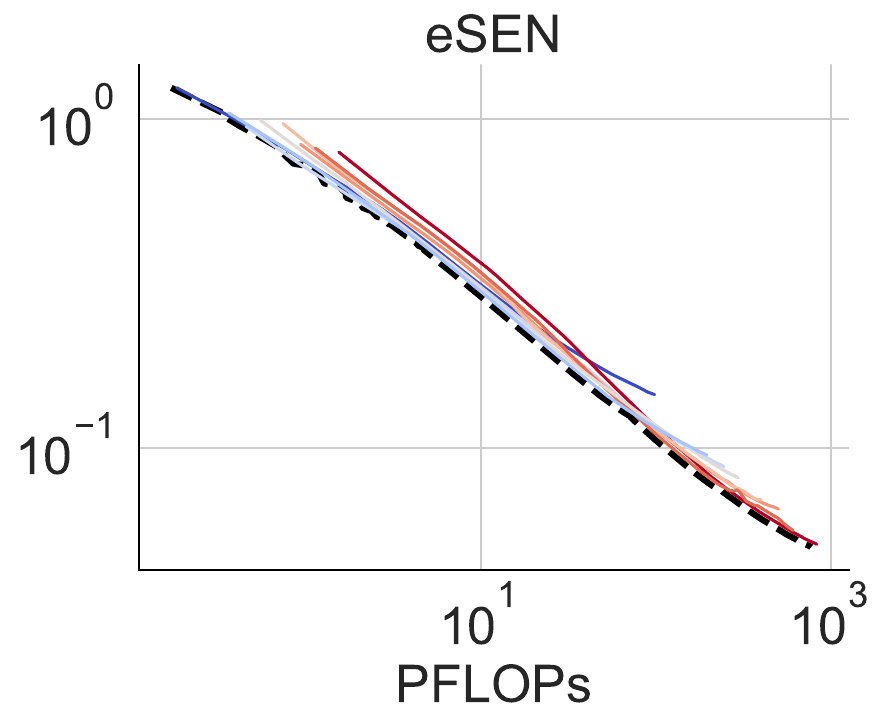}
        \label{fig:subfigB}
    \end{subfigure} 
    \begin{subfigure}[b]{0.24\textwidth} 
        \centering
        \includegraphics[width=\linewidth]{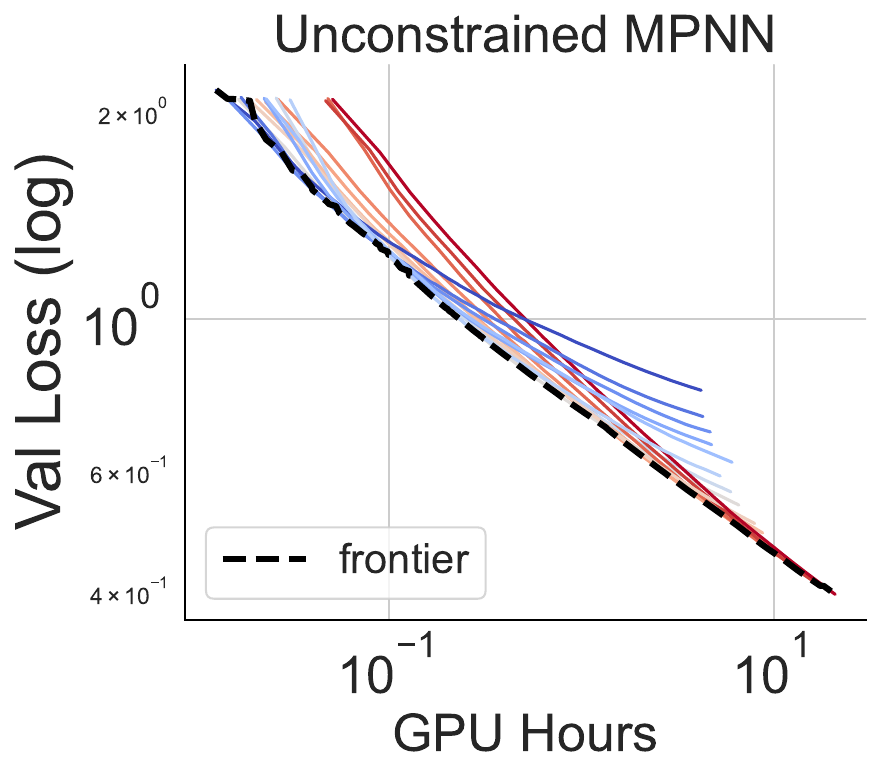} 
        \label{fig:subfigA}
    \end{subfigure}
    \begin{subfigure}[b]{0.24\textwidth}
        \centering
        \includegraphics[width=\linewidth]{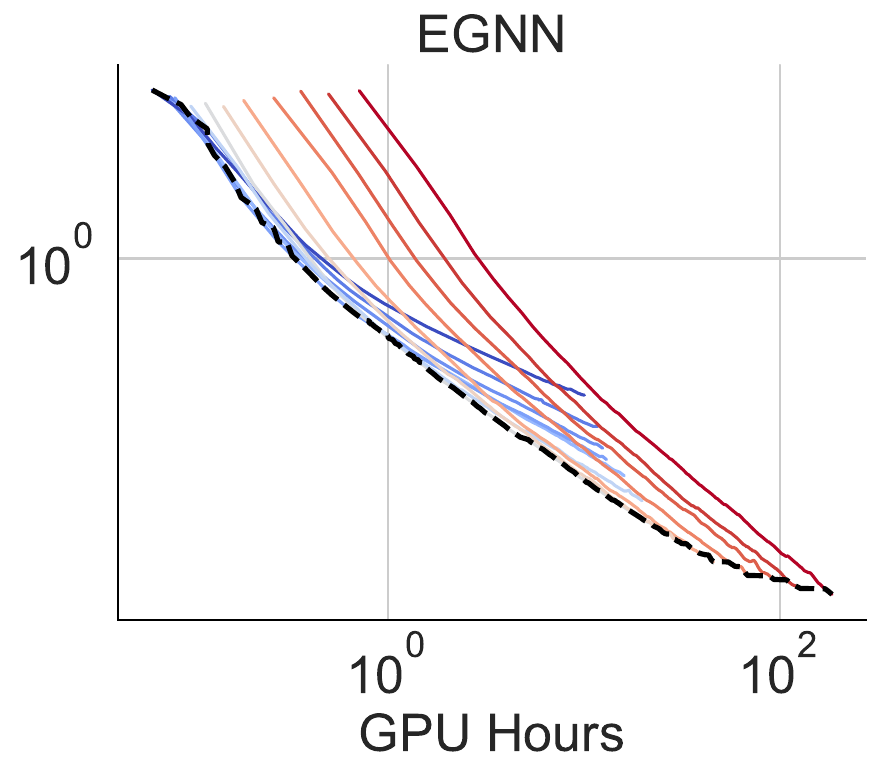}
        % \caption{MC-EGNN}
        \label{fig:subfigB}
    \end{subfigure} 
        \begin{subfigure}[b]{0.24\textwidth}
        \centering
        \includegraphics[width=\linewidth]{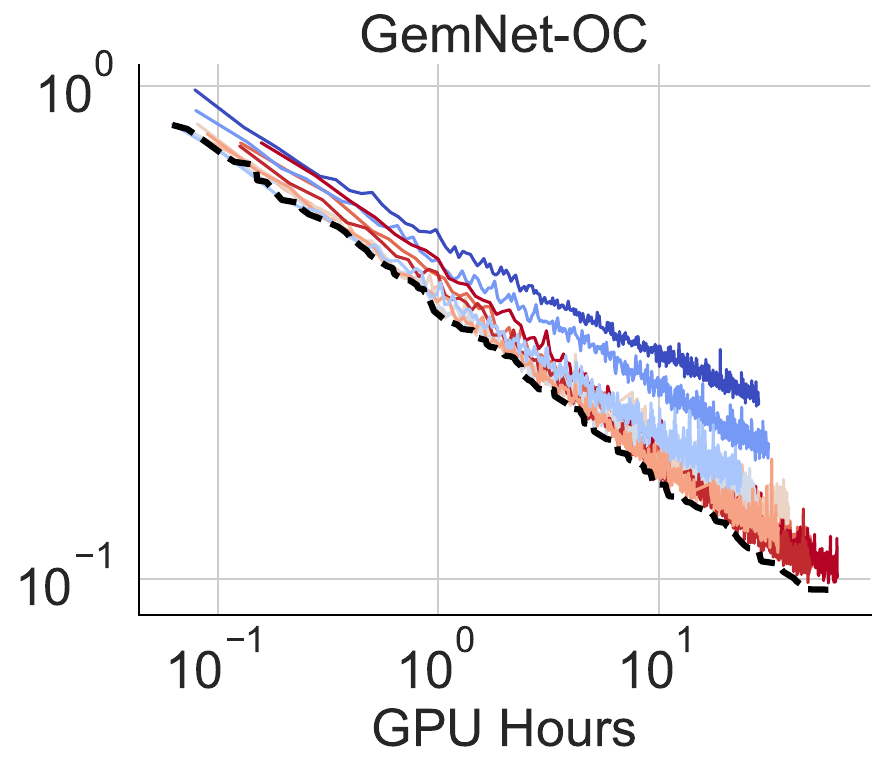}
        % \caption{MC-EGNN}
        \label{fig:subfigB}
    \end{subfigure} 
    \begin{subfigure}[b]{0.24\textwidth}
        \centering
        \includegraphics[width=\linewidth]{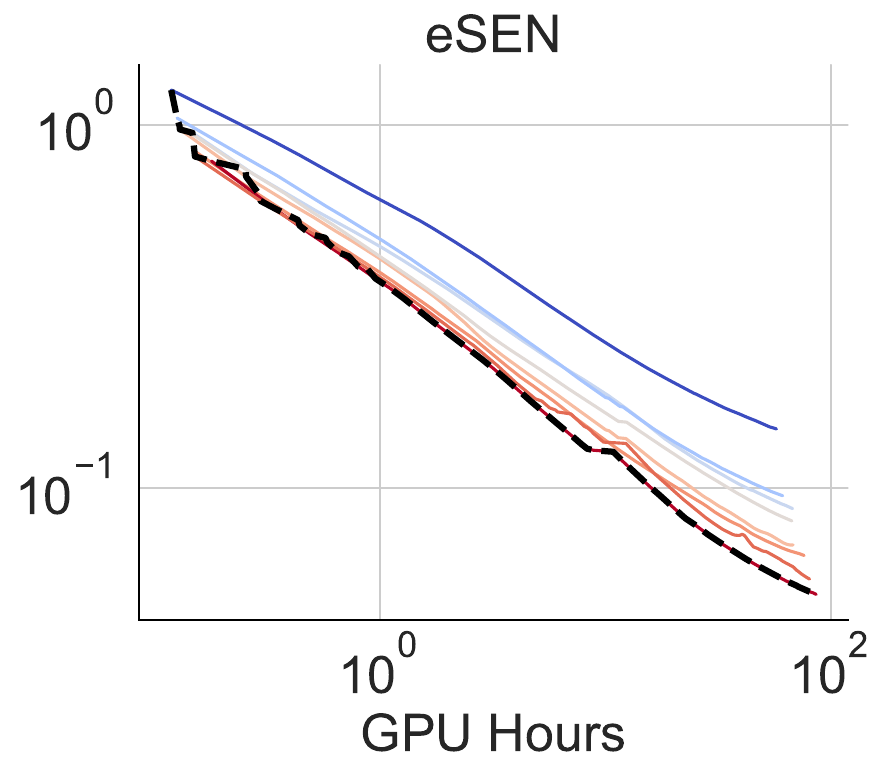}
        \label{fig:subfigB}
    \end{subfigure} 
    \vspace{-10pt}
    \caption{Pareto frontiers of training compute in log–log spaces. \textbf{Top}: Efficient loss-FLOPs frontier. \textbf{Bottom}: Efficient loss-train-time frontier. Across architectures, the log–log frontiers are approximately linear. Line color encodes model size (\textcolor{blue}{small}, \textcolor{red}{large}).}
    \vspace{-2em}
    \label{fig:pareto_front_flops}
\end{figure}
\label{sec:scaling_compute}
\paragraph{Counting FLOPs.} Following \cite{kaplan2020scaling, hoffmann2022chinchilla}, we define the compute as theoretical FLOPs counting $C$ incurred from training a model of $N$ parameters on $D$ training tokens: 
\begin{equation}
    C \approx 3 \times \kappa \times N \times D.
    \label{eq:flops_counting}
\end{equation}
Here, $\kappa$ is an architecture-dependent constant representing the number of FLOPs required for a single forward pass over one input token. During training, each input incurs both a forward and a backward pass, with the latter approximately doubling the FLOP cost. Consequently, the total training cost per token is $3\kappa$. For architectures dominated by dense linear layers, $\kappa \approx 2$, yielding the widely used expression $C \approx 6ND$ for transformer-based language models.

\begin{wrapfigure}{r}{0.3\textwidth}
  \centering
  \includegraphics[width=\linewidth]{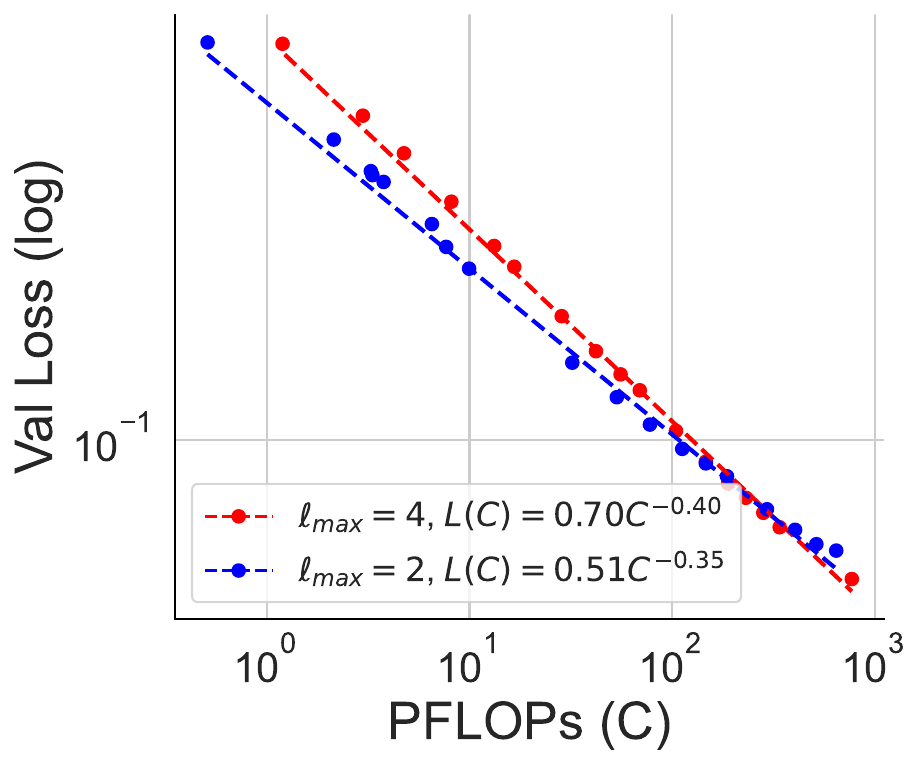}
  \caption{Using higher orders of feature tensors in eSEN leads to better scaling exponents w.r.t compute.}
  \label{fig:esen_different_order}
  \vspace{-20pt}
\end{wrapfigure}

\paragraph{Estimating $\kappa$.} We empirically estimate $\kappa$ by varying $N$ and $D$, recording FLOPs for a pass over $D$, and fitting $C$ vs.\ $ND$. \Cref{fig:flops_used} shows a clear linear trend with distinct $\kappa$ per architecture: unconstrained MPNNs with mostly linear layers  give $\kappa \approx 2$, whereas equivariant architectures incur larger $\kappa$. % \citep{brehmer2025does, wood2025uma}.
\paragraph{Compute-Optimal Frontier.} For compute scaling, we follow approach 1 in \cite{hoffmann2022chinchilla}. For each compute budget, we select the minimum validation loss achieved across runs, yielding the loss–compute Pareto frontiers. As shown in \Cref{fig:pareto_front_flops} \footnote{GemNet-OC exhibits noisier learning curves because it relies on empirically estimated per-layer scaling factors—approximated from a few random batches-rather than explicit normalization (e.g., LayerNorm) to control activation variance.}, loss-compute frontiers across architectures follows in linear relationships in log-log space. We then fit these frontier points to the power law:
\begin{equation}
    L(C)=L_\infty + F_c\,C^{-\gamma_c},  \quad  L(H) = L_\infty  + F_hH^{-\gamma_h},
    \label{eq:scaling_flops}
\end{equation}

where $L_\infty$ is the irreducible loss for the given architecture and dataset, and $F_C$,$\gamma_c$, $F_h$, and $\gamma_h$ are fit parameters. Unlike language modeling with cross-entropy, force-field tasks do not admit a clear theoretical baseline for $L_\infty$ \citep{brehmer2025does, wood2025uma}; therefore, we set $L_\infty \approx 0$ unless noted otherwise. This choice of $L_\infty$ gives exponents that are consistent with the alternative derivation in \cref{sec:compute_optimal_scaling}. \Cref{fig:champion} summarizes our main results, which indicate different exponents for architectures with increasing levels of symmetry expressivity. We also find the argument holds within the same architecture; particularly, \cref{fig:esen_different_order} shows an improvement in compute-scaling exponents as the max order $\ell_{\text{max}}$ increases from $2$ to $4$ in eSEN. {Finally, it is worth noting that we do not study the effects of denoising pretraining, as done for Orb, a non-equivariant model, by \cite{neumann2024orbfastscalableneural}. Consequently, our compute-scaling results are not directly comparable to this line of work, since their compute budget is allocated differently between pretraining and downstream fine-tuning.}
\begin{figure}[h]
    \centering
      \begin{subfigure}[b]{0.24\textwidth} 
        \centering
        \includegraphics[width=\linewidth]{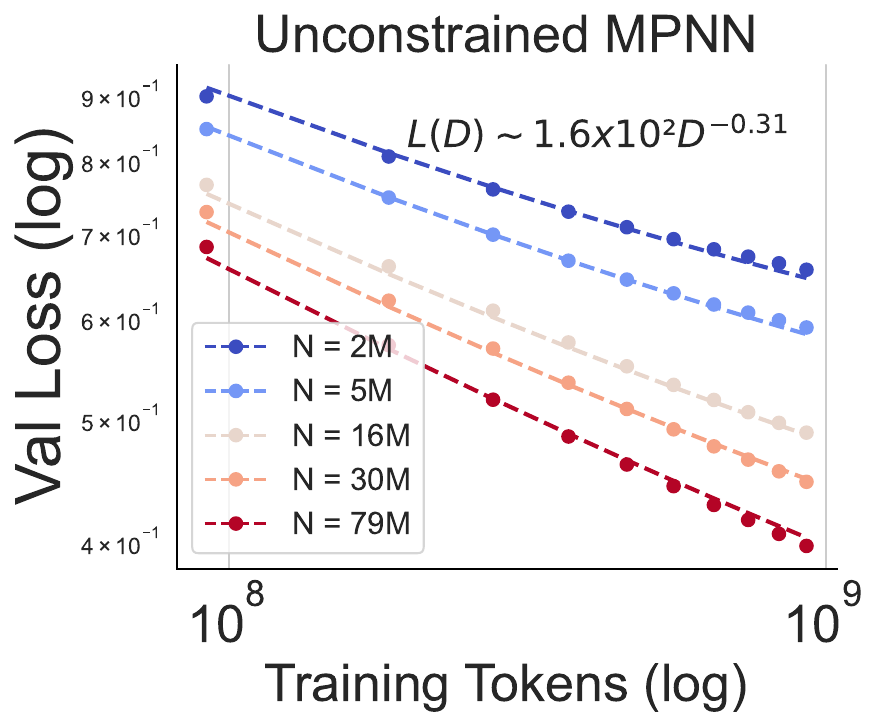}
        \label{fig:subfigA}
    \end{subfigure}
    \begin{subfigure}[b]{0.24\textwidth}
        \centering
        \includegraphics[width=\linewidth]{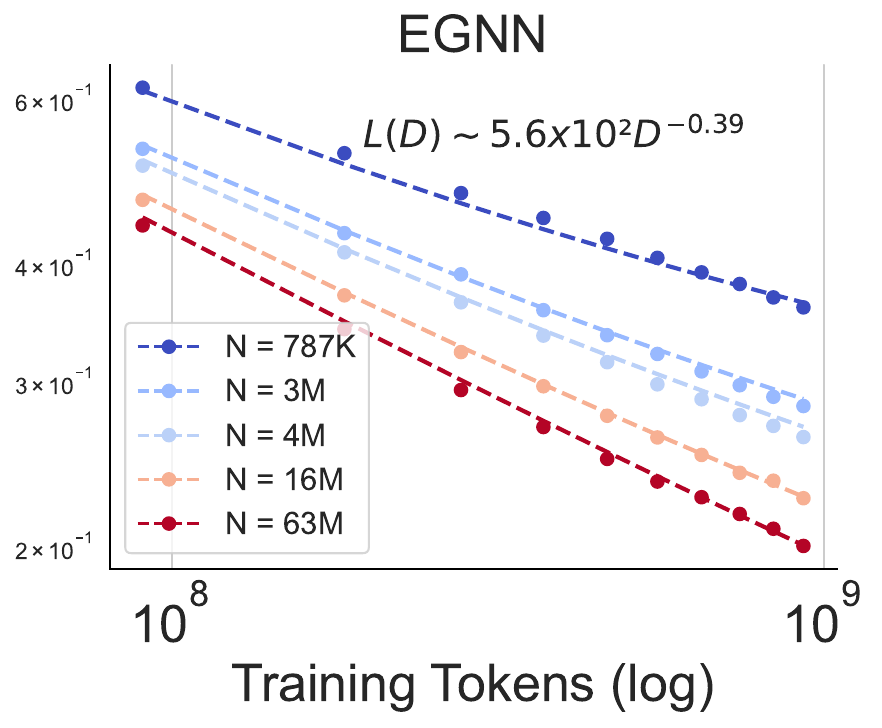}
        \label{fig:subfigB}
    \end{subfigure} 
       \begin{subfigure}[b]{0.24\textwidth}
        \centering
        \includegraphics[width=\linewidth]{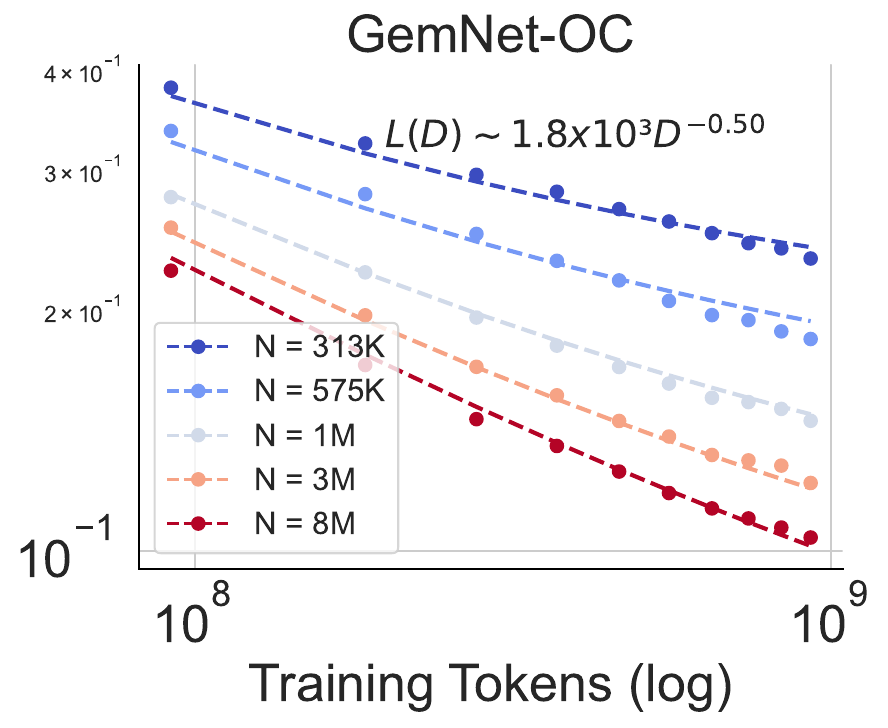}
        \label{fig:subfigB}
    \end{subfigure} 
    \begin{subfigure}[b]{0.24\textwidth}
        \centering
        \includegraphics[width=\linewidth]{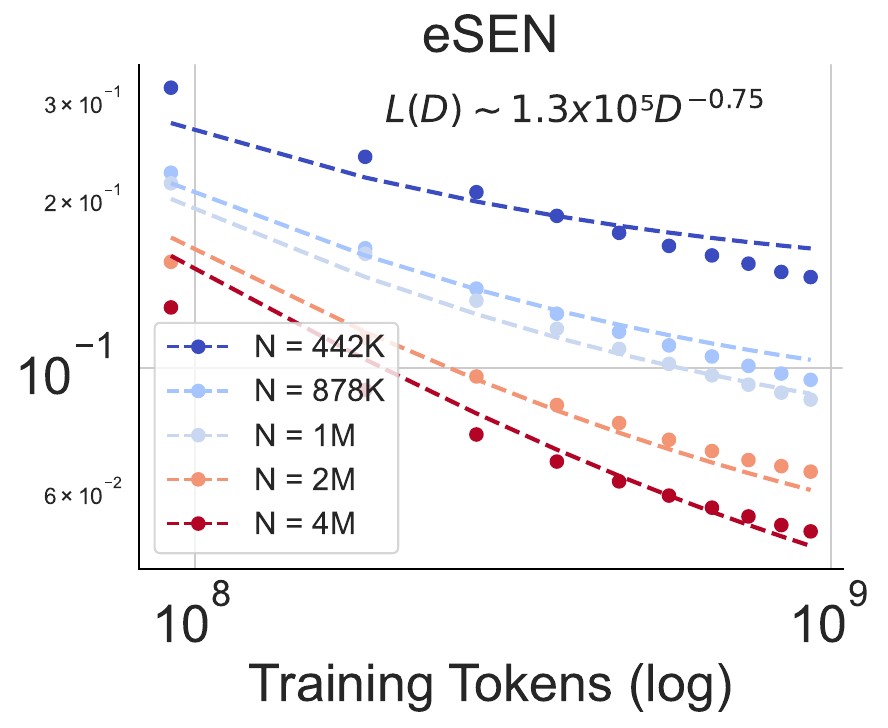}
        \label{fig:subfigB}
    \end{subfigure} 
    \begin{subfigure}[b]{0.24\textwidth} 
        \centering
        \includegraphics[width=\linewidth]{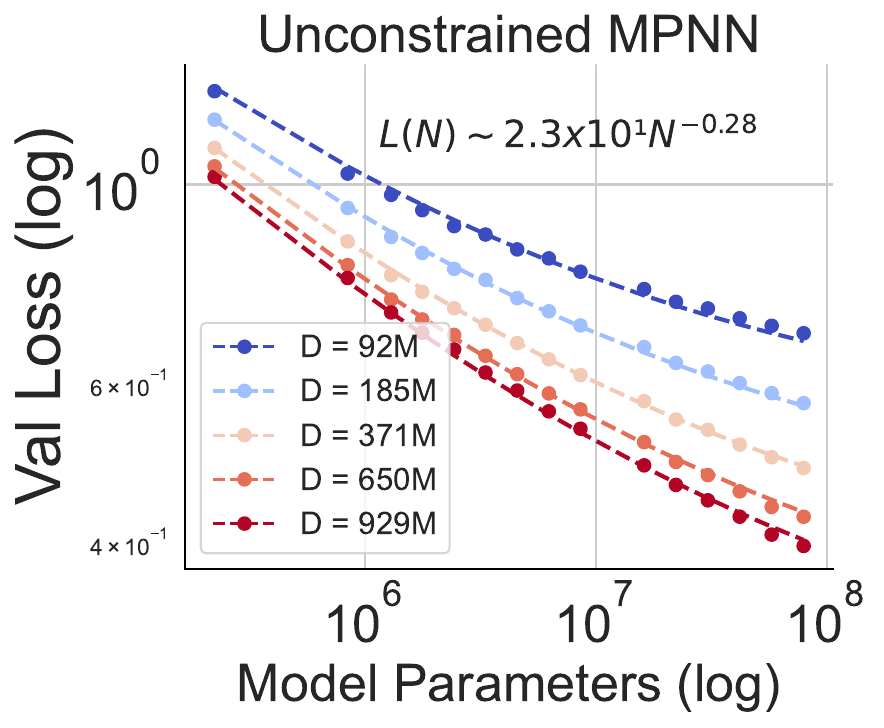}
        \label{fig:subfigA}
    \end{subfigure}
    \begin{subfigure}[b]{0.24\textwidth}
        \centering
        \includegraphics[width=\linewidth]{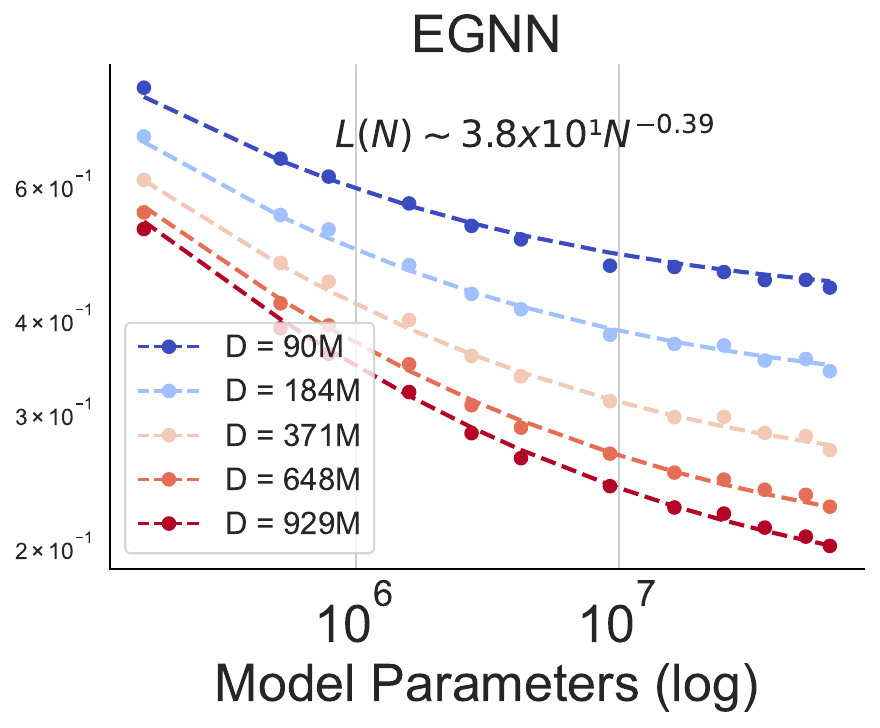}
        \label{fig:subfigB}
    \end{subfigure} 
       \begin{subfigure}[b]{0.24\textwidth}
        \centering
        \includegraphics[width=\linewidth]{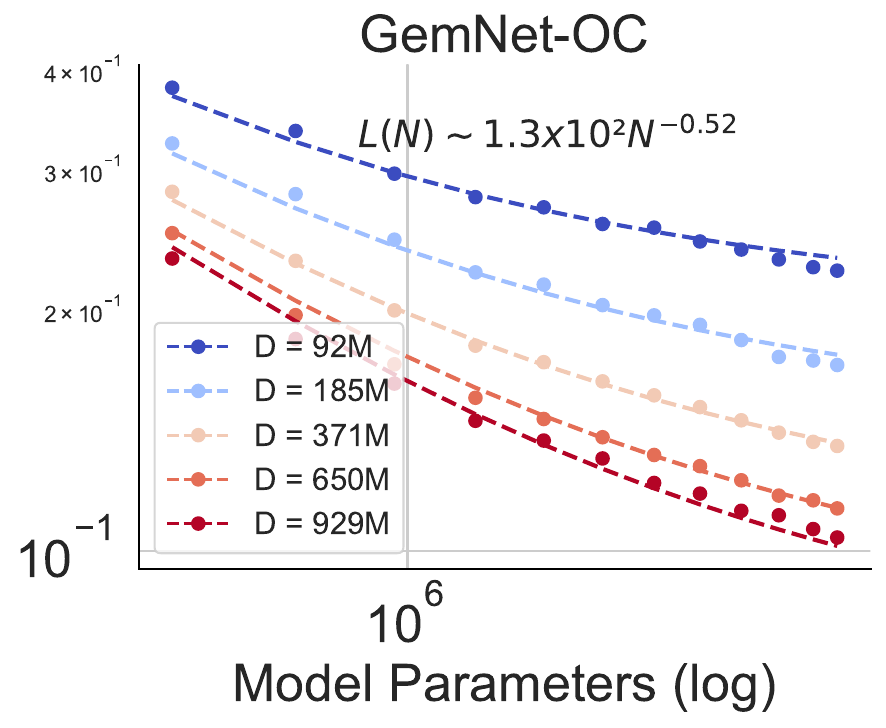}
        \label{fig:subfigB}
    \end{subfigure} 
    \begin{subfigure}[b]{0.24\textwidth}
        \centering
        \includegraphics[width=\linewidth]{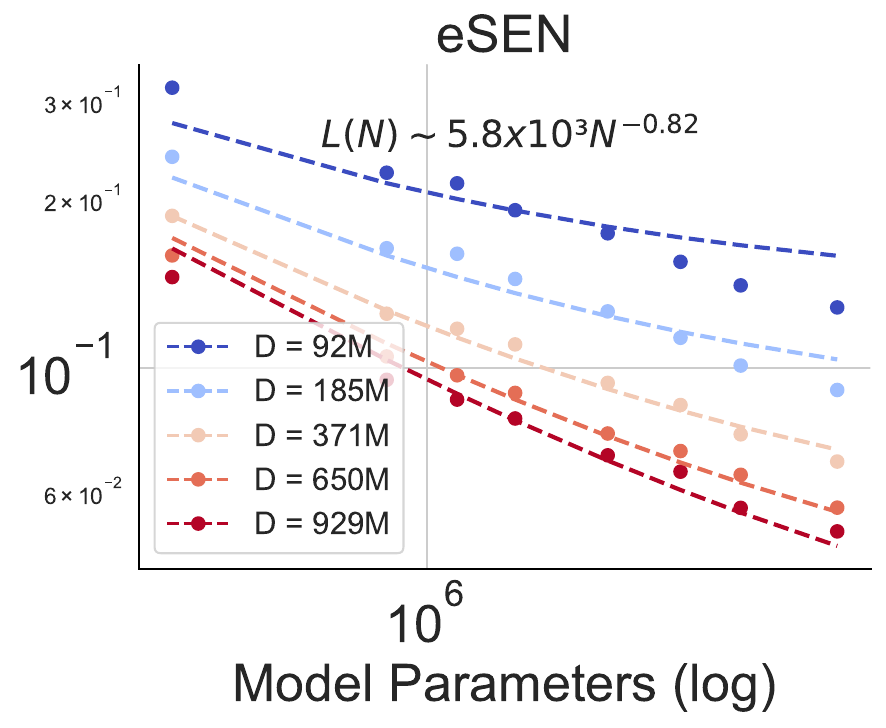}
        \label{fig:subfigB}
    \end{subfigure} 
    \vspace{-10pt}
    \caption{\textbf{Top:} Scaling number of training tokens. \textbf{Bottom:}  Scaling number of parameters}
    \vspace*{-2em}
     \label{fig:scaling_data_size}
\end{figure}
% \vspace{-10pt}
\subsection{Scaling Parameter and Dataset}
\label{sec:scaling_data_model}
\paragraph{Sum-Power-Law.} To analyze scaling in model size $N$ and dataset size $D$, we follow approach 3 of \citep{hoffmann2022chinchilla} and fit the triplets $(N, D, L)$ to the separable power-law model as:
\begin{equation}
    L(N, D) = L_\infty + A \times N ^{-\alpha} + B \times D^{-\beta}.
    \label{eq:scaling_law_N_D}
\end{equation}
$L(N,D)$ is the validation loss represented as a function of $(N, D)$. $L_\infty$ $A, B, \alpha, \beta$ are parameters that we fit. Notably, we found $L_\infty$ to be $\approx 0$ in all architecture families. For each $N$, we measured validation loss at training set fractions $r \in \{0.1, 0.2,\dots, 1.0\}$, that is $D_r = r \cdot D_{\text{max}}$ \footnote{Because the GemNet-OC loss curve is high-variance, we smooth it using an exponential moving average with a smoothing factor of $0.9$.}. 

\paragraph{Scaling Analysis.} \Cref{fig:scaling_data_size} presents our fit for four architectures under study. The top row shows the power-law fit in validation loss when the number of training tokens, $D$, is the limiting factor. Power-law exponents $\beta$, are $.31, .39, .50$ and $.75$ from left to right. The bottom row shows this relationship when the number of model parameters, $N$, is the bottleneck. Here, the exponent $\alpha$ from left to right is $.28, .39, .52$, and $.82$. These results highlight three phenomena:
\begin{itemize}
    \item \textbf{Data Efficiency}: In data-limited scenarios, equivariant models demonstrate superior scaling behaviours compared to unconstrained models, demonstrated by their larger scaling exponents  $\beta$. 
    % This is because equivariant models are more data-efficient and learn more effectively from data that exhibits symmetry. 
    Moreover, equivariance of higher orders translates to larger exponents. 
    \item \textbf{Expressivity}: When bottlenecked by model size, equivariant models exhibit higher scaling exponents with respect to $N$. This occurs because explicit symmetry constraints enable greater expressivity with fewer parameters. 
    % In contrast, unconstrained architectures require a significantly larger number of parameters to achieve comparable performance. 
    Furthermore, the scaling exponent gap between high-order architectures (i.e., eSEN, GemNet-OC) and lower-order ones (i.e., EGNN) is considerable. While higher order representations are known to result in better expressivity \citep{joshi2023gwl}, 
    the fact that the benefit of such representations grows with scale is a novel finding.
    \item {${\alpha \approx \beta}$:} The exponents remain close across architectures. We discuss this finding in \cref{sec:compute_optimal_scaling}. 
\end{itemize}

In brief, we observe larger data-scaling exponents $\beta$ for equivariant networks, consistent with prior reports \citep{nequip,brehmer2025does,wood2025uma}. Meanwhile, our parameter-scaling exponents $\alpha$ are larger for equivariant networks, and this differs from \citep{brehmer2025does}, which report larger $\alpha$ for unconstrained models; note that the tasks are not directly comparable. Together, the increases in both $\beta$ and $\alpha$ for equivariant models change the slope of the compute-optimal frontier under $C \propto ND$, which is one of our main findings.

\subsection{Compute-Optimal Allocation}
\label{sec:compute_optimal_scaling}
\begin{figure}[h]
    \centering
    % Left: the two plots
    \begin{minipage}[t]{0.62\textwidth}
        \centering
        \begin{subfigure}[t]{0.48\textwidth}
            \includegraphics[width=\linewidth]{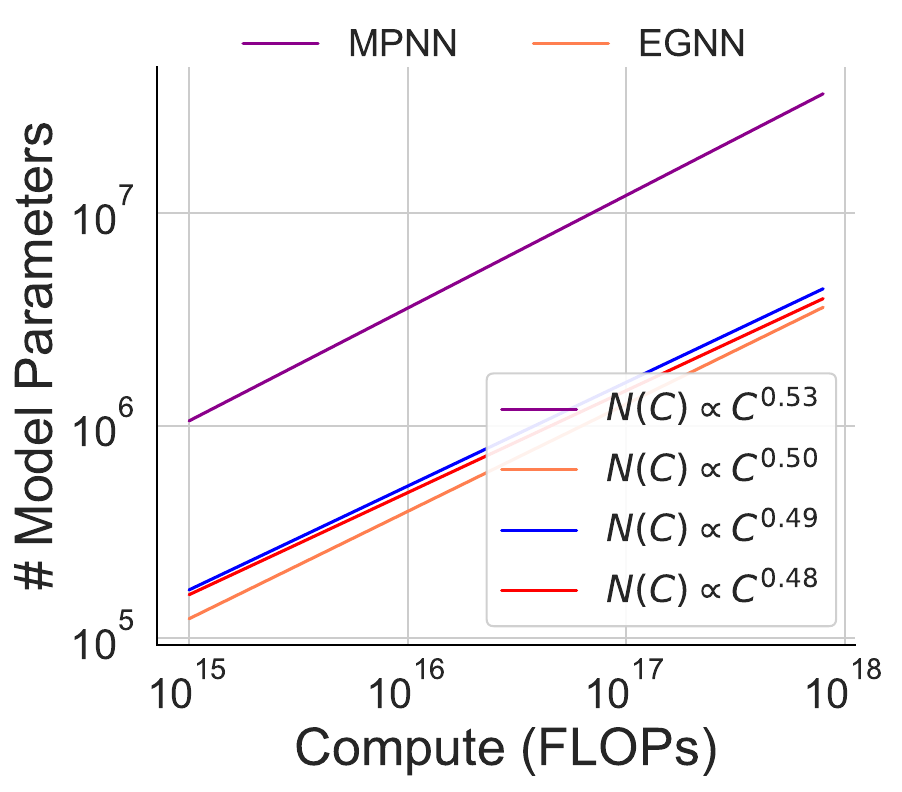}
            \label{fig:subfigA}
        \end{subfigure}\hfill
        \begin{subfigure}[t]{0.48\textwidth}
            \includegraphics[width=\linewidth]{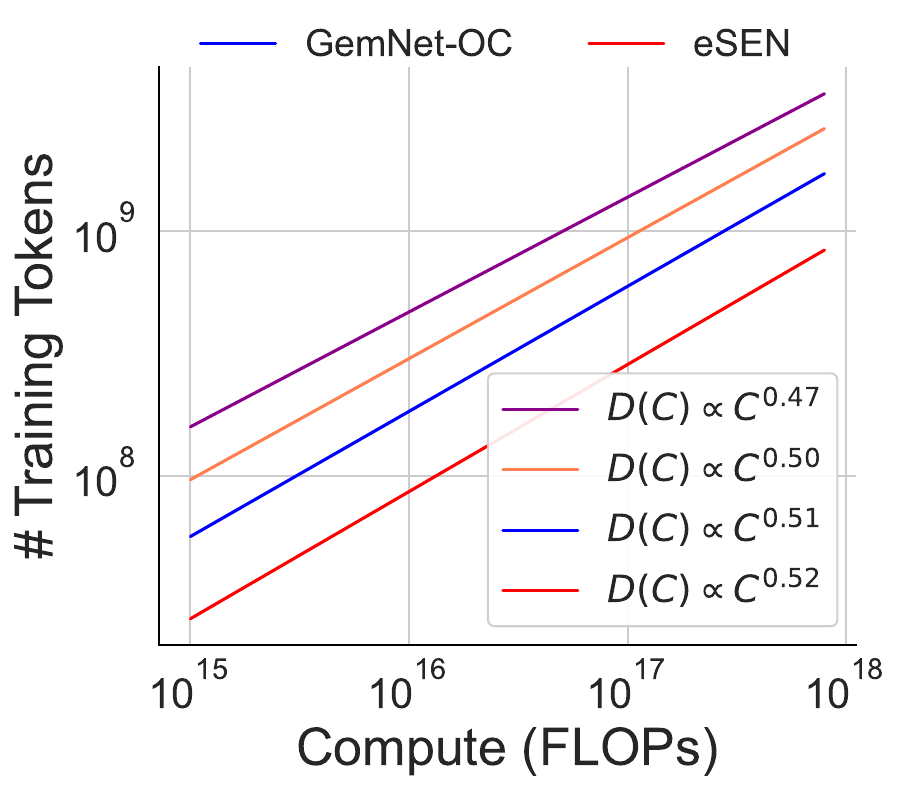}
            \label{fig:subfigB}
        \end{subfigure}
    \end{minipage}\hfill
    % Right: caption, top-aligned with NO top skip
    \begin{minipage}[t]{0.35\textwidth}
        \vspace*{-8em} % remove the caption's top spacing
        \label{fig:optimal_allocation}
         \caption{{Optimal model size (left) and training size (right) at a fixed level of compute (log-log)}}
    \end{minipage}
\end{figure}

We have presented two scaling laws so far: (1) a power law with respect to the \textit{compute-optimal frontier} in \cref{sec:scaling_compute}, and (2) a \textit{sum-power-law} with respect to parameter count and the number of training tokens in \cref{sec:scaling_data_model}. In this section, we discuss the connection between them.
Given a fixed compute budget $C$ (FLOPs), we seek the optimal allocation between model size $N$ and training tokens $D$. We pose this as a constrained optimization problem that combines \cref{eq:scaling_law_N_D} with \cref{eq:flops_counting}; in particular, we have:
\begin{equation}
    N^*(C), D^*(C) = \text{argmin } L(N, D), \quad  3\kappa ND = C. 
    \label{eq:compute_optimal_allocation}
\end{equation}
Recall that $L(N,D)=L_\infty + A N^{-\alpha} + B D^{-\beta}$, with $L_\infty \approx 0$. Let $N^*(C)$ and $D^*(C)$ denote, respectively, the compute-optimal model size and data size for a fixed compute budget $C$. Solving \cref{eq:compute_optimal_allocation} yields $N^*(C) = {G}\xi^{-a} C^a, \quad D^*(C) = G^{-1} \xi^{-b}C^b,\label{eq:opt_N_and_D}$
where $\xi = 3\kappa$, $G = (\frac{\alpha A}{\beta B})^\frac{1}{\alpha + \beta}$, $a = \frac{\beta}{\alpha + \beta}$, $b = \frac{\alpha}{\alpha + \beta}$ \citep{hoffmann2022chinchilla, brehmer2025does}. 
Furthermore, plugging back the results to $L(N, D)$, we get back the loss-compute frontier power law similar to \cref{eq:scaling_flops}: 
\begin{equation}
    L(C) = L(N^*(C), D^*(C)) = F_c C^{-\gamma_c},
    \label{eq:sum_power_law}
\end{equation}
where $F_c = AG^{-\alpha}\xi^\gamma + BG^\beta \xi^\gamma$, and $\gamma_c = \frac{\alpha \beta}{\alpha + \beta}$.
\Cref{tab:consistency_law} presents the values of $F_c$ and $\gamma_c$ obtained from two methods. The results show a good agreement between them, indicating the consistency of our power laws. We further visualize the compute-optimal allocation between model size $N$ and data size $D$ in \cref{fig:optimal_allocation}. Across architectures, we find $a \approx b \approx 0.5$, indicating that parameters and tokens should be scaled in roughly equal proportions, consistent with the Chinchilla allocation for transformer language modelling \citep{hoffmann2022chinchilla}.

\begin{table}[htbp]
\centering
\caption{Compute-optimal scaling law parameters with $95\%$ confidence intervals; see \cref{sec:uncertainty}. Compute is scaled to PFLOPs.}
\label{tab:consistency_law}

\begin{minipage}{0.82\linewidth}
\centering
\small
\setlength{\tabcolsep}{6pt}
\renewcommand{\arraystretch}{1.15}
\newcolumntype{C}{>{\centering\arraybackslash}X}

\begin{tabularx}{\linewidth}{@{} l c CC @{}}
\toprule
\bfseries Architecture & \bfseries Param & \multicolumn{2}{c}{\bfseries Fit Method} \\
\cmidrule(lr){3-4}
 &  & \makecell{Compute-Optimal Frontier\\ \cref{eq:scaling_flops}} & \makecell{Sum-Power-Law\\ \cref{eq:sum_power_law}} \\
\midrule
\multirow{2}{*}{MPNN} & $F_c$        & 0.928 [0.925–0.930] & 0.934 [0.863–0.952] \\
                                    & $\gamma_c$   & 0.142 [0.141–0.143] & 0.146 [0.142–0.159] \\
\addlinespace[2pt]
\multirow{2}{*}{MC-EGNN}            & $F_c$        & 0.775 [0.761–0.792] & 0.811 [0.784–0.832] \\
                                    & $\gamma_c$   & 0.173 [0.169–0.178] & 0.195 [0.188–0.204] \\
\addlinespace[2pt]
\multirow{2}{*}{GemNet-OC}          & $F_c$        & 0.488 [0.485–0.491] & 0.479 [0.424–0.542] \\
                                    & $\gamma_c$   & 0.255 [0.252–0.257] & 0.256 [0.232–0.282] \\
\addlinespace[2pt]
\multirow{2}{*}{eSEN}               & $F_c$        & 0.703 [0.696–0.712] & 0.669 [0.575–0.729] \\
                                    & $\gamma_c$   & 0.403 [0.401–0.406] & 0.392 [0.342–0.451] \\
\bottomrule
\end{tabularx}
\end{minipage}
\end{table}

% \begin{figure}[h]
%     \centering
%       \begin{subfigure}[b]{0.30\textwidth} 
%         \centering
%         \includegraphics[width=\linewidth]{Plots/optimal_allocation/compute_vs_model_size.pdf} 
%         \label{fig:subfigA}
%     \end{subfigure}
%     \begin{subfigure}[b]{0.30\textwidth}
%         \centering
%         \includegraphics[width=\linewidth]{Plots/optimal_allocation/compute_vs_training_tokens.pdf}
%         \label{fig:subfigB}
%     \end{subfigure} 
%     \vspace{-10pt}
%     \caption{Optimal model size (left) and training size (right) at a fixed level of compute (log-log).}
%     \label{fig:optimal_allocation}
% \end{figure}

% Requires: \usepackage{caption,subcaption}
% \begin{table}[h]

\subsection{Effect of Symmetry Loss in Scaling Laws}
\begin{figure}[h]
    \centering
      \begin{subfigure}[b]{0.32\textwidth} 
        \centering
        \includegraphics[width=\linewidth]{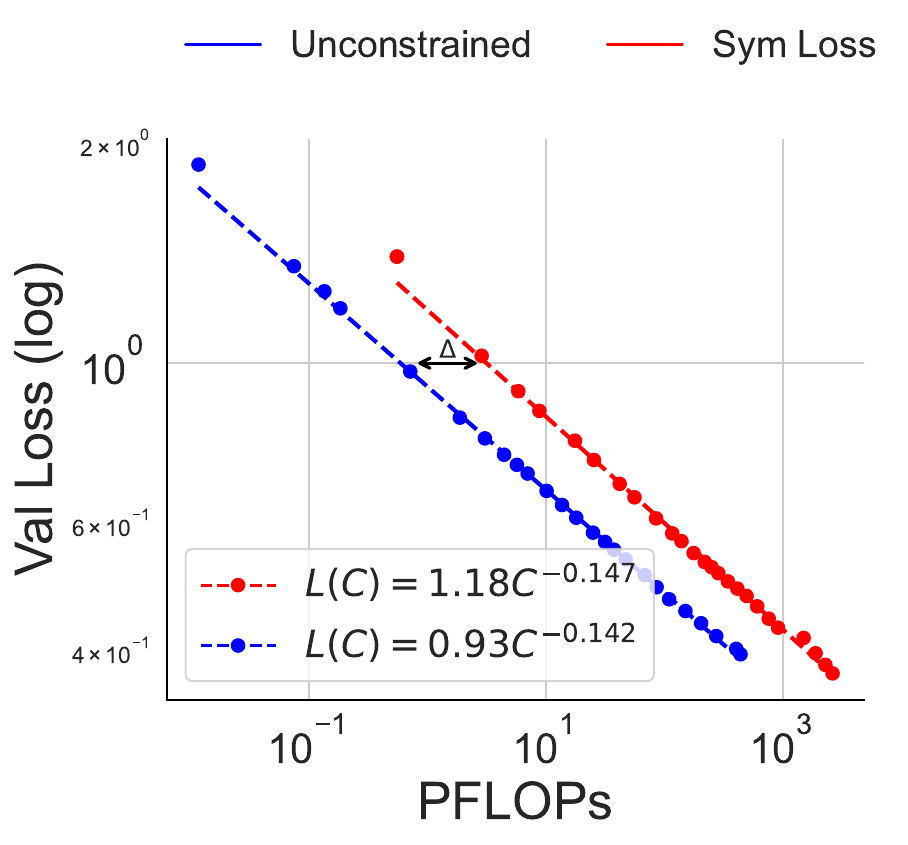}
        \label{fig:subfigA}
    \end{subfigure}
       \begin{subfigure}[b]{0.32\textwidth} 
        \centering
        \includegraphics[width=\linewidth]{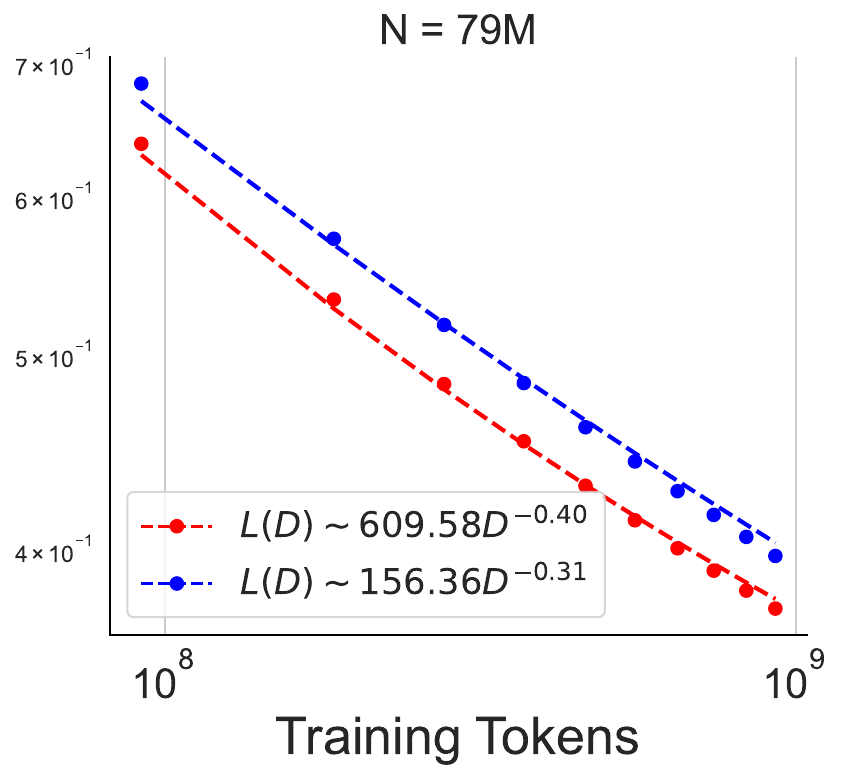}
        \label{fig:subfigA}
    \end{subfigure}
    \begin{subfigure}[b]{0.32\textwidth}
        \centering
        \includegraphics[width=\linewidth]{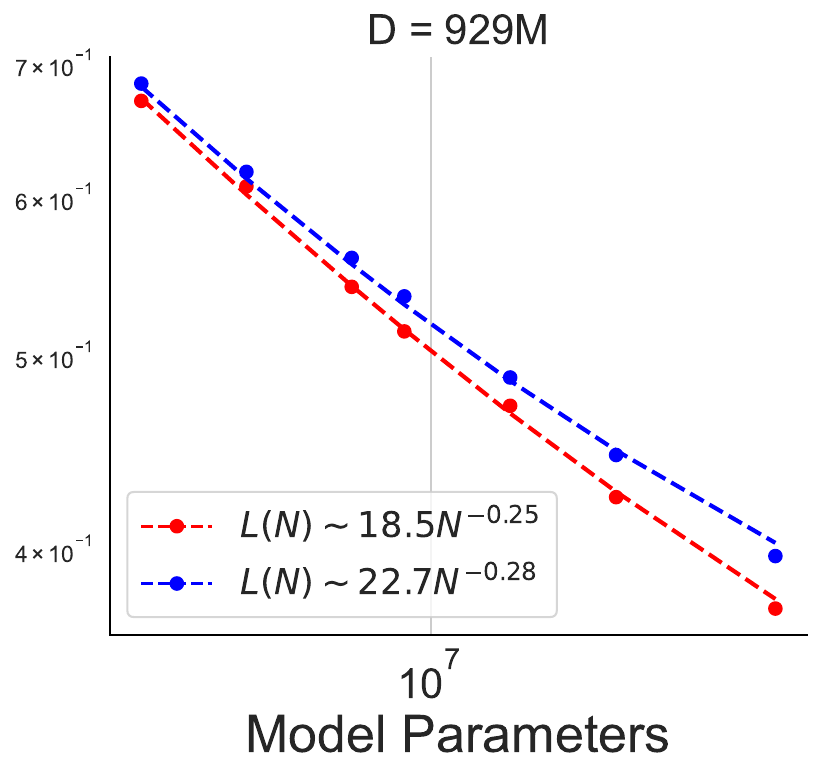}
        \label{fig:subfigB}
    \end{subfigure} 
    \vspace{-10pt}
    \caption{Comparison of scaling exponents for unconstrained MPNNs with/without symmetry regularization. \textbf{Left}: Rightward shift of the log–log loss–compute frontier. \textbf{Middle}: Symmetry loss increases the data-scaling exponent ($\beta$), indicating improved data efficiency. \textbf{Right}: The parameter-scaling exponent ($\alpha$) decreases, suggesting that the regularization benefits larger models more. Validation loss excludes the regularization term (task loss only).}
    \label{fig:scaling_sym_loss}
\end{figure}
In this experiment, we train an unconstrained model augmented with a symmetry-loss term. The loss is
$\mathcal{L}=\mathcal{L}_{\text{obj}}+\lambda\,\mathcal{L}_{\text{sym}}$,
where $\mathcal{L}_{\text{obj}}$ is the task loss in \cref{eq:task_loss} and $\mathcal{L}_{\text{sym}}$ is the symmetry loss in \cref{eq:symmetry_loss} wherein we set $M = 5$. We use unit coefficients for both terms (i.e., $\lambda=1$), as smaller weights ($\lambda \ll 1$) on $\mathcal{L}_{\text{sym}}$ are reported to have negligible effect \citep{elhag2025relaxed}. For validation, we track $\mathcal{L}_{\text{obj}}$, ensuring direct comparison with models trained without the symmetry penalty. We fit the learning-curve trajectories to the functional forms in \cref{eq:scaling_flops} and \cref{eq:scaling_law_N_D}. \Cref{fig:scaling_sym_loss} shows the resulting fits. Compared with models trained without $\mathcal{L}_{\text{sym}}$, we observe:
\begin{itemize}
     \item \textbf{Opposite Changes in Slopes of $D$ and $N$:} Under the scaling form $L(N,D)=L_\infty + A N^{-\alpha} + B D^{-\beta}$, when $N$ is sufficiently large ($N \rightarrow \infty$), adding a symmetry-constraint loss slightly increases the data exponent $\beta$, indicating improved sample-efficiency: the model leverages the regularizer to infer approximate symmetries from data. Conversely, in the infinite-data regime ($D \rightarrow \infty$), a smaller model-size exponent $\alpha$ implies that increasing the parameter count $N$ more effectively reduces loss.
    \item \textbf{Unchanged Compute-Optimal Slope:} We hypothesize that because the $N$- and $D$-slopes change in opposite directions, the induced exponent $\gamma$ with respect to compute $C\propto ND$ is preserved. Furthermore, the sampling-based regularizer in \cref{eq:symmetry_loss} functions as data augmentation: in addition to each original sample, we also evaluate $M$ group-transformed inputs and predict the correspondingly transformed targets. As a result, the training FLOPs scale as
$C_{\text{sym}} = (M+1)\,C_{\text{unconstr}}$, shifting the compute-optimal frontier to the right  by $\Delta \approx \gamma\log(M+1) \approx 0.14 \log_{10}(6)$ in log–log coordinates as shown in \cref{fig:scaling_sym_loss}. Our fits indicate that approximate symmetry enforced via sampling-based augmentation may be unnecessary for compute-optimal scaling, as {the relevant scaling exponents remain unchanged}. 
\end{itemize}
\begin{figure}[h]
    \centering
      \begin{subfigure}[b]{0.32\textwidth} 
        \centering
        \includegraphics[width=\linewidth]{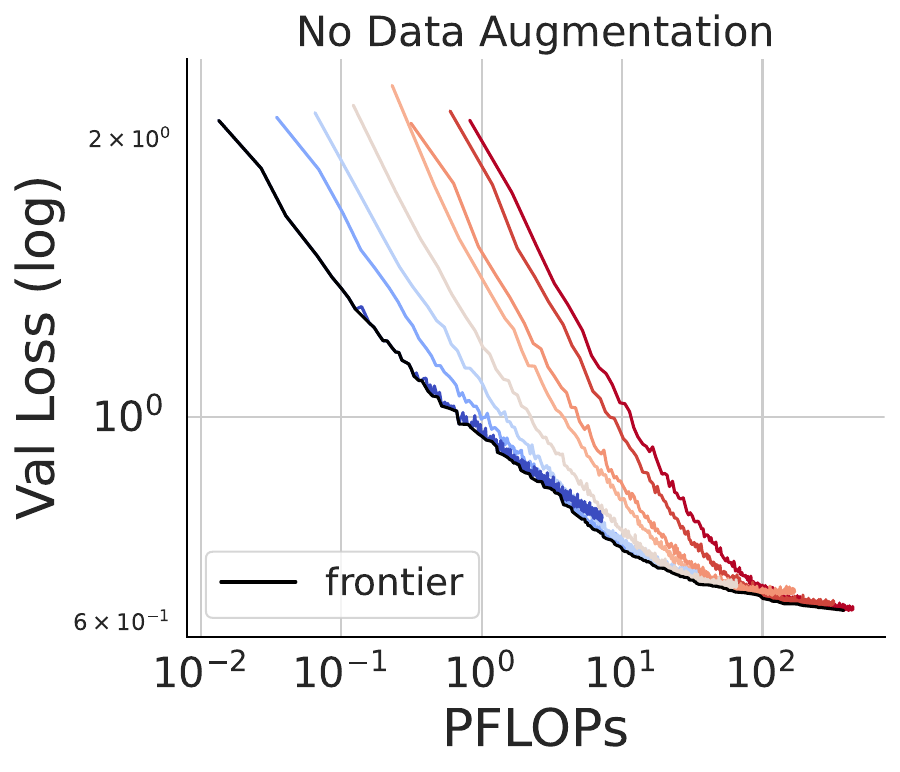}
        \label{fig:subfigA}
    \end{subfigure}
       \begin{subfigure}[b]{0.32\textwidth} 
        \centering
        \includegraphics[width=\linewidth]{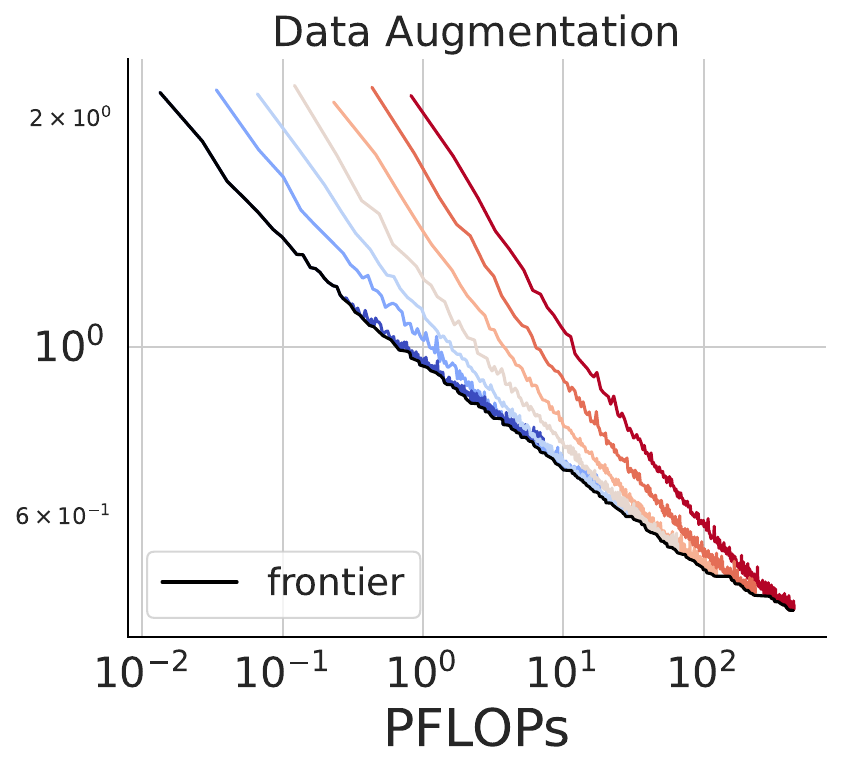}
        \label{fig:subfigA}
    \end{subfigure}
    \begin{subfigure}[b]{0.32\textwidth}
        \centering
        \includegraphics[width=\linewidth]{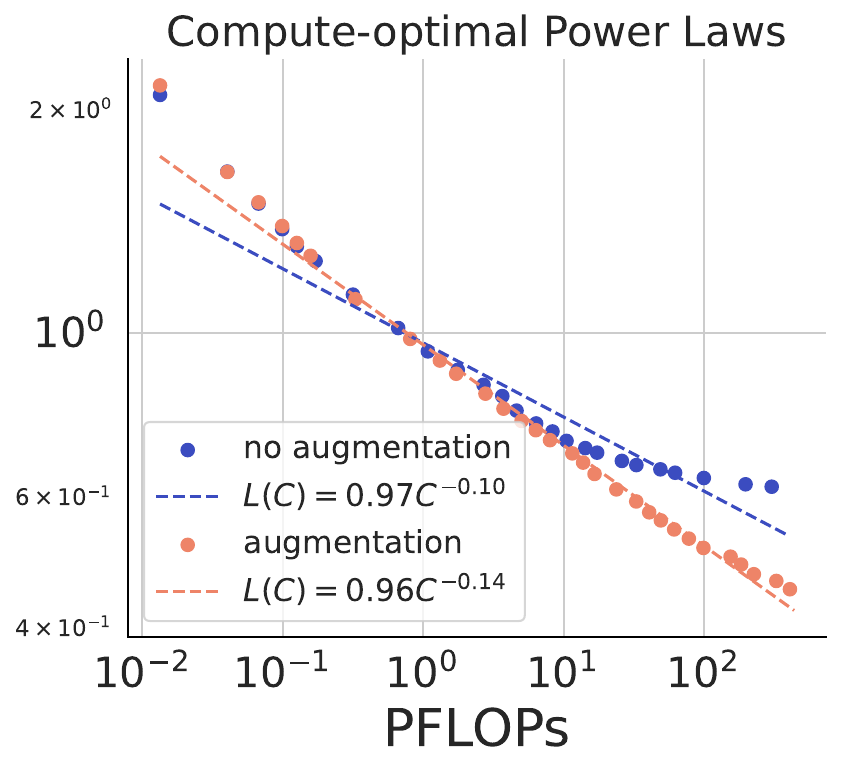}
        \label{fig:subfigB}
    \end{subfigure} 
    \vspace{-10pt}
    \caption{{Loss-compute Pareto frontiers for unconstrained models trained on $1\%$ of the data for 100 epochs, shown without data augmentation (\textbf{Left}) and with data augmentation (\textbf{Middle}). \textbf{Right}: The linear trend in log-log space is broken at late training when data augmentation is omitted, and is recovered when data augmentation is utilized.}}
    \label{fig:ratio-1-epoch-100}
\end{figure}
\vspace{-10pt}
{\subsection{Effect of Multi-Epoch Training in Scaling Laws}
In the previous sections, we present clear scaling-law trends across architectures in the one-pass training regime, keeping our empirical setup aligned with insights from rigorous theoretical works that focus on this setting \citep{paquette2024phases_compute, bordelon2024dynamical_model}. Unfortunately, scaling laws for scenarios in which data is repeated for multiple epochs remain under-explored, both in practice and in theory. To our knowledge, \cite{muennighoff2023dataconstraintscaling} is among the few works investigating this area, showing that under fixed compute, training models for a small number of epochs on repeated data has negligible effects on the loss, behaving almost as if the models are trained on fresh data. We examine this hypothesis in our study by simulating a scenario where data is extremely limited by sampling only $1\%$ of the full dataset, to train the models. The data is repeated over 100 epochs in each training run, which naturally enables the use of data augmentation to improve the data efficiency of unconstrained models. Also, validation losses are recorded after every 1000 gradient steps.}
\begin{wrapfigure}{r}{0.4\textwidth}
  \centering
  \includegraphics[width=\linewidth]{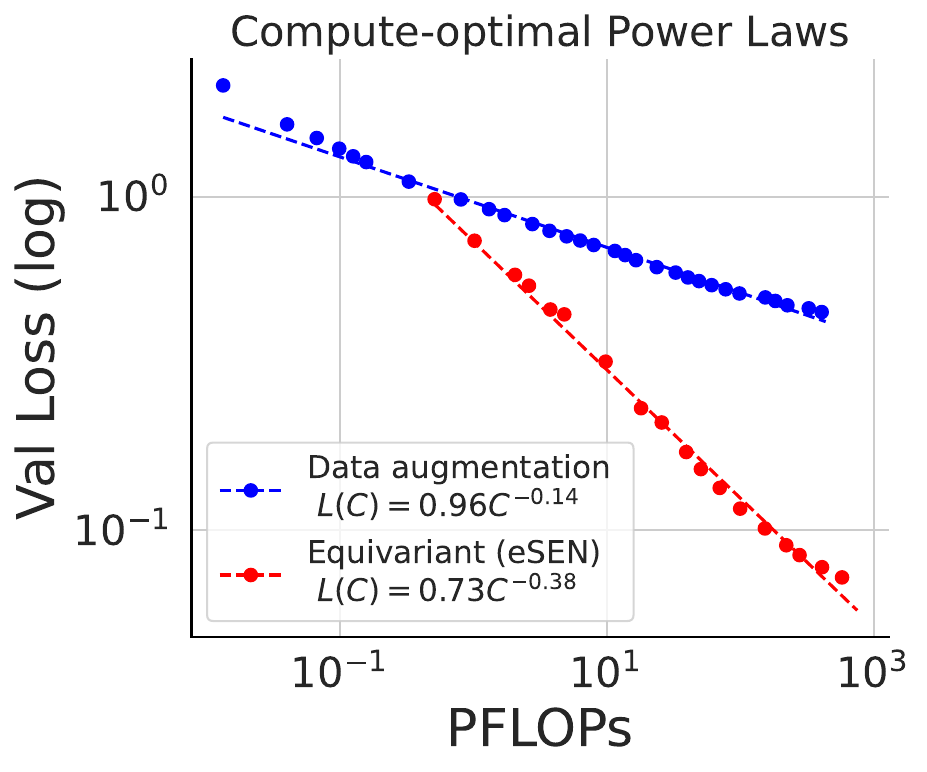}
  \caption{{Scaling with training compute for unconstrained models trained with data augmentation and eSEN. The models are trained on $1\%$ of training dataset for 100 epochs.}}
  \label{fig:ratio_1_comparison}
  \vspace{-20pt}
\end{wrapfigure}
{\Cref{fig:ratio-1-epoch-100} shows the resulting fits in this regime. We observe that for unconstrained models trained without augmentation, the loss-compute frontier follows power-law scaling in early epochs, similar to findings of \cite{muennighoff2023dataconstraintscaling}. However, this power law breaks down when the number of passes exceeds a certain threshold, as heavy data repetition induces overfitting. In contrast, data augmentation substantially stabilizes the learning curves and ultimately recovers the same power law ($\gamma_c \approx 0.14, F_c \approx 0.96$) with respect to the compute-optimal frontier as in the one-epoch training regime, indicating that the effect of data augmentation under low-data regime is the same as adding fresh data in one-pass training over larger datasets. Furthermore, we also examine the power laws of eSEN under this regime, and observe that its compute-optimal power-law holds as the same in one-pass training. Importantly,  \cref{fig:ratio_1_comparison} reveals that the gap between data augmentation and equivariant networks continues to grow as compute increases through multi-epoch training.}
\vspace{-10pt}
\section{Conclusion, Future Works and Limitations}
\label{sec:conclusion}
Our empirical study of scaling laws in the geometric task of interatomic potentials shows that the degree to
which an architecture encodes domain symmetries is correlated with the exponent in its power-law scaling behaviour. The empirical change in exponent is dramatic, suggesting that the role of symmetry potentially extends beyond simply reducing data dimensionality \citep{utkarsh2022scalingmanifold}. This is because the degrees of freedom in the input and output are $\approx 3n$ for $n$ atoms, while the rotation group is only three-dimensional. Our findings, therefore, 
suggest an important future research direction in
developing a theory that explains this scaling behaviour. On the practical side, our work provides a recipe for scaling the model and data size in geometric tasks, such as force fields, and it motivates the development of more scalable models that utilize higher-order representations.

 Other directions for future work are, in part, motivated by the limitations of this work: (1) Our scaling-law analysis focuses on single-epoch, academic-scale settings for NNIPs. Extending it to multi-epoch training and larger models, as well as more diverse models and datasets, is a natural next step. (2) Our study of symmetry losses was confined to one simple choice; it is possible that training with alternative definitions, if scalable, could provide a different scaling behaviour. (3) Our work completely ignores the family of architecture-agnostic equivariant models, such as frame averaging and canonicalization \cite[e.g.,][]{puny2022frame,kaba2023cano,kim2023lps}. We plan to study their scaling laws in the future.
 (4) {Finally, we leave to future work a systematic, large-scale evaluation of denoising pretraining for both unconstrained models \citep{neumann2024orbfastscalableneural, rhodes2025orb} and equivariant networks \citep{liao2024generalizing}}.

\section*{Acknowledgments}
This research is supported by the Canada CIFAR AI Chairs program, Samsung AI Labs, IVADO, and the NSERC Discovery Grant. Computational resources for experiments are provided by the Digital Research Alliance of Canada (Compute Canada) and Mila.

\bibliography{iclr2026_conference}

\begin{thebibliography}{69}
\providecommand{\natexlab}[1]{#1}
\providecommand{\url}[1]{\texttt{#1}}
\expandafter\ifx\csname urlstyle\endcsname\relax
  \providecommand{\doi}[1]{doi: #1}\else
  \providecommand{\doi}{doi: \begingroup \urlstyle{rm}\Url}\fi

\bibitem[Abramson et~al.(2024)Abramson, Adler, Dunger, Evans, Green, Pritzel,
  Ronneberger, Willmore, Ballard, Bambrick, Bodenstein, Evans, Hung, O'Neill,
  Reiman, Tunyasuvunakool, Wu, {\v{Z}}emgulyt{\.{e}}, Arvaniti, Beattie,
  Bertolli, Bridgland, Cherepanov, Congreve, Cowen-Rivers, Cowie, Figurnov,
  Fuchs, Gladman, Jain, Khan, Low, Perlin, Potapenko, Savy, Singh, Stecula,
  Thillaisundaram, Tong, Yakneen, Zhong, Zielinski, {\v{Z}}{\'i}dek, Bapst,
  Kohli, Jaderberg, Hassabis, and Jumper]{abramson2024alphafold3}
Josh Abramson, Jonas Adler, Jack Dunger, Richard Evans, Tim Green, Alexander
  Pritzel, Olaf Ronneberger, Lindsay Willmore, Andrew~J. Ballard, Joshua
  Bambrick, Sebastian~W. Bodenstein, David~A. Evans, Chia-Chun Hung, Michael
  O'Neill, David Reiman, Kathryn Tunyasuvunakool, Zachary Wu, Akvil{\.{e}}
  {\v{Z}}emgulyt{\.{e}}, Eirini Arvaniti, Charles Beattie, Ottavia Bertolli,
  Alex Bridgland, Alexey Cherepanov, Miles Congreve, Alexander~I. Cowen-Rivers,
  Andrew Cowie, Michael Figurnov, Fabian~B. Fuchs, Hannah Gladman, Rishub Jain,
  Yousuf~A. Khan, Caroline M.~R. Low, Kuba Perlin, Anna Potapenko, Pascal Savy,
  Sukhdeep Singh, Adrian Stecula, Ashok Thillaisundaram, Catherine Tong, Sergei
  Yakneen, Ellen~D. Zhong, Michal Zielinski, Augustin {\v{Z}}{\'i}dek, Victor
  Bapst, Pushmeet Kohli, Max Jaderberg, Demis Hassabis, and John~M. Jumper.
\newblock Accurate structure prediction of biomolecular interactions with
  alphafold 3.
\newblock \emph{Nature}, 630\penalty0 (8016):\penalty0 493--500, Jun 2024.
\newblock ISSN 1476-4687.
\newblock \doi{10.1038/s41586-024-07487-w}.
\newblock URL \url{https://doi.org/10.1038/s41586-024-07487-w}.

\bibitem[Ahmad \& Tesauro(1988)Ahmad and Tesauro]{ahmad1998scaling}
Subutai Ahmad and Gerald Tesauro.
\newblock Scaling and generalization in neural networks: A case study.
\newblock In D.~Touretzky (ed.), \emph{Advances in Neural Information
  Processing Systems}, volume~1. Morgan-Kaufmann, 1988.
\newblock URL
  \url{https://proceedings.neurips.cc/paper_files/paper/1988/file/d1f491a404d6854880943e5c3cd9ca25-Paper.pdf}.

\bibitem[Akhound-Sadegh et~al.(2023)Akhound-Sadegh, Perreault-Levasseur,
  Brandstetter, Welling, and Ravanbakhsh]{tara2023liepoint}
Tara Akhound-Sadegh, Laurence Perreault-Levasseur, Johannes Brandstetter, Max
  Welling, and Siamak Ravanbakhsh.
\newblock Lie point symmetry and physics-informed networks.
\newblock In A.~Oh, T.~Naumann, A.~Globerson, K.~Saenko, M.~Hardt, and
  S.~Levine (eds.), \emph{Advances in Neural Information Processing Systems},
  volume~36, pp.\  42468--42481. Curran Associates, Inc., 2023.
\newblock URL
  \url{https://proceedings.neurips.cc/paper_files/paper/2023/file/8493c860bec41705f7743d5764301b94-Paper-Conference.pdf}.

\bibitem[Alabdulmohsin et~al.(2022)Alabdulmohsin, Neyshabur, and
  Zhai]{aladulmohsin2022language_vision_scaling_laws}
Ibrahim~M Alabdulmohsin, Behnam Neyshabur, and Xiaohua Zhai.
\newblock Revisiting neural scaling laws in language and vision.
\newblock In S.~Koyejo, S.~Mohamed, A.~Agarwal, D.~Belgrave, K.~Cho, and A.~Oh
  (eds.), \emph{Advances in Neural Information Processing Systems}, volume~35,
  pp.\  22300--22312. Curran Associates, Inc., 2022.
\newblock URL
  \url{https://proceedings.neurips.cc/paper_files/paper/2022/file/8c22e5e918198702765ecff4b20d0a90-Paper-Conference.pdf}.

\bibitem[Anderson et~al.(2019)Anderson, Hy, and Kondor]{anderson2019cormorant}
Brandon Anderson, Truong~Son Hy, and Risi Kondor.
\newblock Cormorant: Covariant molecular neural networks.
\newblock In H.~Wallach, H.~Larochelle, A.~Beygelzimer, F.~d\textquotesingle
  Alch\'{e}-Buc, E.~Fox, and R.~Garnett (eds.), \emph{Advances in Neural
  Information Processing Systems}, volume~32. Curran Associates, Inc., 2019.
\newblock URL
  \url{https://proceedings.neurips.cc/paper_files/paper/2019/file/03573b32b2746e6e8ca98b9123f2249b-Paper.pdf}.

\bibitem[Bahri et~al.(2024)Bahri, Dyer, Kaplan, Lee, and
  Sharma]{bahri2024explaining_nn_scaling}
Yasaman Bahri, Ethan Dyer, Jared Kaplan, Jaehoon Lee, and Utkarsh Sharma.
\newblock Explaining neural scaling laws.
\newblock \emph{Proceedings of the National Academy of Sciences}, 121\penalty0
  (27):\penalty0 e2311878121, 2024.

\bibitem[Bai et~al.(2025)Bai, Fu, Xie, and Meng]{bai2025regularization}
Yulu Bai, Jiahong Fu, Qi~Xie, and Deyu Meng.
\newblock A regularization-guided equivariant approach for image restoration.
\newblock In \emph{Proceedings of the Computer Vision and Pattern Recognition
  Conference}, pp.\  2300--2310, 2025.

\bibitem[Batatia et~al.(2022)Batatia, Kovacs, Simm, Ortner, and
  Csanyi]{batatia2022mace}
Ilyes Batatia, David~P Kovacs, Gregor Simm, Christoph Ortner, and Gabor Csanyi.
\newblock Mace: Higher order equivariant message passing neural networks for
  fast and accurate force fields.
\newblock In S.~Koyejo, S.~Mohamed, A.~Agarwal, D.~Belgrave, K.~Cho, and A.~Oh
  (eds.), \emph{Advances in Neural Information Processing Systems}, volume~35,
  pp.\  11423--11436. Curran Associates, Inc., 2022.
\newblock URL
  \url{https://proceedings.neurips.cc/paper_files/paper/2022/file/4a36c3c51af11ed9f34615b81edb5bbc-Paper-Conference.pdf}.

\bibitem[Batzner et~al.(2022)Batzner, Musaelian, Sun, Geiger, Mailoa,
  Kornbluth, Molinari, Smidt, and Kozinsky]{nequip}
Simon Batzner, Albert Musaelian, Lixin Sun, Mario Geiger, Jonathan~P. Mailoa,
  Mordechai Kornbluth, Nicola Molinari, Tess~E. Smidt, and Boris Kozinsky.
\newblock E(3)-equivariant graph neural networks for data-efficient and
  accurate interatomic potentials.
\newblock \emph{Nature Communications}, 13\penalty0 (1), May 2022.
\newblock \doi{10.1038/s41467-022-29939-5}.
\newblock URL \url{https://doi.org/10.1038/s41467-022-29939-5}.

\bibitem[Bigi et~al.(2025)Bigi, Langer, and Ceriotti]{bigi2025the}
Filippo Bigi, Marcel~F. Langer, and Michele Ceriotti.
\newblock The dark side of the forces: assessing non-conservative force models
  for atomistic machine learning.
\newblock In \emph{Forty-second International Conference on Machine Learning},
  2025.
\newblock URL \url{https://openreview.net/forum?id=OEl3L8osas}.

\bibitem[Bordelon et~al.(2024)Bordelon, Atanasov, and
  Pehlevan]{bordelon2024dynamical_model}
Blake Bordelon, Alexander Atanasov, and Cengiz Pehlevan.
\newblock A dynamical model of neural scaling laws.
\newblock In Ruslan Salakhutdinov, Zico Kolter, Katherine Heller, Adrian
  Weller, Nuria Oliver, Jonathan Scarlett, and Felix Berkenkamp (eds.),
  \emph{Proceedings of the 41st International Conference on Machine Learning},
  volume 235 of \emph{Proceedings of Machine Learning Research}, pp.\
  4345--4382. PMLR, 21--27 Jul 2024.
\newblock URL \url{https://proceedings.mlr.press/v235/bordelon24a.html}.

\bibitem[Brehmer et~al.(2025)Brehmer, Behrends, Haan, and
  Cohen]{brehmer2025does}
Johann Brehmer, S{\"o}nke Behrends, Pim~De Haan, and Taco Cohen.
\newblock Does equivariance matter at scale?, 2025.
\newblock URL \url{https://openreview.net/forum?id=iIWeyfGTof}.

\bibitem[Caballero et~al.(2023)Caballero, Gupta, Rish, and
  Krueger]{caballero2023broken}
Ethan Caballero, Kshitij Gupta, Irina Rish, and David Krueger.
\newblock Broken neural scaling laws.
\newblock In \emph{The Eleventh International Conference on Learning
  Representations}, 2023.
\newblock URL \url{https://openreview.net/forum?id=sckjveqlCZ}.

\bibitem[Choshen et~al.(2025)Choshen, Zhang, and
  Andreas]{choshen2025hitchhiker}
Leshem Choshen, Yang Zhang, and Jacob Andreas.
\newblock A hitchhiker’s guide to scaling law estimation.
\newblock In \emph{International Conference on Machine Learning}, 2025.

\bibitem[Cortes et~al.(1993)Cortes, Jackel, Solla, Vapnik, and
  Denker]{cortes1993learning}
Corinna Cortes, Lawrence~D Jackel, Sara Solla, Vladimir Vapnik, and John
  Denker.
\newblock Learning curves: Asymptotic values and rate of convergence.
\newblock \emph{Advances in neural information processing systems}, 6, 1993.

\bibitem[Dangovski et~al.(2021)Dangovski, Jing, Loh, Han, Srivastava, Cheung,
  Agrawal, and Solja{\v{c}}i{\'c}]{dangovski2021equivariant}
Rumen Dangovski, Li~Jing, Charlotte Loh, Seungwook Han, Akash Srivastava, Brian
  Cheung, Pulkit Agrawal, and Marin Solja{\v{c}}i{\'c}.
\newblock Equivariant contrastive learning.
\newblock \emph{arXiv preprint arXiv:2111.00899}, 2021.

\bibitem[Defazio et~al.(2024)Defazio, Yang, Mehta, Mishchenko, Khaled, and
  Cutkosky]{defazio2024schedulerfree}
Aaron Defazio, Xingyu Yang, Harsh Mehta, Konstantin Mishchenko, Ahmed Khaled,
  and Ashok Cutkosky.
\newblock The road less scheduled.
\newblock In A.~Globerson, L.~Mackey, D.~Belgrave, A.~Fan, U.~Paquet,
  J.~Tomczak, and C.~Zhang (eds.), \emph{Advances in Neural Information
  Processing Systems}, volume~37, pp.\  9974--10007. Curran Associates, Inc.,
  2024.
\newblock URL
  \url{https://proceedings.neurips.cc/paper_files/paper/2024/file/136b9a13861308c8948cd308ccd02658-Paper-Conference.pdf}.

\bibitem[Deng et~al.(2023)Deng, Zhong, Jun, Riebesell, Han, Bartel, and
  Ceder]{deng2023chgnet}
Bowen Deng, Peichen Zhong, KyuJung Jun, Janosh Riebesell, Kevin Han,
  Christopher~J Bartel, and Gerbrand Ceder.
\newblock Chgnet as a pretrained universal neural network potential for
  charge-informed atomistic modelling.
\newblock \emph{Nature Machine Intelligence}, 5\penalty0 (9):\penalty0
  1031--1041, 2023.

\bibitem[Duval et~al.(2023)Duval, Schmidt, Hern\'{a}ndez-Garc\'{\i}a, Miret,
  Malliaros, Bengio, and Rolnick]{duval2023stochasticframe}
Alexandre~Agm Duval, Victor Schmidt, Alex Hern\'{a}ndez-Garc\'{\i}a, Santiago
  Miret, Fragkiskos~D. Malliaros, Yoshua Bengio, and David Rolnick.
\newblock {FAEN}et: Frame averaging equivariant {GNN} for materials modeling.
\newblock In Andreas Krause, Emma Brunskill, Kyunghyun Cho, Barbara Engelhardt,
  Sivan Sabato, and Jonathan Scarlett (eds.), \emph{Proceedings of the 40th
  International Conference on Machine Learning}, volume 202 of
  \emph{Proceedings of Machine Learning Research}, pp.\  9013--9033. PMLR,
  23--29 Jul 2023.
\newblock URL \url{https://proceedings.mlr.press/v202/duval23a.html}.

\bibitem[Elhag et~al.(2025)Elhag, Rusch, Giovanni, and
  Bronstein]{elhag2025relaxed}
Ahmed A.~A. Elhag, T.~Konstantin Rusch, Francesco~Di Giovanni, and Michael~M.
  Bronstein.
\newblock Relaxed equivariance via multitask learning.
\newblock In \emph{ICLR 2025 Workshop on Machine Learning for Genomics
  Explorations}, 2025.
\newblock URL \url{https://openreview.net/forum?id=8kZSO4WbTh}.

\bibitem[Escriche \& Jegelka(2025)Escriche and Jegelka]{escriche2025learning}
Eduardo~Santos Escriche and Stefanie Jegelka.
\newblock Learning equivariant models by discovering symmetries with learnable
  augmentations.
\newblock \emph{arXiv preprint arXiv:2506.03914}, 2025.

\bibitem[Frey et~al.(2022)Frey, Soklaski, Axelrod, Samsi, Gomez-Bombarelli,
  Coley, and Gadepally]{neural_scale_of_chemical_models}
Nathan Frey, Ryan Soklaski, Simon Axelrod, Siddharth Samsi, Rafael
  Gomez-Bombarelli, Connor Coley, and Vijay Gadepally.
\newblock Neural scaling of deep chemical models.
\newblock \emph{ChemRxiv}, 2022.
\newblock \doi{10.26434/chemrxiv-2022-3s512}.

\bibitem[Frey et~al.(2023)Frey, Soklaski, Axelrod, Samsi, Gomez-Bombarelli,
  Coley, and Gadepally]{frey2023neuralchemicalscaling}
Nathan~C Frey, Ryan Soklaski, Simon Axelrod, Siddharth Samsi, Rafael
  Gomez-Bombarelli, Connor~W Coley, and Vijay Gadepally.
\newblock Neural scaling of deep chemical models.
\newblock \emph{Nature Machine Intelligence}, 5\penalty0 (11):\penalty0
  1297--1305, 2023.

\bibitem[Fu et~al.(2025)Fu, Wood, Barroso-Luque, Levine, Gao, Dzamba, and
  Zitnick]{fu2025eSEN}
Xiang Fu, Brandon~M Wood, Luis Barroso-Luque, Daniel~S. Levine, Meng Gao, Misko
  Dzamba, and C.~Lawrence Zitnick.
\newblock Learning smooth and expressive interatomic potentials for physical
  property prediction.
\newblock In \emph{Forty-second International Conference on Machine Learning},
  2025.
\newblock URL \url{https://openreview.net/forum?id=R0PBjxIbgm}.

\bibitem[Gasteiger et~al.(2020{\natexlab{a}})Gasteiger, Giri, Margraf, and
  G{\"u}nnemann]{dimenet_pp}
Johannes Gasteiger, Shankari Giri, Johannes~T. Margraf, and Stephan
  G{\"u}nnemann.
\newblock Fast and uncertainty-aware directional message passing for
  non-equilibrium molecules.
\newblock In \emph{Machine Learning for Molecules Workshop, NeurIPS},
  2020{\natexlab{a}}.

\bibitem[Gasteiger et~al.(2020{\natexlab{b}})Gasteiger, Groß, and
  Günnemann]{Gasteiger2020Directional}
Johannes Gasteiger, Janek Groß, and Stephan Günnemann.
\newblock Directional message passing for molecular graphs.
\newblock In \emph{International Conference on Learning Representations},
  2020{\natexlab{b}}.
\newblock URL \url{https://openreview.net/forum?id=B1eWbxStPH}.

\bibitem[Gasteiger et~al.(2022)Gasteiger, Shuaibi, Sriram, Günnemann, Ulissi,
  Zitnick, and Das]{gemnet_oc}
Johannes Gasteiger, Muhammed Shuaibi, Anuroop Sriram, Stephan Günnemann,
  Zachary Ulissi, C.~Lawrence Zitnick, and Abhishek Das.
\newblock How do graph networks generalize to large and diverse molecular
  systems?
\newblock \emph{arxiv preprint arxiv:2204.02782}, 2022.

\bibitem[Henighan et~al.(2020)Henighan, Kaplan, Katz, Chen, Hesse, Jackson,
  Jun, Brown, Dhariwal, Gray, et~al.]{henighan2020scaling}
Tom Henighan, Jared Kaplan, Mor Katz, Mark Chen, Christopher Hesse, Jacob
  Jackson, Heewoo Jun, Tom~B Brown, Prafulla Dhariwal, Scott Gray, et~al.
\newblock Scaling laws for autoregressive generative modeling.
\newblock \emph{arXiv preprint arXiv:2010.14701}, 2020.

\bibitem[Hestness et~al.(2017)Hestness, Narang, Ardalani, Diamos, Jun,
  Kianinejad, Patwary, Yang, and Zhou]{hestness2017deep}
Joel Hestness, Sharan Narang, Newsha Ardalani, Gregory Diamos, Heewoo Jun,
  Hassan Kianinejad, Md~Mostofa~Ali Patwary, Yang Yang, and Yanqi Zhou.
\newblock Deep learning scaling is predictable, empirically.
\newblock \emph{arXiv preprint arXiv:1712.00409}, 2017.

\bibitem[Hoffmann et~al.(2022)Hoffmann, Borgeaud, Mensch, Buchatskaya, Cai,
  Rutherford, Casas, Hendricks, Welbl, Clark, et~al.]{hoffmann2022chinchilla}
Jordan Hoffmann, Sebastian Borgeaud, Arthur Mensch, Elena Buchatskaya, Trevor
  Cai, Eliza Rutherford, Diego de~Las Casas, Lisa~Anne Hendricks, Johannes
  Welbl, Aidan Clark, et~al.
\newblock Training compute-optimal large language models.
\newblock \emph{arXiv preprint arXiv:2203.15556}, 2022.

\bibitem[Hu et~al.(2024)Hu, Tu, Han, He, Cui, Long, Zheng, Fang, Huang, Zhao,
  et~al.]{hu2024minicpm}
Shengding Hu, Yuge Tu, Xu~Han, Chaoqun He, Ganqu Cui, Xiang Long, Zhi Zheng,
  Yewei Fang, Yuxiang Huang, Weilin Zhao, et~al.
\newblock Minicpm: Unveiling the potential of small language models with
  scalable training strategies.
\newblock \emph{arXiv preprint arXiv:2404.06395}, 2024.

\bibitem[Jia et~al.(2020)Jia, Wang, Chen, Lu, Lin, Car, E, and
  Zhang]{push_limit_of_md_100m}
Weile Jia, Han Wang, Mohan Chen, Denghui Lu, Lin Lin, Roberto Car, Weinan E,
  and Linfeng Zhang.
\newblock Pushing the limit of molecular dynamics with ab initio accuracy to
  100 million atoms with machine learning.
\newblock In \emph{Proceedings of the International Conference for High
  Performance Computing, Networking, Storage and Analysis}, SC '20. IEEE Press,
  2020.

\bibitem[Joshi et~al.(2023)Joshi, Bodnar, Mathis, Cohen, and Lio]{joshi2023gwl}
Chaitanya~K. Joshi, Cristian Bodnar, Simon~V Mathis, Taco Cohen, and Pietro
  Lio.
\newblock On the expressive power of geometric graph neural networks.
\newblock In Andreas Krause, Emma Brunskill, Kyunghyun Cho, Barbara Engelhardt,
  Sivan Sabato, and Jonathan Scarlett (eds.), \emph{Proceedings of the 40th
  International Conference on Machine Learning}, volume 202 of
  \emph{Proceedings of Machine Learning Research}, pp.\  15330--15355. PMLR,
  23--29 Jul 2023.
\newblock URL \url{https://proceedings.mlr.press/v202/joshi23a.html}.

\bibitem[Kaba et~al.(2023)Kaba, Mondal, Zhang, Bengio, and
  Ravanbakhsh]{kaba2023cano}
S\'{e}kou-Oumar Kaba, Arnab~Kumar Mondal, Yan Zhang, Yoshua Bengio, and Siamak
  Ravanbakhsh.
\newblock Equivariance with learned canonicalization functions.
\newblock In Andreas Krause, Emma Brunskill, Kyunghyun Cho, Barbara Engelhardt,
  Sivan Sabato, and Jonathan Scarlett (eds.), \emph{Proceedings of the 40th
  International Conference on Machine Learning}, volume 202 of
  \emph{Proceedings of Machine Learning Research}, pp.\  15546--15566. PMLR,
  23--29 Jul 2023.
\newblock URL \url{https://proceedings.mlr.press/v202/kaba23a.html}.

\bibitem[Kaplan et~al.(2020)Kaplan, McCandlish, Henighan, Brown, Chess, Child,
  Gray, Radford, Wu, and Amodei]{kaplan2020scaling}
Jared Kaplan, Sam McCandlish, Tom Henighan, Tom~B Brown, Benjamin Chess, Rewon
  Child, Scott Gray, Alec Radford, Jeffrey Wu, and Dario Amodei.
\newblock Scaling laws for neural language models.
\newblock \emph{arXiv preprint arXiv:2001.08361}, 2020.

\bibitem[Kim et~al.(2023{\natexlab{a}})Kim, Lee, Yang, and
  Lee]{kim2023softloss}
Hyunsu Kim, Hyungi Lee, Hongseok Yang, and Juho Lee.
\newblock Regularizing towards soft equivariance under mixed symmetries.
\newblock In Andreas Krause, Emma Brunskill, Kyunghyun Cho, Barbara Engelhardt,
  Sivan Sabato, and Jonathan Scarlett (eds.), \emph{Proceedings of the 40th
  International Conference on Machine Learning}, volume 202 of
  \emph{Proceedings of Machine Learning Research}, pp.\  16712--16727. PMLR,
  23--29 Jul 2023{\natexlab{a}}.
\newblock URL \url{https://proceedings.mlr.press/v202/kim23p.html}.

\bibitem[Kim et~al.(2023{\natexlab{b}})Kim, Nguyen, Suleymanzade, An, and
  Hong]{kim2023lps}
Jinwoo Kim, Dat Nguyen, Ayhan Suleymanzade, Hyeokjun An, and Seunghoon Hong.
\newblock Learning probabilistic symmetrization for architecture agnostic
  equivariance.
\newblock In A.~Oh, T.~Naumann, A.~Globerson, K.~Saenko, M.~Hardt, and
  S.~Levine (eds.), \emph{Advances in Neural Information Processing Systems},
  volume~36, pp.\  18582--18612. Curran Associates, Inc., 2023{\natexlab{b}}.
\newblock URL
  \url{https://proceedings.neurips.cc/paper_files/paper/2023/file/3b5c7c9c5c7bd77eb73d0baec7a07165-Paper-Conference.pdf}.

\bibitem[Klicpera et~al.(2021)Klicpera, Becker, and G{\"u}nnemann]{gemnet}
Johannes Klicpera, Florian Becker, and Stephan G{\"u}nnemann.
\newblock Gemnet: Universal directional graph neural networks for molecules.
\newblock In \emph{Conference on Neural Information Processing (NeurIPS)},
  2021.

\bibitem[Levine et~al.(2025)Levine, Shuaibi, Spotte-Smith, Taylor, Hasyim,
  Michel, Batatia, Cs{\'a}nyi, Dzamba, Eastman, et~al.]{levine2025openmol}
Daniel~S Levine, Muhammed Shuaibi, Evan Walter~Clark Spotte-Smith, Michael~G
  Taylor, Muhammad~R Hasyim, Kyle Michel, Ilyes Batatia, G{\'a}bor Cs{\'a}nyi,
  Misko Dzamba, Peter Eastman, et~al.
\newblock The open molecules 2025 (omol25) dataset, evaluations, and models.
\newblock \emph{arXiv preprint arXiv:2505.08762}, 2025.

\bibitem[Levy et~al.(2023)Levy, Kaba, Gonzales, Miret, and
  Ravanbakhsh]{levy2023using}
Daniel Levy, S{\'e}kou-Oumar Kaba, Carmelo Gonzales, Santiago Miret, and Siamak
  Ravanbakhsh.
\newblock Using multiple vector channels improves e (n)-equivariant graph
  neural networks.
\newblock \emph{arXiv preprint arXiv:2309.03139}, 2023.

\bibitem[Li et~al.(2025)Li, Ye, Pasini, Choi, Wan, Lin, and
  Balaprakash]{li2025scaling}
Chaojian Li, Zhifan Ye, Massimiliano~Lupo Pasini, Jong~Youl Choi, Cheng Wan,
  Yingyan~Celine Lin, and Prasanna Balaprakash.
\newblock Scaling laws of graph neural networks for atomistic materials
  modeling.
\newblock \emph{arXiv preprint arXiv:2504.08112}, 2025.

\bibitem[Liao \& Smidt(2023)Liao and Smidt]{liao2023equiformer}
Yi-Lun Liao and Tess Smidt.
\newblock Equiformer: Equivariant graph attention transformer for 3d atomistic
  graphs.
\newblock In \emph{The Eleventh International Conference on Learning
  Representations}, 2023.
\newblock URL \url{https://openreview.net/forum?id=KwmPfARgOTD}.

\bibitem[Liao et~al.(2024{\natexlab{a}})Liao, Smidt, Shuaibi, and
  Das]{liao2024generalizing}
Yi-Lun Liao, Tess Smidt, Muhammed Shuaibi, and Abhishek Das.
\newblock Generalizing denoising to non-equilibrium structures improves
  equivariant force fields.
\newblock \emph{Transactions on Machine Learning Research}, 2024{\natexlab{a}}.
\newblock ISSN 2835-8856.
\newblock URL \url{https://openreview.net/forum?id=whGzYUbIWA}.

\bibitem[Liao et~al.(2024{\natexlab{b}})Liao, Wood, Das, and
  Smidt]{liao2024equiformerv}
Yi-Lun Liao, Brandon~M Wood, Abhishek Das, and Tess Smidt.
\newblock Equiformerv2: Improved equivariant transformer for scaling to
  higher-degree representations.
\newblock In \emph{The Twelfth International Conference on Learning
  Representations}, 2024{\natexlab{b}}.
\newblock URL \url{https://openreview.net/forum?id=mCOBKZmrzD}.

\bibitem[Liu et~al.(2024)Liu, Mao, Chen, Zhao, Shah, and
  Tang]{liu2024graphscaling}
Jingzhe Liu, Haitao Mao, Zhikai Chen, Tong Zhao, Neil Shah, and Jiliang Tang.
\newblock Neural scaling laws on graphs.
\newblock \emph{arXiv e-prints}, pp.\  arXiv--2402, 2024.

\bibitem[Loukas(2020)]{Loukas2020What}
Andreas Loukas.
\newblock What graph neural networks cannot learn: depth vs width.
\newblock In \emph{International Conference on Learning Representations}, 2020.
\newblock URL \url{https://openreview.net/forum?id=B1l2bp4YwS}.

\bibitem[Maloney et~al.(2022)Maloney, Roberts, and Sully]{maloney2022solvable}
Alexander Maloney, Daniel~A Roberts, and James Sully.
\newblock A solvable model of neural scaling laws.
\newblock \emph{arXiv preprint arXiv:2210.16859}, 2022.

\bibitem[Muennighoff et~al.(2023)Muennighoff, Rush, Barak, Le~Scao, Tazi,
  Piktus, Pyysalo, Wolf, and Raffel]{muennighoff2023dataconstraintscaling}
Niklas Muennighoff, Alexander Rush, Boaz Barak, Teven Le~Scao, Nouamane Tazi,
  Aleksandra Piktus, Sampo Pyysalo, Thomas Wolf, and Colin~A Raffel.
\newblock Scaling data-constrained language models.
\newblock In A.~Oh, T.~Naumann, A.~Globerson, K.~Saenko, M.~Hardt, and
  S.~Levine (eds.), \emph{Advances in Neural Information Processing Systems},
  volume~36, pp.\  50358--50376. Curran Associates, Inc., 2023.
\newblock URL
  \url{https://proceedings.neurips.cc/paper_files/paper/2023/file/9d89448b63ce1e2e8dc7af72c984c196-Paper-Conference.pdf}.

\bibitem[Neumann et~al.(2024)Neumann, Gin, Rhodes, Bennett, Li, Choubisa,
  Hussey, and Godwin]{neumann2024orbfastscalableneural}
Mark Neumann, James Gin, Benjamin Rhodes, Steven Bennett, Zhiyi Li, Hitarth
  Choubisa, Arthur Hussey, and Jonathan Godwin.
\newblock Orb: A fast, scalable neural network potential, 2024.
\newblock URL \url{https://arxiv.org/abs/2410.22570}.

\bibitem[Paquette et~al.(2024)Paquette, Paquette, Xiao, and
  Pennington]{paquette2024phases_compute}
Elliot Paquette, Courtney Paquette, Lechao Xiao, and Jeffrey Pennington.
\newblock 4+3 phases of compute-optimal neural scaling laws.
\newblock In A.~Globerson, L.~Mackey, D.~Belgrave, A.~Fan, U.~Paquet,
  J.~Tomczak, and C.~Zhang (eds.), \emph{Advances in Neural Information
  Processing Systems}, volume~37, pp.\  16459--16537. Curran Associates, Inc.,
  2024.
\newblock \doi{10.52202/079017-0526}.
\newblock URL
  \url{https://proceedings.neurips.cc/paper_files/paper/2024/file/1dccfc3ee01871d05e33457c61037d59-Paper-Conference.pdf}.

\bibitem[Passaro \& Zitnick(2023)Passaro and Zitnick]{passaro2023reduce}
Saro Passaro and C.~Lawrence Zitnick.
\newblock Reducing {SO}(3) convolutions to {SO}(2) for efficient equivariant
  {GNN}s.
\newblock In Andreas Krause, Emma Brunskill, Kyunghyun Cho, Barbara Engelhardt,
  Sivan Sabato, and Jonathan Scarlett (eds.), \emph{Proceedings of the 40th
  International Conference on Machine Learning}, volume 202 of
  \emph{Proceedings of Machine Learning Research}, pp.\  27420--27438. PMLR,
  23--29 Jul 2023.
\newblock URL \url{https://proceedings.mlr.press/v202/passaro23a.html}.

\bibitem[Pengmei et~al.(2025)Pengmei, Shen, Wang, Collins, and
  Rangwala]{pengmei2025pushing}
Zihan Pengmei, Zhengyuan Shen, Zichen Wang, Marcus~D. Collins, and Huzefa
  Rangwala.
\newblock Pushing the limits of all-atom geometric graph neural networks:
  Pre-training, scaling, and zero-shot transfer.
\newblock In \emph{The Thirteenth International Conference on Learning
  Representations}, 2025.
\newblock URL \url{https://openreview.net/forum?id=4S2L519nIX}.

\bibitem[Petrache \& Trivedi(2023)Petrache and
  Trivedi]{petrache2023approximation}
Mircea Petrache and Shubhendu Trivedi.
\newblock Approximation-generalization trade-offs under (approximate) group
  equivariance.
\newblock In A.~Oh, T.~Naumann, A.~Globerson, K.~Saenko, M.~Hardt, and
  S.~Levine (eds.), \emph{Advances in Neural Information Processing Systems},
  volume~36, pp.\  61936--61959. Curran Associates, Inc., 2023.
\newblock URL
  \url{https://proceedings.neurips.cc/paper_files/paper/2023/file/c35f8e2fc6d81f195009a1d2ae5f6ae9-Paper-Conference.pdf}.

\bibitem[Puny et~al.(2022)Puny, Atzmon, Smith, Misra, Grover, Ben-Hamu, and
  Lipman]{puny2022frame}
Omri Puny, Matan Atzmon, Edward~J. Smith, Ishan Misra, Aditya Grover, Heli
  Ben-Hamu, and Yaron Lipman.
\newblock Frame averaging for invariant and equivariant network design.
\newblock In \emph{International Conference on Learning Representations}, 2022.
\newblock URL \url{https://openreview.net/forum?id=zIUyj55nXR}.

\bibitem[Qu \& Krishnapriyan(2024)Qu and Krishnapriyan]{qu2024scalable}
Eric Qu and Aditi~S. Krishnapriyan.
\newblock The importance of being scalable: Improving the speed and accuracy of
  neural network interatomic potentials across chemical domains.
\newblock In A.~Globerson, L.~Mackey, D.~Belgrave, A.~Fan, U.~Paquet,
  J.~Tomczak, and C.~Zhang (eds.), \emph{Advances in Neural Information
  Processing Systems}, volume~37, pp.\  139030--139053. Curran Associates,
  Inc., 2024.
\newblock URL
  \url{https://proceedings.neurips.cc/paper_files/paper/2024/file/fad8e1915f66161581bb127ccf01092e-Paper-Conference.pdf}.

\bibitem[Rhodes et~al.(2025)Rhodes, Vandenhaute, {\v{S}}imkus, Gin, Godwin,
  Duignan, and Neumann]{rhodes2025orb}
Benjamin Rhodes, Sander Vandenhaute, Vaidotas {\v{S}}imkus, James Gin, Jonathan
  Godwin, Tim Duignan, and Mark Neumann.
\newblock Orb-v3: atomistic simulation at scale.
\newblock \emph{arXiv preprint arXiv:2504.06231}, 2025.

\bibitem[Satorras et~al.(2021)Satorras, Hoogeboom, and Welling]{egnn}
V\'{\i}ctor~Garcia Satorras, Emiel Hoogeboom, and Max Welling.
\newblock E(n) equivariant graph neural networks.
\newblock In \emph{International Conference on Machine Learning (ICML)}, 2021.

\bibitem[Sharma \& Kaplan(2022)Sharma and Kaplan]{utkarsh2022scalingmanifold}
Utkarsh Sharma and Jared Kaplan.
\newblock Scaling laws from the data manifold dimension.
\newblock \emph{Journal of Machine Learning Research}, 23\penalty0
  (9):\penalty0 1--34, 2022.
\newblock URL \url{http://jmlr.org/papers/v23/20-1111.html}.

\bibitem[Snell et~al.(2025)Snell, Lee, Xu, and Kumar]{snell2025scaling}
Charlie~Victor Snell, Jaehoon Lee, Kelvin Xu, and Aviral Kumar.
\newblock Scaling {LLM} test-time compute optimally can be more effective than
  scaling parameters for reasoning.
\newblock In \emph{The Thirteenth International Conference on Learning
  Representations}, 2025.
\newblock URL \url{https://openreview.net/forum?id=4FWAwZtd2n}.

\bibitem[Sutton(2019)]{sutton2019bitter}
Richard Sutton.
\newblock The bitter lesson.
\newblock \emph{Incomplete Ideas (blog)}, 13\penalty0 (1):\penalty0 38, 2019.

\bibitem[Sypetkowski et~al.(2024)Sypetkowski, Wenkel, Poursafaei, Dickson,
  Suri, Fradkin, and Beaini]{sypetkowski2024gnnscaling}
Maciej Sypetkowski, Frederik Wenkel, Farimah Poursafaei, Nia Dickson, Karush
  Suri, Philip Fradkin, and Dominique Beaini.
\newblock On the scalability of gnns for molecular graphs.
\newblock In A.~Globerson, L.~Mackey, D.~Belgrave, A.~Fan, U.~Paquet,
  J.~Tomczak, and C.~Zhang (eds.), \emph{Advances in Neural Information
  Processing Systems}, volume~37, pp.\  19870--19906. Curran Associates, Inc.,
  2024.
\newblock URL
  \url{https://proceedings.neurips.cc/paper_files/paper/2024/file/2345275663a15ee92a06bc957be54a2c-Paper-Conference.pdf}.

\bibitem[Thomas et~al.(2018)Thomas, Smidt, Kearnes, Yang, Li, Kohlhoff, and
  Riley]{thomas2018tensor}
Nathaniel Thomas, Tess Smidt, Steven Kearnes, Lusann Yang, Li~Li, Kai Kohlhoff,
  and Patrick Riley.
\newblock Tensor field networks: Rotation-and translation-equivariant neural
  networks for 3d point clouds.
\newblock \emph{arXiv preprint arXiv:1802.08219}, 2018.

\bibitem[Tong et~al.(2025)Tong, Hoang, Liu, Broeck, and Niepert]{tong2025rao}
Vinh Tong, Trung-Dung Hoang, Anji Liu, Guy Van~den Broeck, and Mathias Niepert.
\newblock Rao-blackwell gradient estimators for equivariant denoising
  diffusion.
\newblock \emph{arXiv preprint arXiv:2502.09890}, 2025.

\bibitem[Topping et~al.(2022)Topping, Giovanni, Chamberlain, Dong, and
  Bronstein]{topping2022oversquashing}
Jake Topping, Francesco~Di Giovanni, Benjamin~Paul Chamberlain, Xiaowen Dong,
  and Michael~M. Bronstein.
\newblock Understanding over-squashing and bottlenecks on graphs via curvature.
\newblock In \emph{International Conference on Learning Representations}, 2022.
\newblock URL \url{https://openreview.net/forum?id=7UmjRGzp-A}.

\bibitem[Wang et~al.(2024)Wang, Elhag, Jaitly, Susskind, and
  Bautista]{wang2024swallowing}
Yuyang Wang, Ahmed A.~A. Elhag, Navdeep Jaitly, Joshua~M. Susskind, and
  Miguel~\'{A}ngel Bautista.
\newblock Swallowing the bitter pill: Simplified scalable conformer generation.
\newblock In Ruslan Salakhutdinov, Zico Kolter, Katherine Heller, Adrian
  Weller, Nuria Oliver, Jonathan Scarlett, and Felix Berkenkamp (eds.),
  \emph{Proceedings of the 41st International Conference on Machine Learning},
  volume 235 of \emph{Proceedings of Machine Learning Research}, pp.\
  50400--50418. PMLR, 21--27 Jul 2024.
\newblock URL \url{https://proceedings.mlr.press/v235/wang24q.html}.

\bibitem[Wood et~al.(2025)Wood, Dzamba, Fu, Gao, Shuaibi, Barroso-Luque,
  Abdelmaqsoud, Gharakhanyan, Kitchin, Levine, et~al.]{wood2025uma}
Brandon~M Wood, Misko Dzamba, Xiang Fu, Meng Gao, Muhammed Shuaibi, Luis
  Barroso-Luque, Kareem Abdelmaqsoud, Vahe Gharakhanyan, John~R Kitchin,
  Daniel~S Levine, et~al.
\newblock Uma: A family of universal models for atoms.
\newblock \emph{arXiv preprint arXiv:2506.23971}, 2025.

\bibitem[Yang et~al.(2021)Yang, Hu, Babuschkin, Sidor, Liu, Farhi, Ryder,
  Pachocki, Chen, and Gao]{yang2021muP}
Ge~Yang, Edward Hu, Igor Babuschkin, Szymon Sidor, Xiaodong Liu, David Farhi,
  Nick Ryder, Jakub Pachocki, Weizhu Chen, and Jianfeng Gao.
\newblock Tuning large neural networks via zero-shot hyperparameter transfer.
\newblock In M.~Ranzato, A.~Beygelzimer, Y.~Dauphin, P.S. Liang, and J.~Wortman
  Vaughan (eds.), \emph{Advances in Neural Information Processing Systems},
  volume~34, pp.\  17084--17097. Curran Associates, Inc., 2021.
\newblock URL
  \url{https://proceedings.neurips.cc/paper_files/paper/2021/file/8df7c2e3c3c3be098ef7b382bd2c37ba-Paper.pdf}.

\bibitem[Yang et~al.(2024)Yang, Rao, Dehmamy, Walters, and
  Yu]{yang2024symminformed}
Jianke Yang, Wang Rao, Nima Dehmamy, Robin Walters, and Rose Yu.
\newblock Symmetry-informed governing equation discovery.
\newblock In A.~Globerson, L.~Mackey, D.~Belgrave, A.~Fan, U.~Paquet,
  J.~Tomczak, and C.~Zhang (eds.), \emph{Advances in Neural Information
  Processing Systems}, volume~37, pp.\  65297--65327. Curran Associates, Inc.,
  2024.
\newblock URL
  \url{https://proceedings.neurips.cc/paper_files/paper/2024/file/77fa0e7d45c6687f1958de0b31e9fc05-Paper-Conference.pdf}.

\bibitem[Zhai et~al.(2022)Zhai, Kolesnikov, Houlsby, and
  Beyer]{zhai2022scaling}
Xiaohua Zhai, Alexander Kolesnikov, Neil Houlsby, and Lucas Beyer.
\newblock Scaling vision transformers.
\newblock In \emph{Proceedings of the IEEE/CVF conference on computer vision
  and pattern recognition}, pp.\  12104--12113, 2022.

\end{thebibliography}
\bibliographystyle{iclr2026_conference}
\clearpage
\newpage
\appendix
\section{Related Works}
\label{sec:related_works}
\paragraph{Neural Scaling Laws.} Numerous studies investigating the scaling behavior of neural networks \citep{ahmad1998scaling, henighan2020scaling, hoffmann2022chinchilla, kaplan2020scaling, utkarsh2022scalingmanifold} demonstrate a predictable relationship: performance improves as model size $N$, dataset size $D$, and computational budget $C$ increase. Various functional forms have been proposed to model these scaling laws. Using test error $\epsilon$ as the evaluation metric, \citet{cortes1993learning} and \citet{hestness2017deep} proposed the functional form $L = a x^{-b} + L_\infty$, where $b>0$, $L_\infty \ge 0$ denotes the irreducible error, and $x$ can be $N$, $D$, or $C$. While useful, this form may result in infinite error as the scaling variable $x$ approaches zero (e.g., for models performing random guessing in classification tasks). To address this limitation, \citet{zhai2022scaling} introduced a more general form, $L = a(x+c)^{-b} + L_\infty$, where the parameter $c$ represents an effective offset, indicating the scale at which performance significantly surpasses random guessing. The coefficients of these power laws have been empirically explored across various research domains and tasks, including autoregressive generative modeling \citep{henighan2020scaling, kaplan2020scaling, hoffmann2022chinchilla} and computer vision \citep{zhai2022scaling, henighan2020scaling, aladulmohsin2022language_vision_scaling_laws}. Further research by \citet{hoffmann2022chinchilla} and \citet{snell2025scaling} has improved methodologies for studying scaling laws, allowing for the determination of optimal scaling strategies to achieve the best performance on specific tasks under given constraints. Additionally, \citet{caballero2023broken} introduced ``broken'' scaling laws to better model and extrapolate neural network scaling behaviours, particularly when the scaling functions exhibit non-monotonic transitions.

\paragraph{Scaling laws for MPNNs on Molecular Graphs.} Learning accurate molecular representations is a fundamental challenge in drug discovery and computational chemistry. Numerous studies have investigated the scalability of graph neural networks (GNNs), particularly message-passing neural networks (MPNNs), for predicting molecular properties. For 2D molecular graphs, prior research by \cite{liu2024graphscaling, sypetkowski2024gnnscaling, li2025scaling} has demonstrated the promising scalability of MPNNs, showing that network performance follows a power-law scaling behaviour with increases in both dataset and model sizes. Similar scaling trends have also been observed by \cite{frey2023neuralchemicalscaling, li2025scaling, wood2025uma} for $E(3)/SE(3)$ equivariant MPNNs trained on 3D atomistic systems. In contrast to these observations, \cite{pengmei2025pushing} demonstrate that scaling behaviours of geometric GNNs deviate from conventional power laws across different settings, including self-supervised, supervised, and unsupervised learning. 

Our work extends scaling-law analysis across NNIP architectures. Unlike \cite{wood2025uma}, which derives compute-optimal scaling for equivariant models on mixed materials–molecule datasets with periodic coordinates and other auxiliaries, we focus on molecules using only atomic 3D coordinates and atomic numbers. Despite using the eSEN same backbone, \citet{wood2025uma} report that, for dense models \footnote{The authors also study the effect of linear mixture-of-experts, while we don't consider this in our work.}, the compute-optimal strategy scales model size $N$ faster than data size $D$, whereas in our setting we observe nearly equal scaling between $N$ and $D$; though the tasks are different. Relative to \cite{frey2023neuralchemicalscaling}, our study uses a substantially larger force-field dataset, enabling more robust scaling estimates that consider different architectures.

\section{Supporting Figures}
\begin{figure}[h]
    \centering
      \begin{subfigure}[b]{0.3\textwidth} 
        \centering
        \includegraphics[width=\linewidth]{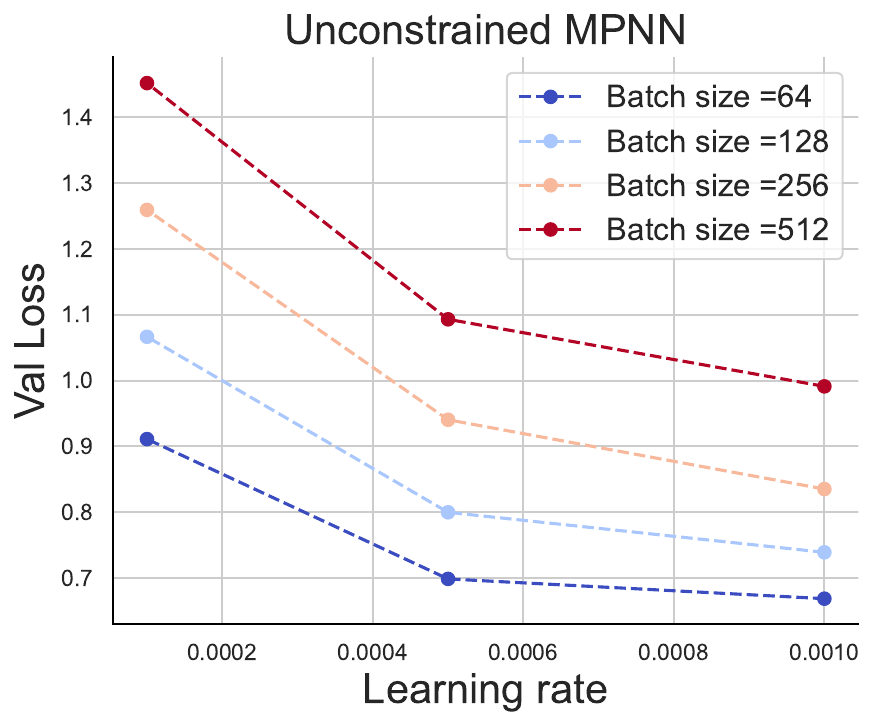}
    \end{subfigure}
    \begin{subfigure}[b]{0.3\textwidth}
        \centering
        \includegraphics[width=\linewidth]{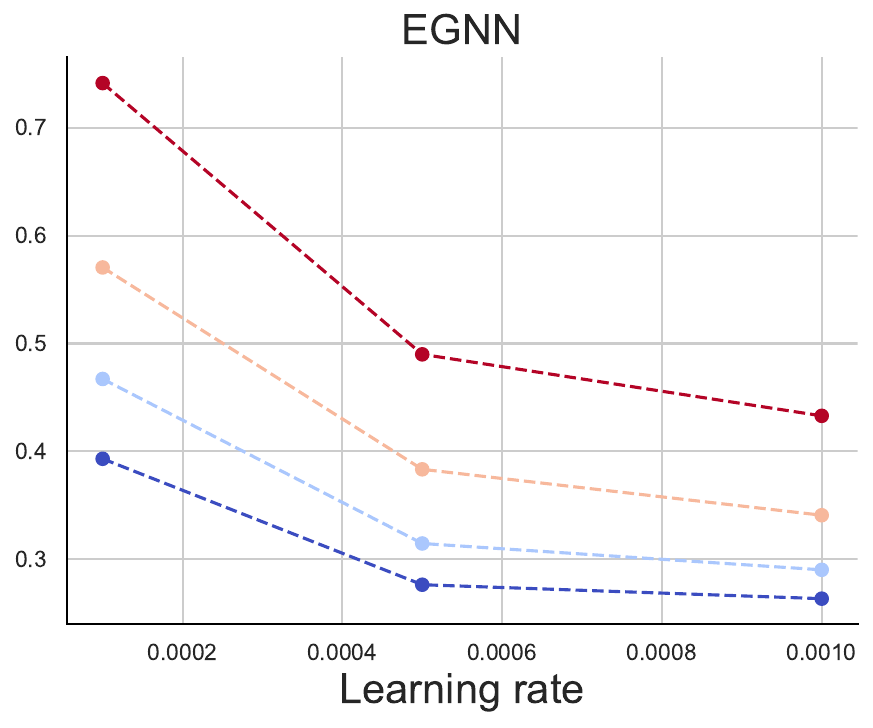}
    \end{subfigure} 
    \begin{subfigure}[b]{0.3\textwidth}
        \centering
        \includegraphics[width=\linewidth]{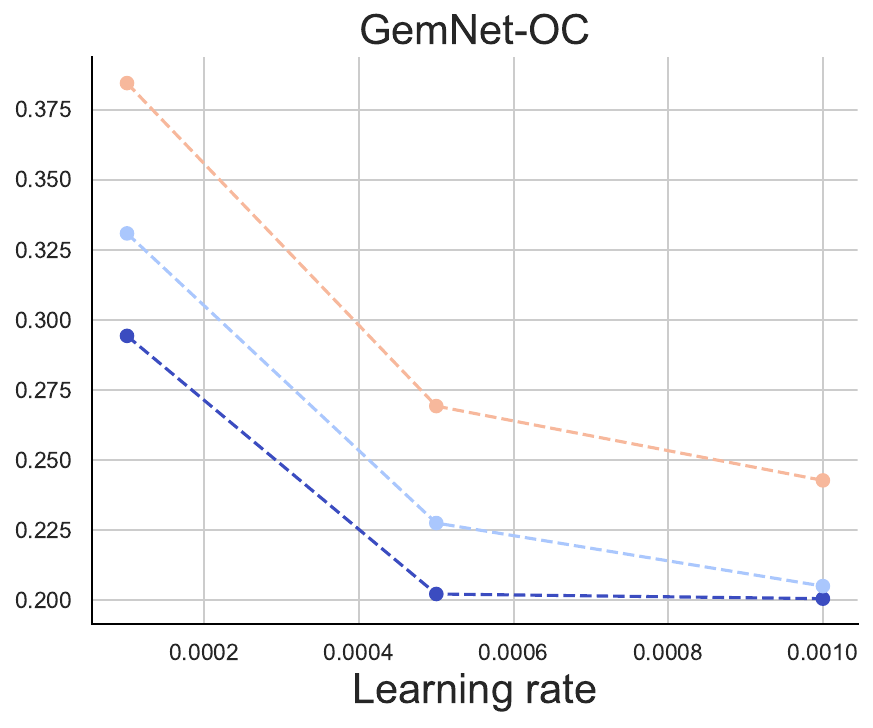}
    \end{subfigure}
    \begin{subfigure}[b]{0.29\textwidth} 
        \centering
        \includegraphics[width=\linewidth]{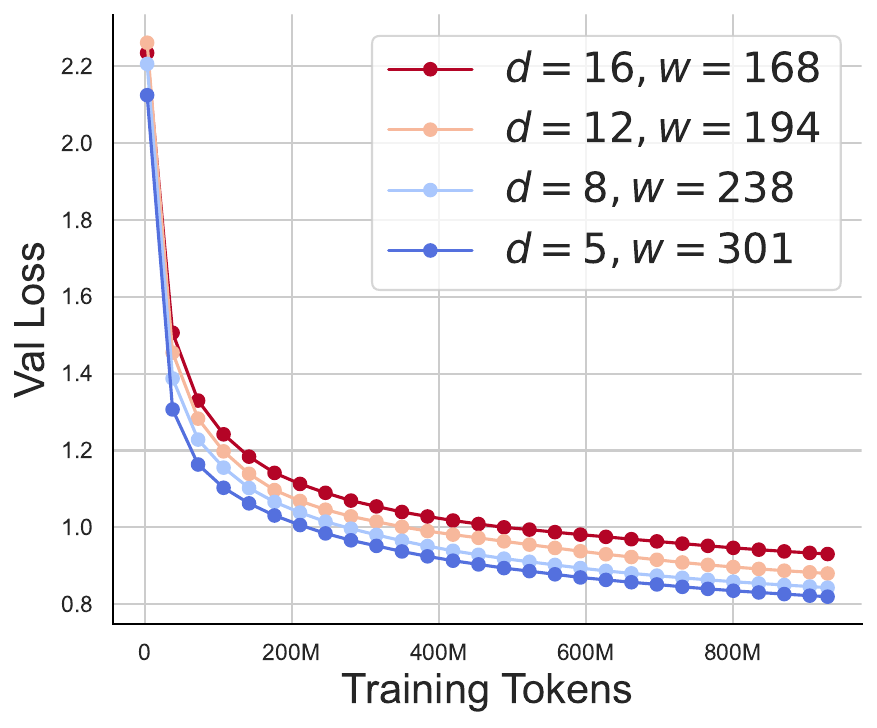}
        \label{fig:subfigA}
    \end{subfigure}
     % Adds horizontal space between subfigures
    \begin{subfigure}[b]{0.29\textwidth}
        \centering
        \includegraphics[width=\linewidth]{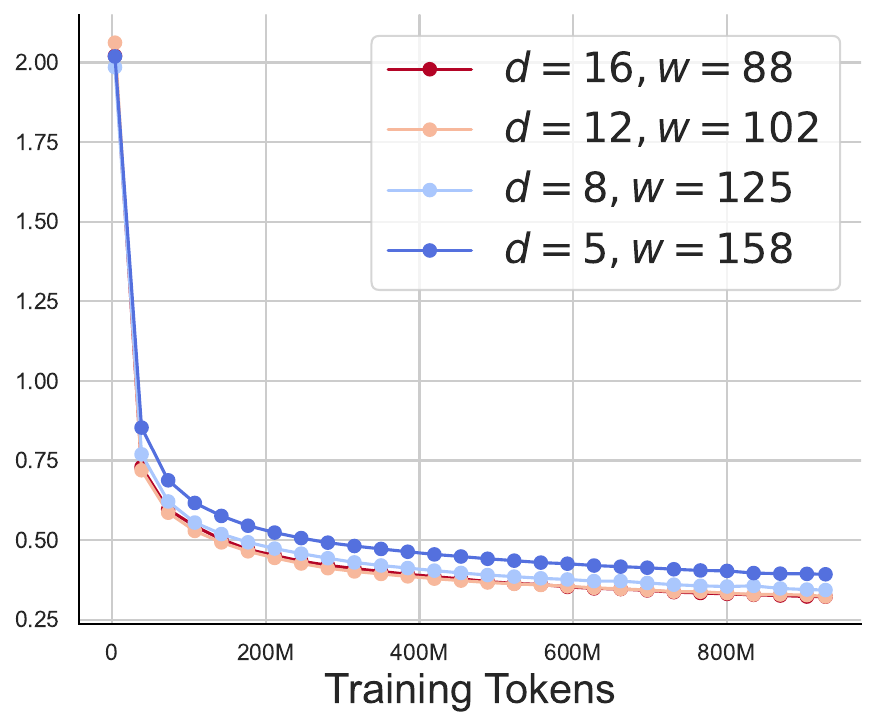}
        \label{fig:subfigB}
    \end{subfigure} 
    \begin{subfigure}[b]{0.29\textwidth}
        \centering
        \includegraphics[width=\linewidth]{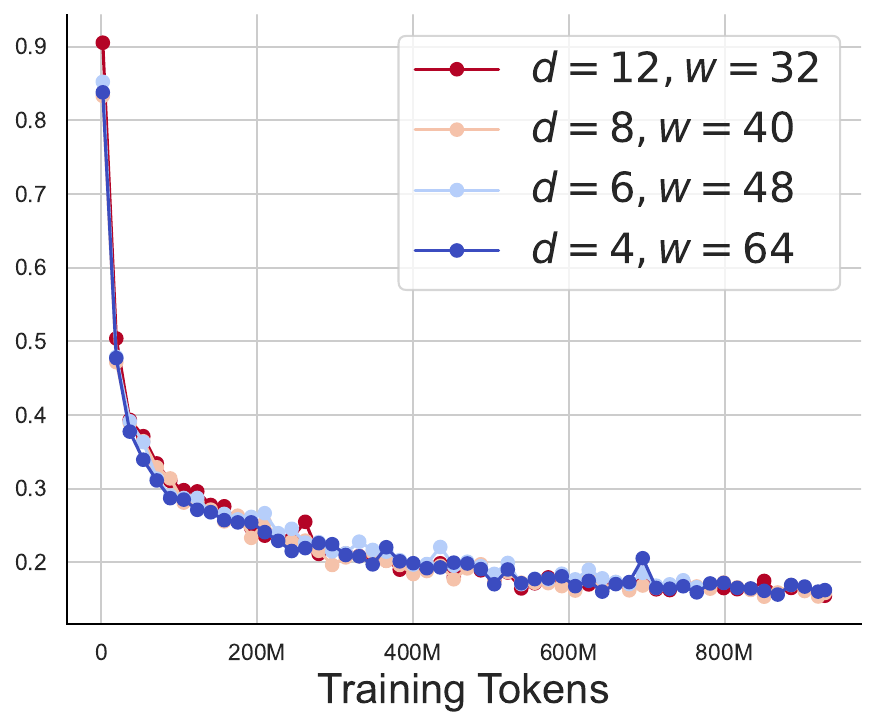}
        \label{fig:subfigB}
    \end{subfigure} 
    \vspace{-10pt}
    \caption{A sweep over batch size-learning rate (top row), and depth-width (bottom row) for three architectures with numbers of parameters are approximately equal to $1M$.}
    \label{fig:batch_size_learning_rate}
\end{figure}

\section{Design Principles for Geometric Message Passing}
\label{sec:background_mpnn}
\begin{table}[h]
    \caption{Architectures and their expressivity.}
    \centering
    \begin{tabular}{lccc}
        \toprule 
         Architectures  & Characteristic & Tensor Order ($\ell$) & Body Order ($\nu$) \\
         \midrule
        Unconstrained MPNN & unconstrained & -  & 2 \\   
        GemNet-OC & invariant & 0 & 4 \\
         EGNN & equivariant & 1 & 2 \\
         eSEN & equivariant & $\ge 2$ & 2 \\
         \bottomrule
    \end{tabular}
    \label{tab:model_expressivity}
\end{table}

Following \cite{joshi2023gwl}, we categorize the models and their expressivity in \Cref{tab:model_expressivity}. The tensor order $\ell$ denotes the order of the geometric tensor embeddings processed by each model. The body order refers to the number of nodes participating in a message function. Because GemNet-OC incorporates dihedral-angle information, its two-hop messages depend simultaneously on the states of four nodes, and thus it is classified as four-body. By contrast, the remaining architectures use one-hop message passing that depends only on the source and target nodes, i.e., two-body.

Let $\{h_1^{(t)}, \dots, h_n^{(t)}\}$ denote the node embeddings at layer $t$, with $h_i^{(0)}$ initialized from input features $z_i$. For a distance cutoff $c>0$, define the neighborhood
\[
\mathcal{N}(v)\;=\;\{\,u\neq v \mid \|x_u - x_v\|\le c\,\}.
\]
We update node $v$ by aggregating messages from its neighbors using a permutation-invariant (mean) aggregator, making the overall layer permutation-equivariant:
\begin{equation}
h_v^{(t+1)} \;=\; \frac{1}{|\mathcal{N}(v)|}\sum_{u\in \mathcal{N}(v)} m_{u\to v}^{(t)},
\qquad
m_{u\to v}^{(t)} \;=\; \phi_m^{(t)}\!\big(\cdot),
\label{eq:message_passing}
\end{equation}
Let $r_{uv} := x_u - x_v$ be the relative position vector and $w := \dim h_v^{(t)}$ be the embedding width. The symmetry properties (e.g., $E(3)$/$SE(3)$ invariance or equivariance) are determined by architectural choices in $\phi_m^{(t)}$, in particular, how it uses $r_{uv}$ (e.g., through rotational invariants such as $\|r_{uv}\|$ or via equivariant tensor constructions). We discuss specific message constructions in the next section.
\subsection{Unconstrained Message Passing}
Following \citet{duval2023stochasticframe}, geometry is injected directly:
\begin{equation}
m_{u\to v}^{(t)}=\phi_m\!\big(h_u^{(t)},\,h_v^{(t)},\,r_{uv},\,\|r_{uv}\|_2\big),
\label{eq:non_equivariant}
\end{equation}
with $h_u^{(t)}, h_v^{(t)} \in \mathbb{R}^w$ and $\phi_m: \mathbb{R}^{2w + 4} \to \mathbb{R}^w$ is an MLP. Because raw vectors ${r}_{uv}$ are processed without symmetry constraints, rotational equivariance is not enforced, and thus the node embeddings $h_v^{(t)}$ are not rotationally invariant. 

\subsection{Directional Message Passing}
GemNet-T \citep{gemnet} and GemNet-OC \citep{gemnet_oc} construct messages from multi–body $E(3)$-invariant geometric features, including pairwise distances, bond (three-body) angles, and dihedral (four-body) angles. Define the bond angle
$\varphi_{uvk} := \angle\!\big(r_{vu}, r_{vk}\big)$ and the dihedral angle
$\omega_{uvkj}$ as the angle between the planes $(u,v,k)$ and $(v,k,j)$
(e.g., via normals $n_1 \propto r_{vu}\!\times\! r_{vk}$ and $n_2 \propto r_{kj}\!\times\! r_{kv}$).
Then the message from $u$ to $v$ is
\begin{equation}
    m_{u \to v}^{(t)}
    \;=\;
    \sum_{\substack{k \in \mathcal{N}(v)\setminus \{u\} \\[2pt]
                      j \in \mathcal{N}(k)\setminus \{u,v\}}}
    \phi_m\!\Big(h_u^{(t)},\, h_v^{(t)},\, \|r_{uv}\|_2,\, \varphi_{uvk},\, \omega_{uvkj}\Big),
\end{equation}
where $h_u^{(t)}, h_v^{(t)} \in \mathbb{R}^w$ are scalar node embeddings and
$\phi_m$ is a learnable function operating on $E(3)$-invariant inputs (distance, angles) together with scalar features. Because the geometric inputs are $E(3)$-invariant (rotation/translation invariant) and $h$’s are scalar channels, the resulting message is $E(3)$-invariant as well. Linear and bilinear interactions inside $\phi_m$ do not affect this invariance so long as they act on invariant/scalar quantities. 

\subsection{Cartesian-Vector Message Passing}
The original EGNN uses a single vector channel (node coordinates) \citep{egnn}, which limits expressive power \citep{joshi2023gwl}.
To address this, we use a multi-channel extension, MC-EGNN \citep{levy2023using}. Each atom $v$ carries both scalar features
$h_v^{(t)} \in \mathbb{R}^w$ and $E$ vector channels $X_v^{(t)} \in \mathbb{R}^{3\times E}$, with the relative vector
$X_{uv}^{(t)} := X_u^{(t)} - X_v^{(t)} \in \mathbb{R}^{3\times E}$.

MC-EGNN maintains invariance/equivariance by updating the invariant node embeddings $h_v$ and the multi-channel equivariant vectors
$X_v$ using invariant messages. In particular, messages depend only on rotation–translation invariants:
\begin{equation}
m_{u\to v}^{(t)}
= \phi_m\!\big(h_u^{(t)},\, h_v^{(t)},\, \|X_{uv}^{(t)}\|_{e}^{\,2}\big),
\qquad
\phi_m:\mathbb{R}^{\,2w+E}\!\to\!\mathbb{R}^{\,w},
\label{eq:egnn_inv_message}
\end{equation}
where $\|X_{uv}\|_{e}\in\mathbb{R}^E$ denotes the channel-wise Euclidean norm (applied over the $3$ spatial components).
The invariant message $m_{u\to v}^{(t)}$ from \Cref{eq:egnn_inv_message} is then used in \Cref{eq:message_passing} to update
$h_v^{(t+1)}$. In addition, MC-EGNN updates the vectors via a channel mixer:
\begin{equation}
X_v^{(t+1)} \;=\; X_v^{(t)} \;+\;
\frac{1}{|\mathcal{N}(v)|}\sum_{u\in\mathcal{N}(v)} \frac{1}{E}\,
X_{uv}^{(t)}\,\Phi_x\!\big(m_{u\to v}^{(t)}\big),
\label{eq:egnn_vector_update}
\end{equation}
where $\Phi_x$ is a linear map with weights $W_x:\mathbb{R}^{\,w}\!\to\!\mathbb{R}^{\,E\times E'}$ followed by a reshape. This channel mixing preserves
equivariance because $E(3)$ actions (and permutations) act on the $3$D indices but \emph{not} on the channel index.

\paragraph{$\Theta(1)$-Variance Scaling.} Our empirical analyses suggest that increasing $E$ significantly improves performance; see \cref{sec:increase_vector_channels}. This, in turn, suggests that studying scaling laws requires scaling both the invariant dimension $w$ and the equivariant channels $E$. However, very large $E$ can be computationally expensive, and scaling arbitrarily may cause exploding gradients due to the matrix-valued function $\Phi_x$ and the matrix product $X_{uv}\Phi_x(m_{u\to v})$. To scale properly, one should ensure that each layer's output scales as $\Theta(1)$ and gradient update scales as $\Theta(1)$ \citep{yang2021muP}, i.e., are invariant across widths. Let $W_x \sim \mathcal{N}(0,\sigma^2)$ with $\sigma = \Theta(\sqrt{\min(w,EE^\prime)/w^2})$. Setting $E = E^\prime \approx \sqrt{w}$ yields $\sigma \approx \Theta(1/\sqrt{w})$. Under $\mu$P, we assume the entries of $m_{u\to v}$ and $X_{uv}$ have variance $\Theta(1)$; then $W_x m_{u\to v}$ also has $\Theta(1)$-entries, and since \textit{reshape} has no parameters, the entries of $\Phi_x(m_{u\to v})$ remain $\Theta(1)$. For the matrix product, because both $X_{uv}$ and $\Phi_x(m_{u\to v})$ have $\Theta(1)$-variance entries, $(X_{uv}\Phi_x(m_{u\to v}))_{3\times E}$, which sums over $E$, scales as $\Theta(E)$. To keep gradient updates stable across widths, we scale $X_{uv}\Phi_x(m_{u\to v})$ by $1/E \approx 1/\sqrt{w}$, instead of $1/\sqrt{E}$, which has the same effect as scaling the logits of dot-product attention by $1/\text{emb\_dim}$ rather than $1/\sqrt{\text{embed\_dim}}$ as discussed by \cite{yang2021muP}. 

\subsection{High-order Tensor Message Passing}
\paragraph{Irreducible Representations.}
High-order equivariant models (e.g., \citet{thomas2018tensor,anderson2019cormorant,nequip,batatia2022mace,liao2024equiformerv,wood2025uma})  use $SO(3)$ irreducible representations (irreps) as node embeddings: 
\begin{equation}
h_u^{(t)} \;=\; \bigoplus_{\ell=0}^{\ell_{\max}} h_{u,\ell}^{(t)},\qquad
h_{u,\ell}^{(t)}\in \mathbb{R}^{C_\ell}\!\otimes\!\mathbb{V}^{(\ell)},\;\; \dim\mathbb{V}^{(\ell)}=2\ell+1,
\label{eq:so3_representation}
\end{equation}
with $\bigoplus$ denotes concatenation of multiple order-$\ell$ tensors that are expanded by $C_\ell$ channels, and thus total dimension is $d := \sum_{\ell=0}^{\ell_{\max}} C_\ell(2\ell+1)$. Assume all orders have the same number of channels $C$, the atom embedding has a size of $w = C(\ell_{\max}{+}1)^2$.

\paragraph{SO(3) Convolution.}
Let $\hat{{r}}_{uv}={r}_{uv}/\|{r}_{uv}\|$. Messages couple source irreps with spherical harmonics:

\begin{equation}
   m_{{u \to v}, \ell_3} ^{(t)} =  \sum_{\ell_1, \ell_2}  w_{\ell_1 \ell_2 \ell_3} \bigoplus_{m_3} \sum_{m_1, m_2}
 {h}_{u,\, (\ell_1,  m_1)}^{(t)}  C_{(\ell_1, m_1),\, (\ell_2 m_2)}^{\,(\ell_3 m_3)}
  Y_{\ell_2}^{m_2}\!\left(\hat{r}_{uv}\right)\,
  \label{eq:so3_convolution}
\end{equation}
where  $C_{\ell_1 m_1,\, \ell_2 m_2}^{\,\ell_3 m_3}$ are Clebsch–Gordan coefficients, $w_{\ell_1, \ell_2, \ell_3}$ are learnable weights, and $Y_\ell$ is the order-$\ell$ spherical harmonics of the unit displacement vector $\hat{r}_{uv}$, $|\ell_1 - \ell_3| \le \ell_2 \le |\ell_1 + \ell_3|$ and $m_i \in \{-\ell_i, \dots, \ell_i\}$.

\paragraph{Efficient Convolution.} eSCN/eSEN \citep{passaro2023reduce, fu2025eSEN} sparsify  \cref{eq:so3_convolution} by rotating vector $r_{uv}$ into an edge-aligned frame. Let $R_{uv} \in \mathbb{R}^{3\times3}$ such that $R_{uv}\hat{r}_{uv} = (0,1,0)$. Then $Y^{m_2}_{\ell_2}(R_{uv} \hat{r}_{uv}) = 0$ unless $m_2 = 0$. Therefore, we can simplify \cref{eq:so3_convolution} as: 
\begin{equation}
     m_{{u \to v}, \ell_3} ^{(t)} = D_{\ell_3}^{-1} \sum_{\ell_1, \ell_2} w_{\ell_1, \ell_2, \ell_3} \bigoplus_{m_3} \sum_{m_1}\tilde{h}^{(t)}_{u, (\ell_1, m_1)} C^{(\ell_3, m_3)}_{(\ell_1, m_1), (\ell_2, 0)}, 
\end{equation}
here $\tilde{h}^{(t)}_{(\ell_1, m_1)} = D_{\ell_1} h^{(t)}_{(\ell_1, m_1)}$ where we denote $D_{\ell_1} :=D_{\ell_1}(R_{uv}) $ and $D_{\ell_3} := D_{\ell_3}(R_{uv})$ denote Wigner-D matrices of order $\ell_1$ and $\ell_3$, respectively. The output is rotated back by $D_{\ell_3}$ to ensure equivariance, and without loss of generality, we re-scale $Y^{m_2}_{\ell_2}(R_{uv} \hat{r}_{uv})$ to 1. Given that $m_2 = 0$, $C_{(\ell_1,m_1),\, (\ell_2,0)}^{(\ell_3,m_3)}$ are non-zero only when $m_1 = \pm m_3$. This further simplifies the computation to: 
\begin{equation}
     m_{{u \to v}, \ell_3} ^{(t)} =  D_{\ell_3}^{-1}\sum_{\ell_1 ,\ell_2}w_{\ell_1, \ell_2, \ell_3}\bigoplus_{m_3} \Big ( \tilde{h}^{(t)}_{v, (\ell_1, m_3)} C^{(\ell_3, m_3)}_{(\ell_1, m_3), (\ell_2, 0)} +  \tilde{h}^{(t)}_{v, (\ell_1, -m_3)} C^{(\ell_3, m_3)}_{(\ell_1, -m_3), (\ell_2, 0)}\Big ).
     \label{eq:reduce}
\end{equation}
Rearranging \cref{eq:reduce}, we obtain: 
\begin{equation}
    m_{{u \to v}, \ell_3} ^{(t)} =   D_{\ell_3}^{-1} \sum_{\ell_1} \bigoplus_{m_3} \Big( \tilde{h}^{(t)}_{v, (\ell_1, m_3)} \sum_{\ell_2} w_{\ell_1, \ell_2, \ell_3}  C^{(\ell_3, m_3)}_{(\ell_1, m_3), (\ell_2, 0)}  + \tilde{h}^{(t)}_{v, (\ell_1, -m_3)} \sum_{\ell_2} w_{\ell_1, \ell_2, \ell_3}  C^{(\ell_3, m_3)}_{(\ell_1, -m_3), (\ell_2, 0)} \Big ).
\end{equation}
\cite{passaro2023reduce} propose to replace the Clesbh-Gordon coefficients with parameterized weights as: 
\begin{align}
    \tilde{w}_{m_3}^{(\ell_1, \ell_3)} &= \sum_{\ell_2} w_{\ell_1, \ell_2, \ell_3}  C^{(\ell_3, m_3)}_{(\ell_1, m_3), (\ell_2, 0)} = \sum_{\ell_2} w_{\ell_1, \ell_2, \ell_3}  C^{(\ell_3, -m_3)}_{(\ell_1, -m_3), (\ell_2, 0)}, \quad \text{for } m \ge 0, \\
    \tilde{w}_{-m_3}^{(\ell_1, \ell_3)} &= \sum_{\ell_2} w_{\ell_1, \ell_2, \ell_3}  C^{(\ell_3, m_3)}_{(\ell_1, -m_3), (\ell_2, 0)} = -\sum_{\ell_2} w_{\ell_1, \ell_2, \ell_3}  C^{(\ell_3, -m_3)}_{(\ell_1, m_3), (\ell_2, 0)}, \quad \text{for } m < 0.
    \label{eq:weights}
\end{align}
Plugging back this into \cref{eq:reduce}, we obtain:
\begin{equation}
    m_{{u \to v}, \ell_3} ^{(t)} =   D_{\ell_3}^{-1} \sum_{\ell_1}  y^{(\ell_1)}_{\ell_3}, 
\end{equation}
where:
\begin{align}
y^{(\ell_1)}_{\ell_3, 0} &=  \tilde{w}^{(\ell_1, \ell_3)}_{0} \tilde{h}^{(t)}_{u, (\ell_1, 0)}  \\
\begin{pmatrix}
        y^{(\ell_1)}_{(\ell_3, m_3)} \\
        y^{(\ell_1)}_{(\ell_3, -m_3)} 
    \end{pmatrix} &= \begin{pmatrix}
        \tilde{w}^{(\ell_1, \ell_3)}_{m_3} &  - \tilde{w}^{(\ell_1, \ell_3)}_{-m_3} \\
        \tilde{w}^{(\ell_1, \ell_3)}_{-m_3} & \tilde{w}^{(\ell_1, \ell_3)}_{m_3}
    \end{pmatrix} \cdot \begin{pmatrix}
        \tilde{h}^{(t)}_{u, (\ell_1, m_3)} \\
         \tilde{h}^{(t)}_{u, (\ell_1, -m_3)}
    \end{pmatrix} \quad \text{for } m_3 > 0.  
    \label{eq:parameterized_kernel}
\end{align}

Therefore, the overall computation is reduced to an equivalent $SO(2)$ linear operation with a parameterized kernel as in \cref{eq:parameterized_kernel}. In eSEN, SO(2) blocks are applied only for values $|m_3| \le m_{max} \le \ell_{max}$; we set $m_{max} = 2$ as similar to the default setting. 

\begin{table}[h]
\centering
\caption{Scaling ladder for unconstrained MPNN.}
\begin{tabular}{@{}ccc@{}}
\toprule
\textbf{Depth} & \textbf{Width} & \textbf{\# Params} \\ \midrule
5     & 128   & 222404    \\
5     & 256   & 838020    \\
5     & 320   & 1293284   \\
5     & 375   & 1763133   \\
5     & 441   & 2422704   \\
5     & 517   & 3311714   \\
5     & 607   & 4543769   \\
5     & 712   & 6226800   \\
5     & 835   & 8535023   \\
5     & 1150  & 16101729  \\
5     & 1349  & 22109506  \\
5     & 1583  & 30389712  \\
5     & 1857  & 41755644  \\
5     & 2179  & 57415632  \\
5     & 2557  & 78974296  \\ \bottomrule
\end{tabular}
\label{tab:scaling_ladder_mpnn}
\end{table}

\begin{table}[h]
\caption{Scaling ladder for EGNN. We scale the number of channels for equivariant vectors as $\sqrt{w}$, ensuring stable update and $\Theta(1)$-variance entries when scaling $w$.}
\centering
\begin{tabular}{@{}ccc@{}}
\toprule 
\textbf{Depth} $d$ & \textbf{Width} $w$ & \textbf{\# Params} \\ \midrule
12    & 32    & 156358    \\
12    & 48    & 306199    \\
12    & 56    & 403992    \\
12    & 64    & 516553    \\
12    & 81    & 786925    \\
12    & 102   & 1195553   \\
12    & 120   & 1590131   \\
12    & 160   & 2745453   \\
12    & 200   & 4231015   \\
12    & 300   & 9214218   \\
12    & 500   & 24959524  \\
12    & 600   & 35673024  \\
12    & 721   & 51092972  \\
12    & 800   & 63035228  \\ \bottomrule
\end{tabular}
\label{tab:scaling_ladder_egnn}
\end{table}

\section{Experimental Setup}
\label{sec:experimental_setup}
\paragraph{Graph Construction.} 
Except for GemNet-OC, geometric graphs are built with a radial cutoff of 6~\text{\AA} and a maximum of 30 neighbors per atom. For GemNet-OC, using the same cutoff occasionally failed on molecules where its high-order message passing requires a larger candidate neighborhood. Accordingly, we use a 10~\text{\AA} cutoff and cap the neighborhood at 50 neighbors for this architecture.
\paragraph{Direct-Force Prediction.} Although the pre-training performance of direct-force models does not always transfer seamlessly to certain downstream physical tasks, these models provide substantial efficiency gains during pre-training. We focus on the scaling behaviour of direct-force models in the pre-training regime, i.e., training across large collections of molecular systems, and therefore employ them consistently throughout our experiments. 
\paragraph{Energy Reference and Normalization.} Following \cite{wood2025uma}\footnote{\url{https://github.com/facebookresearch/fairchem/tree/main}}, we apply the same reference scheme to the energies. Then, we normalize those referenced energies as $e^\prime = (e - \mu)/\sigma$, where $\mu$ and $\sigma$ are the sample mean and standard deviation estimated from the training set. Atomic forces for atom $i$, $f_i^\prime = -\partial e /\partial x_i$, are invariant to constant energy shifts; accordingly, we rescale them by the same factor: $f_i’ = f_i/\sigma$.

\begin{table}[h]
\caption{Scaling ladder for eSEN. Since eSEN use high-order tensors expanded into $C$ channels each, the embedding width $w = (\ell_{\text{max}} + 1)^2C$.}
\centering
\begin{tabular}{@{}cccccc@{}}
\toprule
\textbf{Depth} & $\ell_{\text{max}}$ &  $m_{\text{max}}$ & \textbf{\# Channels} $C$ & \textbf{\# Hidden Channels} & \textbf{\# Params}  \\ \midrule
12    & 4  & 2  & 4  & 32   & 441778    \\
12    & 4  & 2  & 8  & 32   & 878434    \\
12    & 4  & 2  & 10 & 32   & 1100746   \\
12    & 4  & 2  & 12 & 32   & 1325714   \\
12    & 4  & 2  & 16 & 32   & 1783618   \\
12    & 4  & 2  & 20 & 32   & 2252146   \\
12    & 4  & 2  & 32 & 32   & 3721474   \\ \bottomrule
\end{tabular}
\label{tab:scaling_ladder_esen}
\end{table}
\begin{table}[h]
\centering
\caption{Scaling ladder for GemNet-OC.}
\begin{tabular}{@{}ccc@{}}
\toprule
\textbf{Depth} & \textbf{Width} & \textbf{\# Params} \\ \midrule
4     & 32    & 312992    \\
4     & 48    & 575472    \\
4     & 64    & 937280    \\
4     & 80    & 1398416   \\
4     & 96    & 1958880   \\
4     & 112   & 2618672   \\
4     & 128   & 3377792   \\
4     & 144   & 4236240   \\
4     & 160   & 5194016   \\
4     & 176   & 6251120   \\
4     & 192   & 7407552   \\
4     & 204   & 8340060  \\ \bottomrule
\end{tabular}
\label{tab:scaling_ladder_gemnet}
\end{table}

\paragraph{Training Hardware.} We trained all models across the different architectural families under identical hardware conditions on NVIDIA 40GB A100 GPUs of the same compute cluster - each run within a single unit. This setup naturally incorporates overheads such as data loading, CPU bottlenecks, and metrics logging. During training, we continuously monitored validation losses along with the corresponding wall-clock training times. The reported training times thus include forward and backward passes over training samples, as well as forward passes over validation samples during intermediate evaluation checkpoints. We chose batch sizes to strike a balance between performance and computational efficiency, as using extremely small batch sizes is impractical. Based on the empirical results in \Cref{fig:batch_size_learning_rate}, we set the batch size to 128 for EGNN, and 64 for the remaining architectures \footnote{For fair comparison, we double EGNN's reported training time.}. These batch sizes were fixed across model sizes within each model family. 
\paragraph{Scaling Ladder.} Tables \ref{tab:scaling_ladder_mpnn}, \ref{tab:scaling_ladder_egnn}, \ref{tab:scaling_ladder_esen}, \ref{tab:scaling_ladder_gemnet} detail model sizes of all architectures used in this study. 

\begin{table}[h]
\small
\caption{Scaling parameters for \textit{sum-power-law in \cref{eq:scaling_law_N_D}} with $95\%$ confidence intervals.}
\centering 
\begin{tabular}{lcccc}
\toprule 
\bf Architecture  & $\log_{10}(A)$ & $\log_{10}(B)$ & $\alpha$  & $\beta$ \\ \midrule
Unconstrained MPNN  & 1.356 [1.307-1.371] & 2.194 [2.147-2.608]  & 0.276 [0.266- 0.278] & 0.311 [0.301-0.368] \\
EGNN   & 1.582 [1.494,1.692]  & 2.750 [2.660-2.863]  & 0.387 [0.370-0.408] & 0.394 [0.382-0.401] \\
GemNet-OC & 2.109 [1.901-2.422]  & 3.261 [2.842-3.607]  & 0.524 [0.484-0.584]  & 0.499 [0.444-0.544] \\
eSEN  & 3.760 [3.119-4.333] & 5.129 [4.348-6.224] & 0.817 [0.706-0.918] & 0.753  [0.662-0.888] \\ \bottomrule
\end{tabular}
\label{tab:scaling_params_sum_power_law}
\end{table}

\section{Uncertainty in Scaling Laws.} 
\label{sec:uncertainty}
Due to the cost of training NNIPs, we perform scaling study within a range of compute and model sizes and do not consider tuning other hyper-parameters, such as weight decay. Thus, we construct $95\%$ confidence intervals on the fit parameters of \cref{eq:scaling_flops} and  \cref{eq:scaling_law_N_D} from 1000 non-parametric bootstraps. \Cref{tab:scaling_params_sum_power_law} demonstrates the values of fit parameters of \cref{eq:scaling_law_N_D} along with confidence intervals shown in parenthesis.

\section{Effect of Scaling Vector Channels} 
\label{sec:increase_vector_channels}
\begin{figure}[h]
    \centering
    \includegraphics[width=0.5\linewidth]{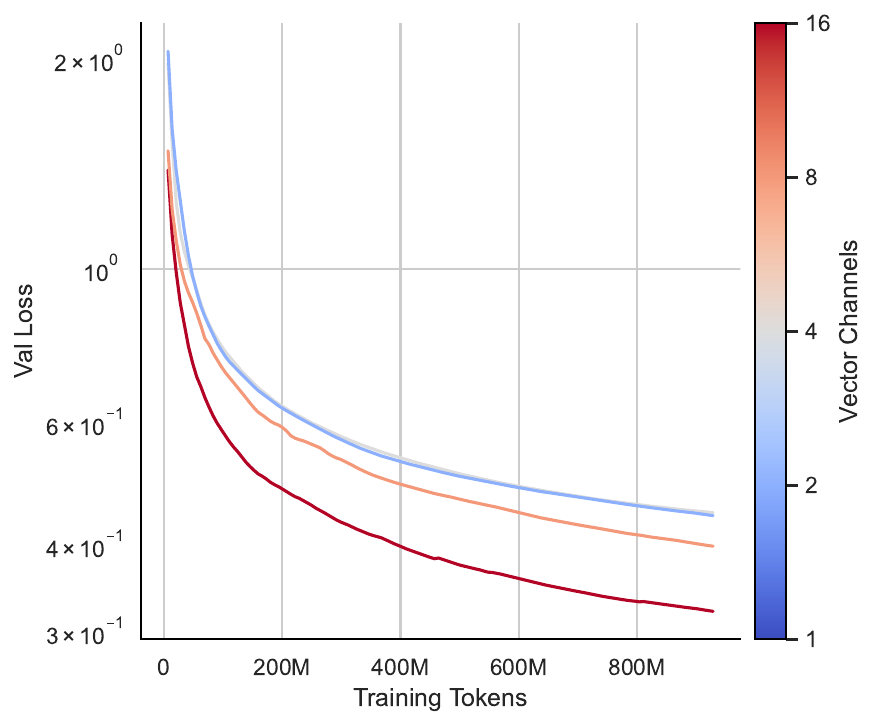}
    \caption{Effect of scaling number of equivariant channels in EGNN}
    \label{fig:scaling_vector_channels}
\end{figure}

We evaluate the effect of multi-channel vectors in EGNN by plotting validation loss against the number of vector channels $E$ in \cref{fig:scaling_vector_channels}. Loss consistently decreases as $E$ grows.

\section{{Symmetry Advantages on Datasets of diverse Molecular types}}
\label{sec:diverse_dataset}
\begin{figure}[h]
    \centering
      \begin{subfigure}[b]{0.32\textwidth} 
        \centering
        \includegraphics[width=\linewidth]{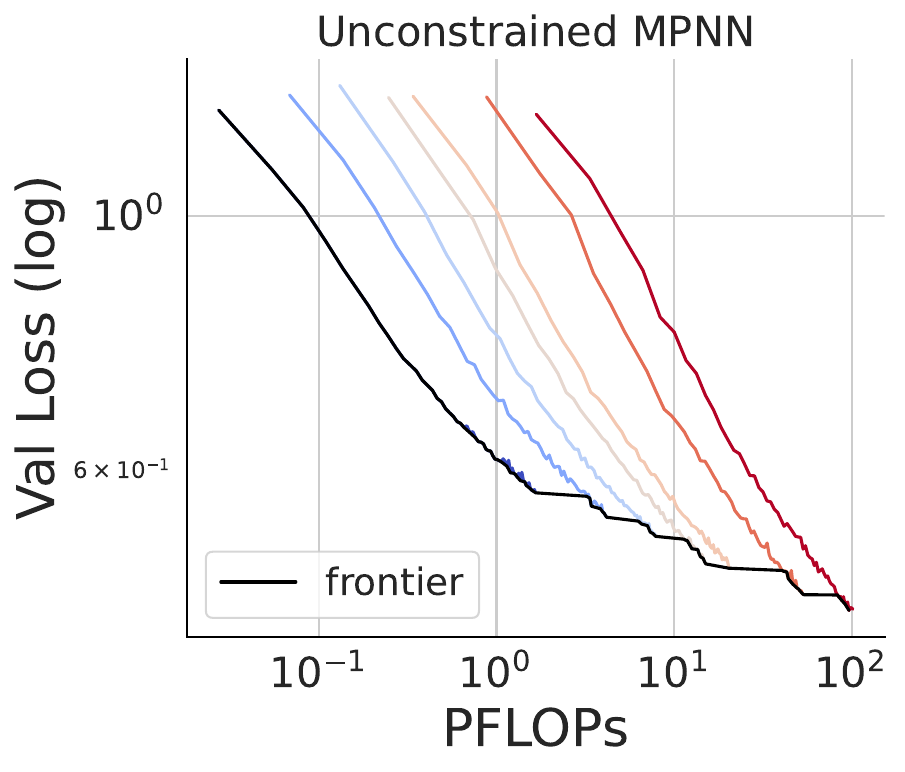}
        \label{fig:subfigA}
    \end{subfigure}
       \begin{subfigure}[b]{0.32\textwidth} 
        \centering
        \includegraphics[width=\linewidth]{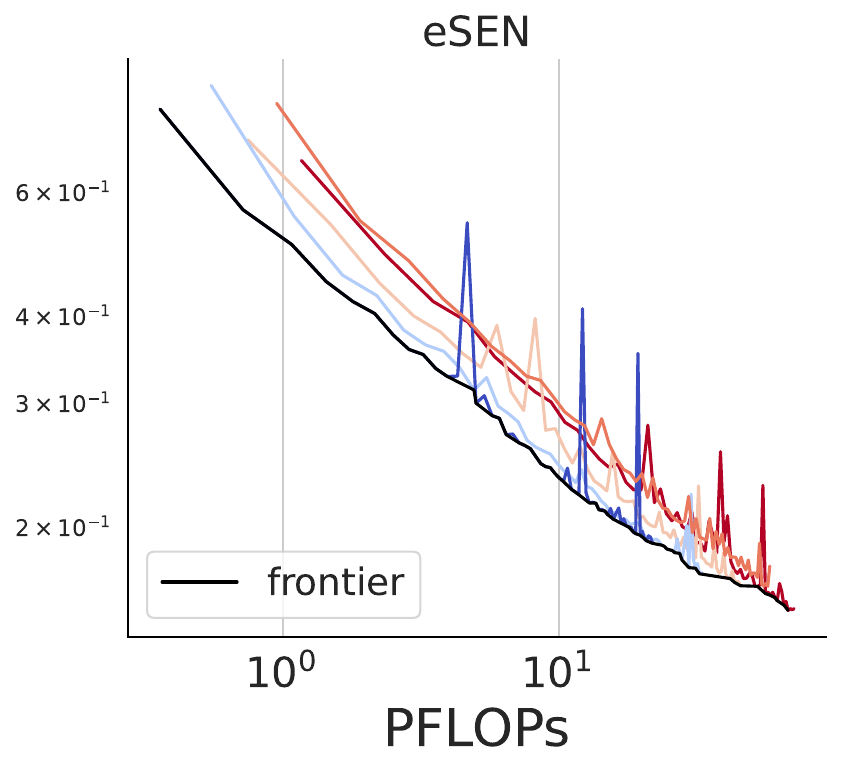}
        \label{fig:subfigA}
    \end{subfigure}
    \begin{subfigure}[b]{0.32\textwidth}
        \centering
        \includegraphics[width=\linewidth]{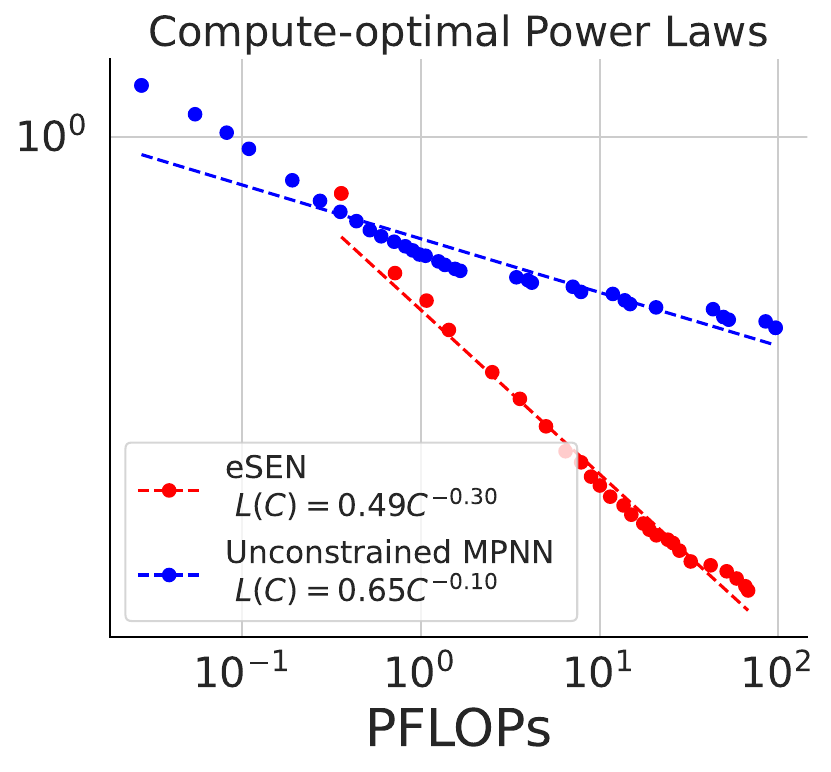}
        \label{fig:subfigB}
    \end{subfigure} 
    %\vspace{-10pt}
    \caption{{Loss-compute Pareto frontiers for unconstrained models (\textbf{Left}) and eSEN (\textbf{Middle}) trained on OMol-4M split. \textbf{Right}: The compute-optimal power laws reveal that the benefits of symmetry persist at scale, even when models are trained on datasets containing a high diversity of molecular types.}}
    \label{fig:scaling_laws_omol_4M}
\end{figure}

{In this experiment, we investigate whether the advantages of symmetry diminish when models are trained on datasets containing more diverse molecular types, e.g., electrolytes, metal complexes, and biomolecules rather than only neutral species. We train an unconstrained MPNN and eSEN on the 4M split of the OpenMol dataset, which is sampled uniformly across the diverse molecular types mentioned above. For validation, we use a held-out subset of $79$K samples from the entire 4M split, using the remainder for training. As shown in \cref{fig:scaling_laws_omol_4M}, we observe that the benefits gained by symmetry-aware design still persist even with this highly diverse dataset. Specifically, the difference in scaling exponents between the unconstrained MPNN and eSEN remains significant, a result similar to our findings for the neutral species split. It is important to note that the scaling exponents for each architecture itself may differ from those observed in the neutral split. This is expected because the coefficients of neural scaling laws are also dependent on the training dataset \citep{maloney2022solvable, bahri2024explaining_nn_scaling, bordelon2024dynamical_model}.}

\section{{Effect of Test-Time Augmentation in Scaling Laws}}\label{sec:test-time}
{An unconstrained model $\phi_\theta$ can be made equivariant at test time via group averaging (GA) with no additional training cost.  
\begin{equation}
    f_\theta(x) = \frac{1}{M} \sum_{i=1}^M \rho_{\text{out}}(g^{-1}_i) \phi_\theta(\rho_\text{in}(g_i) x), 
\end{equation} where $g_i \sim \mu_G$ is a sample from the Haar measure, $\rho_{\text{in}}$ and $\rho_{\text{out}}$ denote linear actions on input and output of $\phi_\theta$, respectively. 
In this section, we explore the effect of GA on the performance of unconstrained models at scale, especially when the model's parameter count and number of group elements increase.}

\begin{figure}[h]
    \centering
      \begin{subfigure}[b]{0.4\textwidth} 
        \centering
        \includegraphics[width=\linewidth]{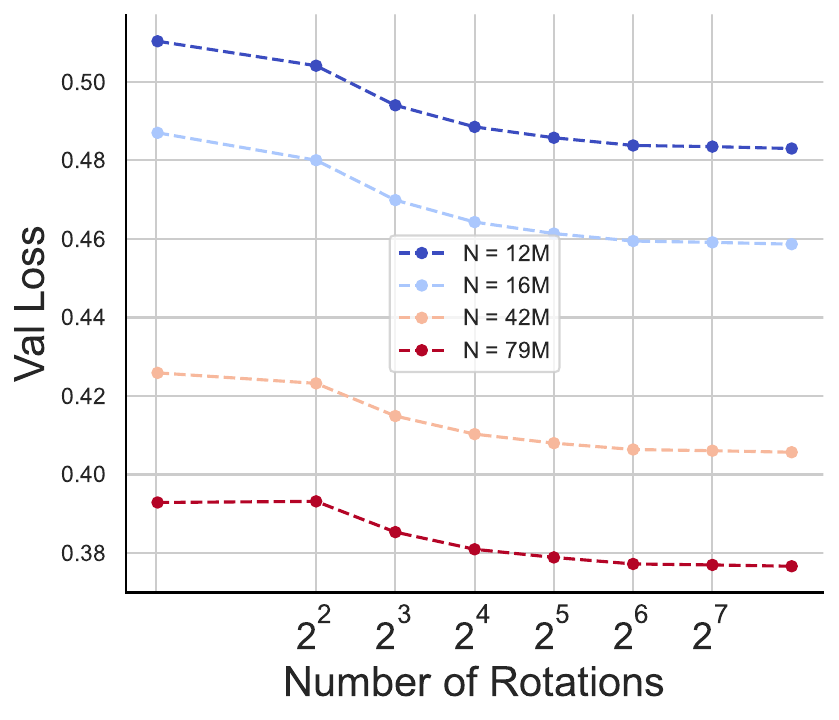}
        \label{fig:subfigA}
    \end{subfigure}
    \hspace{15pt}
       \begin{subfigure}[b]{0.4\textwidth} 
        \centering
        \includegraphics[width=\linewidth]{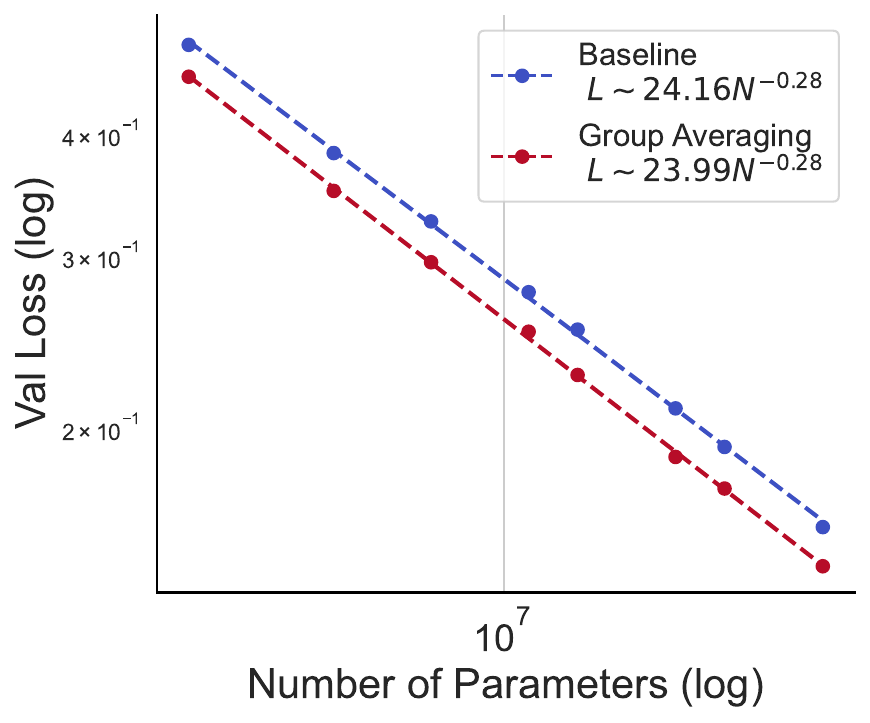}
        \label{fig:subfigA}
    \end{subfigure}
    \caption{{Scaling analysis for group averaging (GA) at test time. \textbf{Left}: The benefit derived from increasing the number of rotations saturates beyond a certain threshold. \textbf{Right}: Scaling performance with respect to parameter count shows that utilizing group averaging results in a slight downward shift of the linear trend in log-log space, while the critical scaling exponent remains unchanged.}}
    \label{fig:scaling_laws_test_time_augmentation}
\end{figure}

{\Cref{fig:scaling_laws_test_time_augmentation} shows that group averaging yields only a minimal improvement in the performance of $\phi_\theta$ across all model sizes. Furthermore, this improvement saturates as the number of rotations, $M$, increases beyond a certain threshold. To better understand this behavior, we fit scaling laws with respect to the number of parameters $N$ for both the baseline and the GA models: 
\begin{equation}
    L - L_D = AN^{-\alpha}.
\end{equation}}
{We use the parameter scaling relationship to analyze the impact of group averaging (GA). $L_D$ represents the lowest achievable loss for $\phi_\theta$ at a fixed data size $D$; see \cref{eq:sum_power_law}. For this analysis, we utilize models trained on the largest dataset, $D_{\max} \approx 9.2 \times 10^8$ (atoms). Based on the results in \cref{sec:scaling_laws}, the data-limited loss is $L_D \approx 1.6 \times 10^{-2} \times D_{\max} \approx 0.223$. We consider the GA performance at the onset of saturation, specifically at $M=32$. The fit results in \cref{fig:scaling_laws_test_time_augmentation} show that group averaging preserves the power-law exponent $\alpha$ that governs the scaling of $\phi_\theta$. As the parameter count increases, the performance shows only a slight downward shift in the log-log scaling relationship, due to a minor change in the multiplying constant $A$.}
\section{{Decomposing Energy and Force Loss}}
\label{sec:loss-decompose}
\begin{figure}[h]
    \centering
      \begin{subfigure}[b]{0.24\textwidth} 
        \centering
        \includegraphics[width=\linewidth]{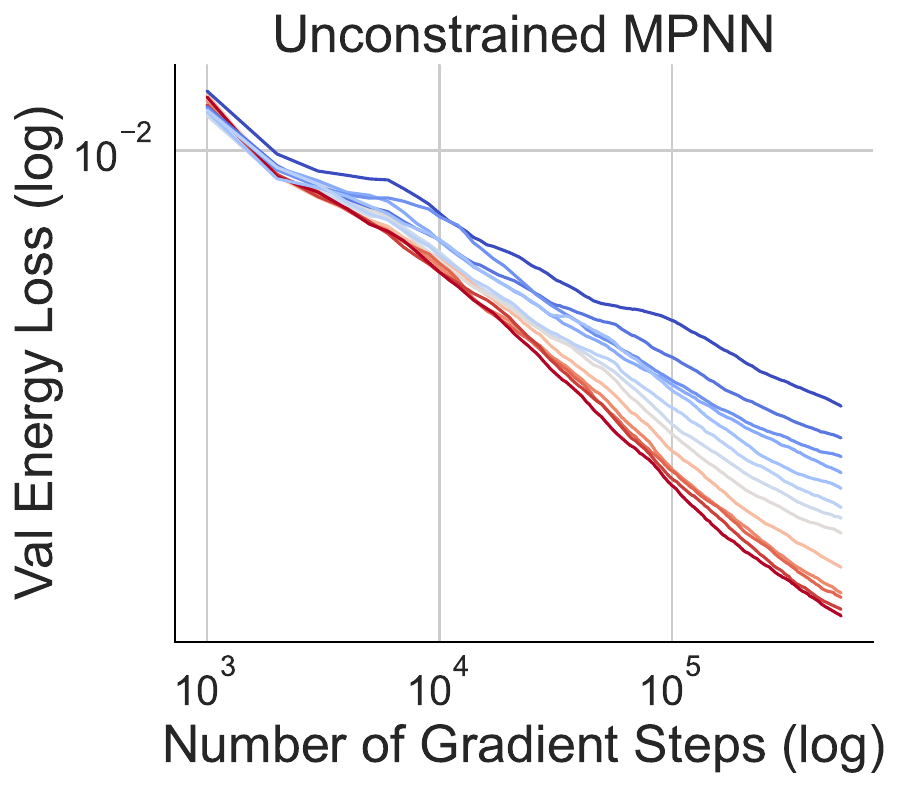}
        \label{fig:subfigA}
    \end{subfigure}
    \begin{subfigure}[b]{0.24\textwidth}
        \centering
        \includegraphics[width=\linewidth]{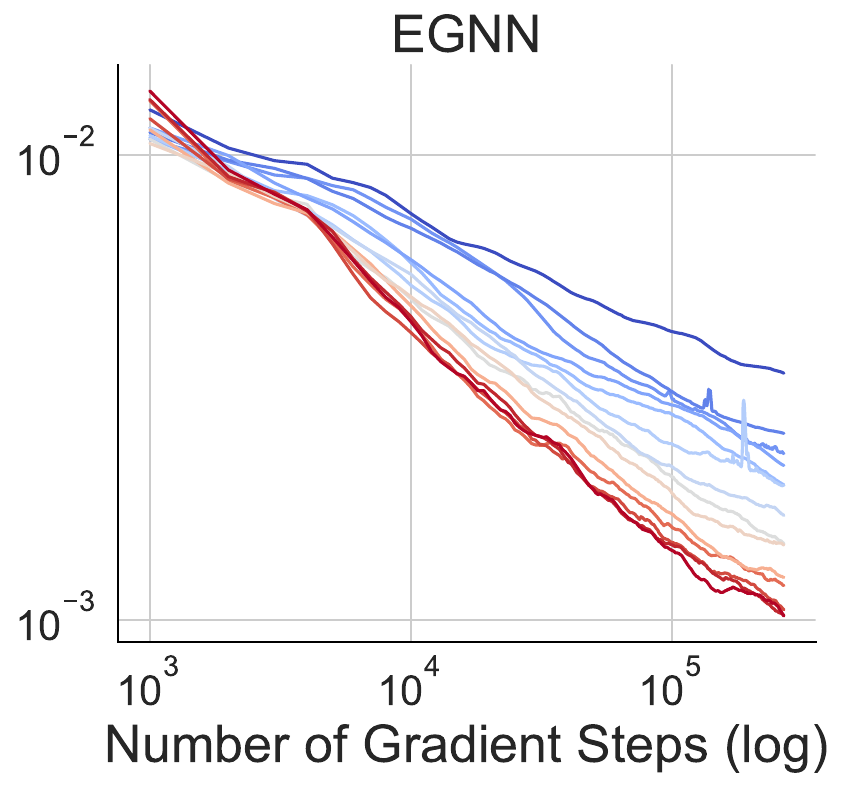}
        \label{fig:subfigB}
    \end{subfigure} 
       \begin{subfigure}[b]{0.24\textwidth}
        \centering
        \includegraphics[width=\linewidth]{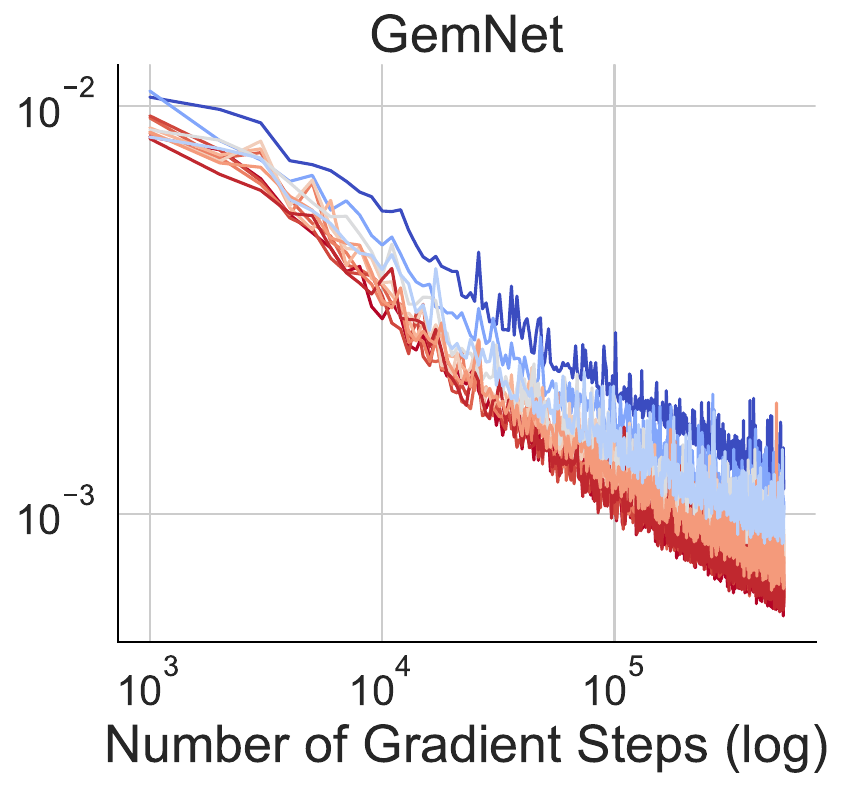}
        \label{fig:subfigB}
    \end{subfigure} 
    \begin{subfigure}[b]{0.24\textwidth}
        \centering
        \includegraphics[width=\linewidth]{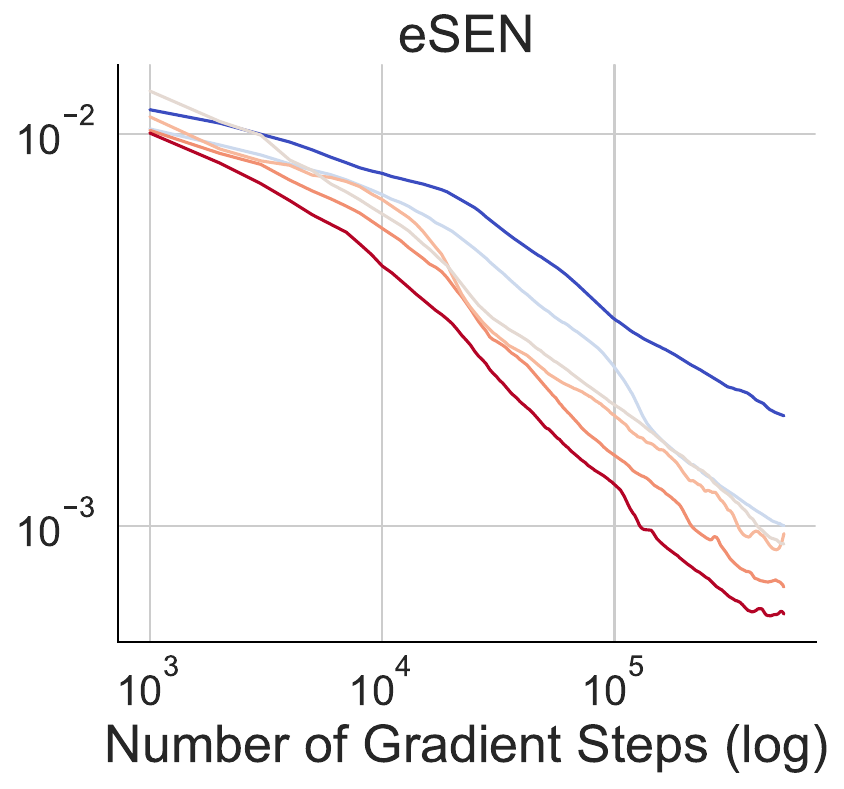}
        \label{fig:subfigB}
    \end{subfigure} 
    \begin{subfigure}[b]{0.24\textwidth} 
        \centering
        \includegraphics[width=\linewidth]{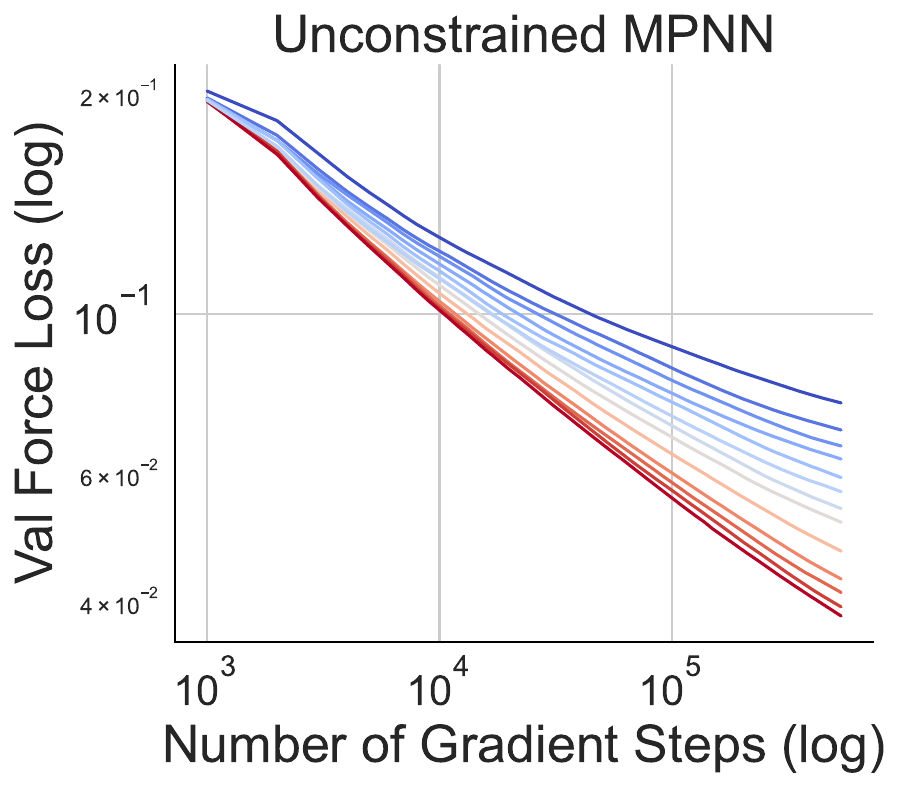}
        \label{fig:subfigA}
    \end{subfigure}
    \begin{subfigure}[b]{0.24\textwidth}
        \centering
        \includegraphics[width=\linewidth]{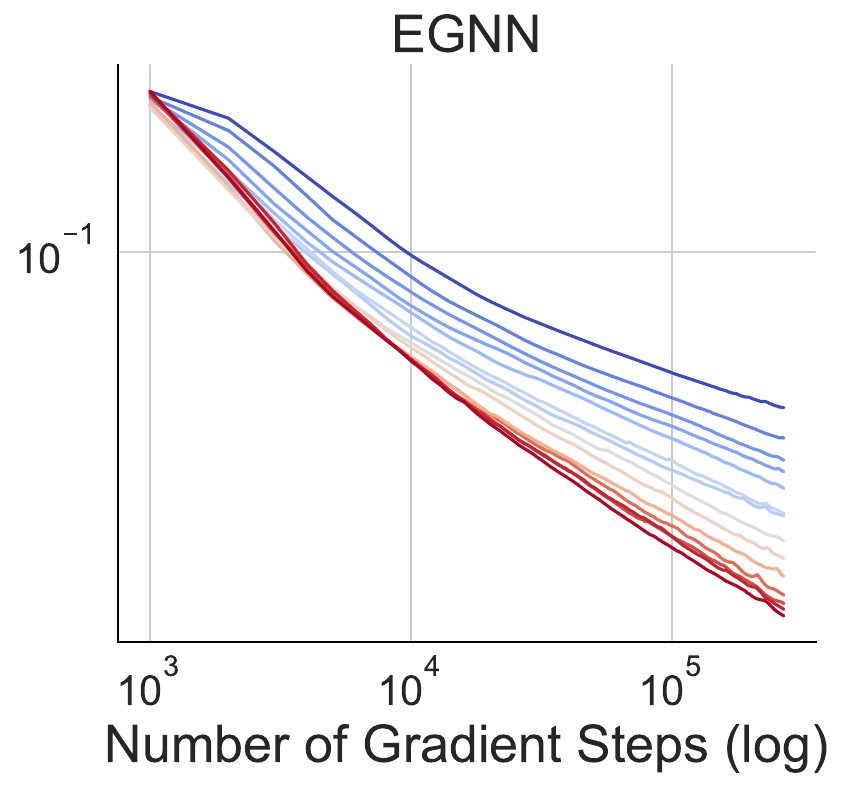}
        \label{fig:subfigB}
    \end{subfigure} 
       \begin{subfigure}[b]{0.24\textwidth}
        \centering
        \includegraphics[width=\linewidth]{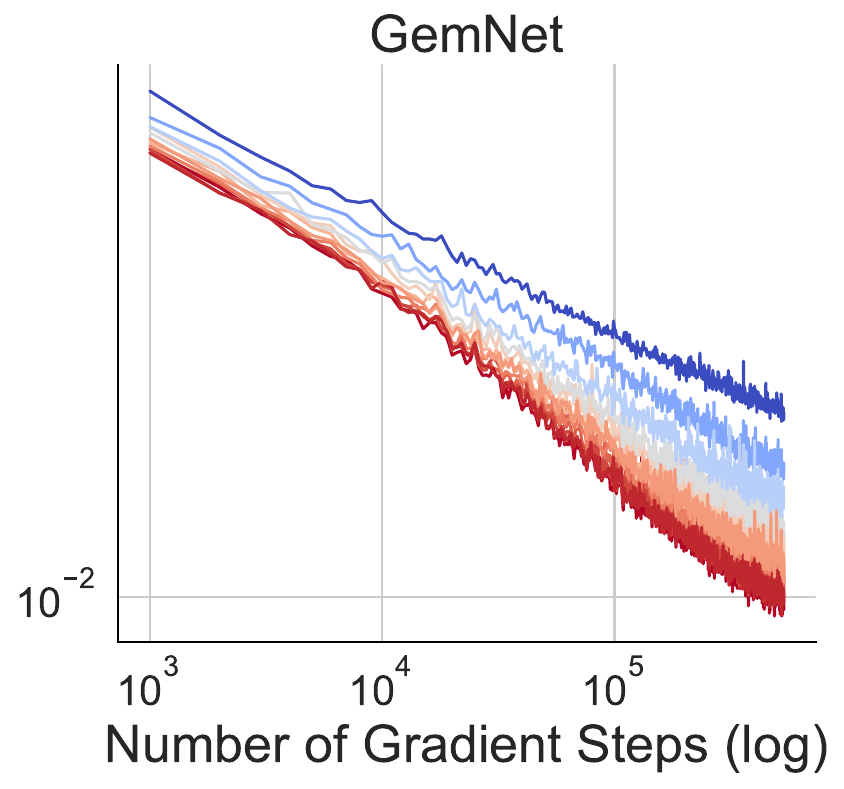}
        \label{fig:subfigB}
    \end{subfigure} 
    \begin{subfigure}[b]{0.24\textwidth}
        \centering
        \includegraphics[width=\linewidth]{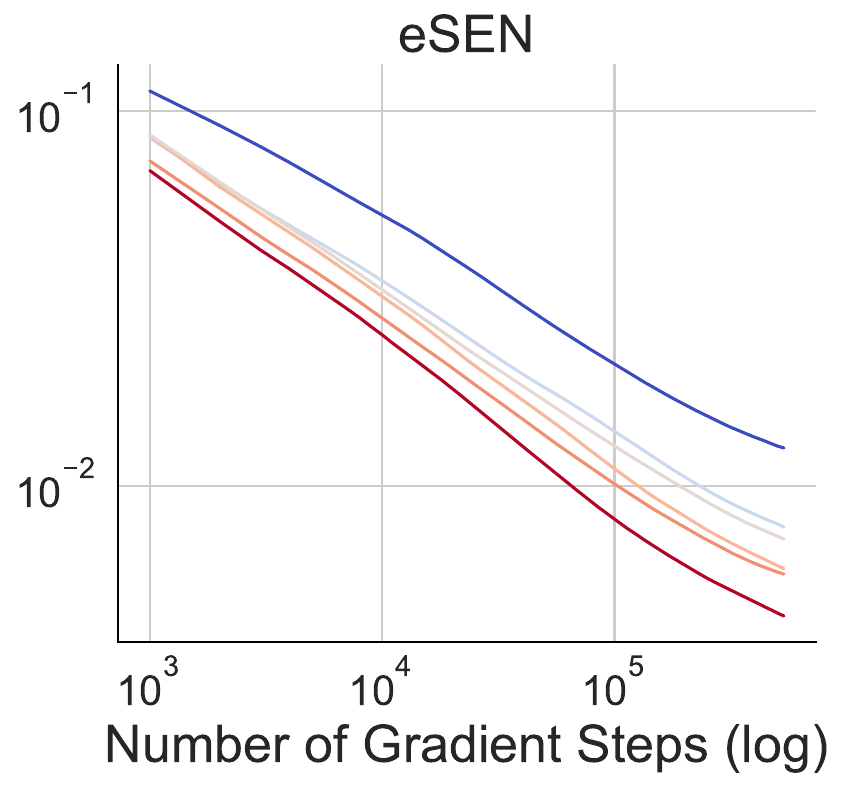}
        \label{fig:subfigB}
    \end{subfigure} 
    \vspace{-10pt}
    \caption{{Learning curves of all models, including Energy Loss (\textbf{Top}) and Force Loss (\textbf{Bottom})}. Line color encodes model size (\textcolor{blue}{small}, \textcolor{red}{large})}.
    \vspace*{-2em}
     \label{fig:energy_and_force_loss}
\end{figure}

{\Cref{fig:energy_and_force_loss} shows the decomposition of the total loss in \cref{eq:task_loss} into its energy and force components. We observe that the energy learning curves are relatively noisy, whereas the force losses remain smooth across model sizes for each architecture.}

\section{{Effect of translation invariance}}
\label{sec:translation_invariance}
\begin{figure}[h]
    \centering
    \includegraphics[width=0.5\linewidth]{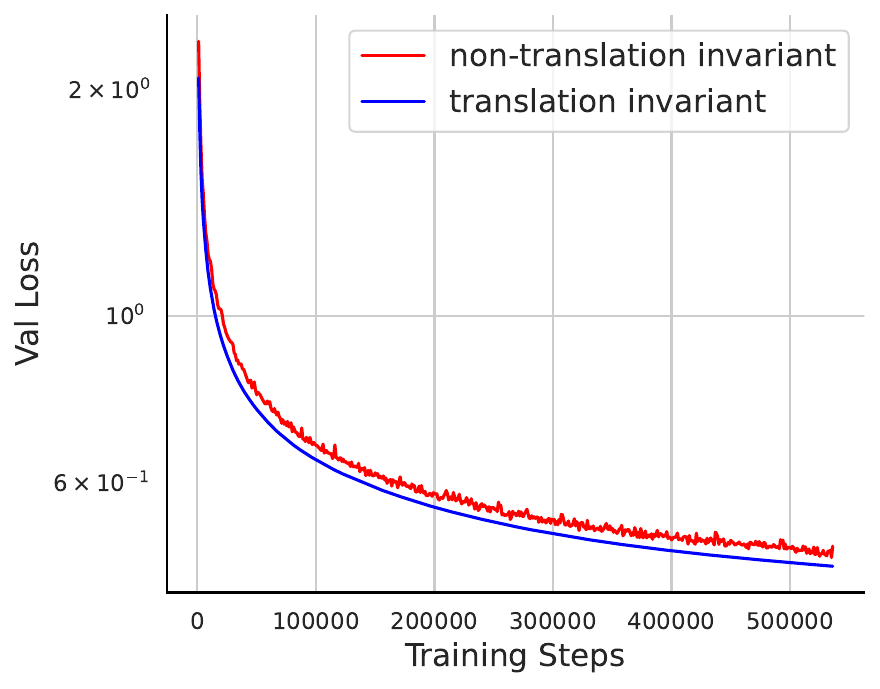}
    \caption{{MPNNs exhibit a performance drop when translation invariance is not maintained.}}
    \label{fig:placeholder}
\end{figure}

\section{{Training Instabilities of Vanilla Transformers}}
\label{sec:instabilites_transformer}
\begin{figure}[h]
    \centering
    \includegraphics[width=0.5\linewidth]{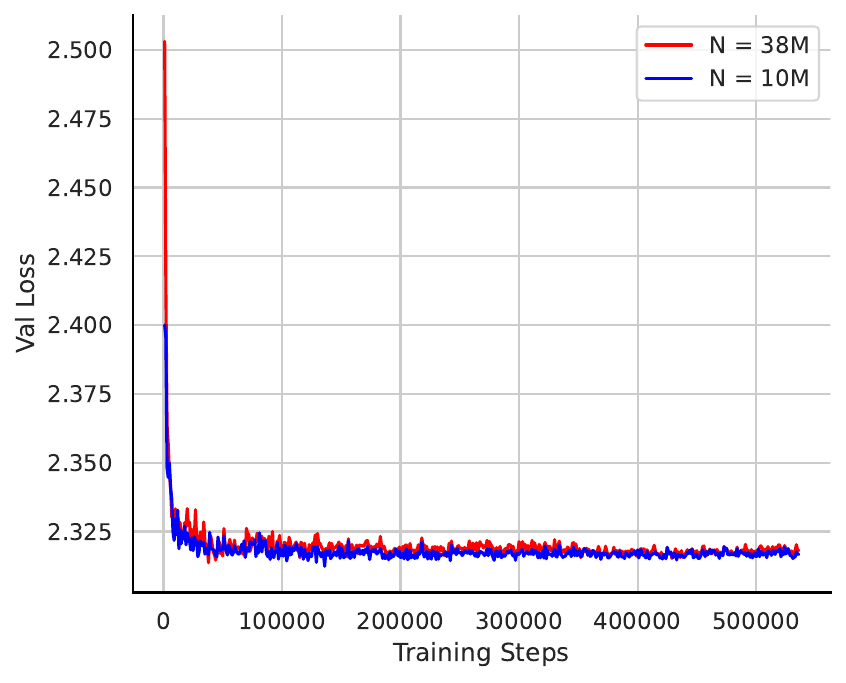}
    \caption{{Training instabilities of vanilla transformers on force-field tasks.}}
    \label{fig:transformer_result}
\end{figure}

{
We further push the limit of lacking inductive biases by training a vanilla transformer for force-field tasks. In particular, we train a GPT-style encoder in which atomic coordinates are fed directly into the model by concatenating them with embeddings computed from atomic numbers. As a result, E(3) equivariance is completely ignored in this setup. Moreover, because the transformer learns global attention over all atom nodes, the inductive bias of local neighborhoods, which is critical in NNIPs, is also abandoned. As shown in \Cref{fig:transformer_result}, vanilla transformers fail to exhibit meaningful learning, as the learning curves saturate rapidly and remain unchanged for the remainder of training.
}

\section{Usage of Large Language Models}
We used LLMs to assist with the writing of the paper (mainly for polishing) and also for coding.

\end{document}